%% file: main.tex
\documentclass[11pt]{article}

\usepackage{acl}

\usepackage{times}
\usepackage{latexsym}

\usepackage[T1]{fontenc}

\usepackage[utf8]{inputenc}

\usepackage{microtype}

\usepackage{inconsolata}

\usepackage{graphicx}

\usepackage{subcaption} 
\usepackage{enumitem}
\usepackage{makecell}        
\usepackage{booktabs}        
\usepackage[table,xcdraw]{xcolor} 
\usepackage{pifont}          
\usepackage{graphicx}        
\usepackage{caption}         
\usepackage{float}

\usepackage{amsmath,amssymb,amsfonts}
\usepackage{graphicx}
\usepackage{textcomp}
\usepackage{multirow}
\usepackage{hyperref}

\usepackage{makecell}

\usepackage{tabularx}
\usepackage{booktabs}
\usepackage[table]{xcolor}
\usepackage{array}
\usepackage{caption}  
\usepackage{booktabs}
\usepackage{booktabs}
\usepackage{graphicx}
\usepackage{subcaption}  
\usepackage{caption}
\usepackage{subcaption}
\usepackage{adjustbox}
\usepackage{colortbl}
\usepackage{graphicx}
\usepackage{natbib}

\usepackage{pifont}
\newcommand{\cmark}{\ding{51}}  
\newcommand{\xmark}{\ding{55}}  

\definecolor{softred}{RGB}{231,76,60}     
\definecolor{softgreen}{RGB}{46,204,113}  
\definecolor{softyellow}{RGB}{241,196,15} 

\usepackage[ruled,vlined]{algorithm2e}

\title{ChartAnchor: Chart Grounding with Structural-Semantic Fidelity and Data Recovery}

\author{Xinhang Li, Jingbo Zhou, Pengfei Luo, Yixiong Xiao, Tong Xu \\Baidu Research, USTC\\
\{lixinhang, zhoujingbo, luopengfei, xiaoyixiong\}@baidu.com,  {tongxu@ustc.edu.cn}}

\newcommand{\name}{ChartAnchor\xspace}

\begin{document}
\maketitle

\input{Section/abstract}

\input{Section/intro}

\input{Section/related}
\input{Section/MyBench}

\input{Section/dataset}

\input{Section/metrics}
\input{Section/exp}

\input{Section/conclusion}

\bibliography{main}
\appendix

\input{Section/appendix}

\end{document}

%% file: Section/abstract.tex
\begin{abstract}
Recent advances in multimodal large language models (MLLMs) highlight the need for benchmarks that rigorously evaluate structured chart comprehension. 
Chart grounding refers to the bidirectional alignment between a chart’s visual appearance and the structured semantics. This task requires models to produce a symbolic specification that faithfully captures the chart’s visual and structural intent, while also recovering the underlying tabular data with precise values and relationships. Chart grounding directly reflects a model’s capabilities in numerical reasoning, multimodal alignment, and structural reconstruction, and has several important applications in real-world scenarios.
Existing benchmarks, constrained by narrow chart diversity, isolated tasks, and incomplete evaluation frameworks, fail to holistically assess grounding. To address this, we propose \name, a comprehensive benchmark of 8k+ chart-table-code triples spanning 30 chart types drawn from diverse real-world and augmented sources. \name introduces two complementary tasks: chart-to-code generation and controlled chart-to-table reconstruction, enabling cross-validation of visual and numerical fidelity. A multi-level evaluation framework integrates semantic validation, stylistic analysis, and perceptual metrics to assess both structural and content-level correctness. Extensive experiments on MLLMs reveal critical limitations in numerical precision and code synthesis, emphasizing the need for structured reasoning beyond surface-level perception. By unifying symbolic and data-driven grounding, \name establishes a rigorous foundation for chart grounding, offering meaningful insights for advancing MLLMs in scientific, financial, and industrial domains.

\end{abstract}

%% file: Section/intro.tex
\section{Introduction}

“A picture is worth a thousand words.” Charts exemplify this principle by converting complex datasets into intuitive visual representations, enabling rapid and effective communication of quantitative insights. As a result, data visualizations are essential tools across domains such as science, finance, journalism, and public policy. With the rise of multimodal large language models (MLLMs), there is increasing interest in extending their capabilities to understand and reason about charts—by jointly interpreting visual features and underlying symbolic structures.

One of the core capabilities of MLLMs in chart understanding and reasoning lies in the challenge of chart grounding---aligning a chart’s visual presentation (marks, axes, layout, colours, chart type) with its structured semantics (tabular data, scales). A grounding model must therefore (i) recover the exact data values from the image (chart → table) and (ii) capture the visual–structural encoding that turns those values into graphics (chart → code). Although such structural encodings could be summarised in natural language, we formalise them as executable plotting code: code provides an unambiguous, machine-verifiable specification of marks, scales, and layout, whereas free-text descriptions are often underspecified and cannot guarantee identical rendering. Mastery of both tasks evidences genuine numerical reasoning, spatial-layout comprehension, and multimodal alignment.

Chart grounding has significant applications in real-world scenarios. Fundamental use cases include the extraction of numerical values from charts for in-depth data analysis, as well as the regeneration of chart code to facilitate the creation of similar visualizations using new data. More importantly, chart grounding enables MLLMs to develop a comprehensive understanding of charts, providing a foundation for more advanced tasks. Potential applications include: (1) chat-based chart modification, in which users issue natural language instructions to alter a chart’s visual presentation or underlying data; (2) chart-grounded retrieval, where detailed chart code and data derived from charts can be indexed for downstream tasks such as retrieval-augmented generation (RAG); and (3) multi-chart reasoning, where grounding information from multiple charts and natural language instructions are integrated to support complex reasoning tasks. Therefore, it is crucial to assess the chart grounding capabilities of MLLMs---a important dimension that has not yet been fully explored.

While recent chart-related benchmarks such as  visual QA (e.g., ChartQA\citep{masry2022chartqa}, PlotQA\citep{methani2020plotqa}) and the summarization (e.g., Chart-to-Text\citep{kantharaj2022chart}, ChartSumm~\citep{rahman2023chartsumm}) have advanced chart understanding, they evaluate only unstructured outputs—offering limited insight into a model’s ability to recover symbolic or structured content. Tasks like chart-to-table and chart-to-code begin to address structured aspects, but each only tells part of the story. Chart-to-table benchmarks assess data recovery but ignore style and structure. Conversely, chart-to-code (e.g. ChartMimic\citep{yang2024chartmimic}, Plot2Code\citep{wu2024plot2code}) benchmarks are predominantly focus on  visual appearance,
including layout, styling, and chart-surface text. While they may assess textual accuracy on the chart surface, these benchmarks still judge success primarily through visual similarity and do not enforce consistency with the underlying structured data, which poses serious risks in data-sensitive scenarios.

Critically, most tasks in chart understanding are evaluated in isolation. This fragmented approach allows models to exploit superficial shortcuts: for instance, question-answering systems may guess answers based solely on surface-level cues, without genuinely interpreting the underlying chart structure. While code-generation models may reproduce a chart’s appearance while silently altering the underlying data. Without unified evaluation, such partial successes can mask critical failures.  Compounding the issue, existing benchmarks are limited in scope --- they cover only a narrow range of chart types, exclude domain-specific formats, and depend on a single plotting library. This narrow design fails to capture the diversity of real-world visualizations and lacks robust, comprehensive evaluation metrics.

To bridge these gaps, we propose \name, a large-scale benchmark specifically designed for comprehensive chart grounding. It comprises 8,068 chart–table–code triples spanning 30 diverse chart types, sourced from over 6,533 real-world examples manually collected by us, along with 1,535 augmented instances derived from existing datasets.
The images and corresponding code are drawn from a selection of popular plotting libraries, reflecting diverse visual styles and implementation patterns found in real-world settings.
Crucially, it introduces two complementary tasks designed to probe different facets of grounding. The first task, Chart-to-Code Generation, requires the model to synthesize executable Python code use indicated plotting library that replicates a given chart. This assesses the model’s understanding of stylistic and structural components—such as axis configuration, data-to-mark mapping, and layout decisions—while implicitly requiring correct data recovery. The second task, Controlled Chart-to-Table Reconstruction, focuses explicitly on data accuracy: given column headers, the model is required to extract the tabular data from the chart image, isolating numerical precision and structural alignment. By providing headers, the task eliminates label ambiguity and allows for a more focused evaluation of data fidelity.
Beyond task design, \name introduces a four-dimensional evaluation framework tailored to assessing and guiding the development of MLLMs for chart grounding and understanding. It provides the first unified diagnostic system that jointly evaluates functional validity (execution pass rate), visual rigor (verification of chart type, color, text, and layout), semantic data fidelity (structured tuple matching), and perceptual consistency (CLIP-based semantic alignment). Rather than relying on limited or isolated metrics, this framework enables comprehensive assessment of both code-level reasoning and data-level understanding, laying a foundation for building multimodal models that integrate computational precision with visual and semantic coherence.

%% file: Section/related.tex
\section{Related Work}

\begin{figure*}[htbp]
\centering
\begin{minipage}[t]{0.25\textwidth}
\centering
\includegraphics[width=\textwidth]{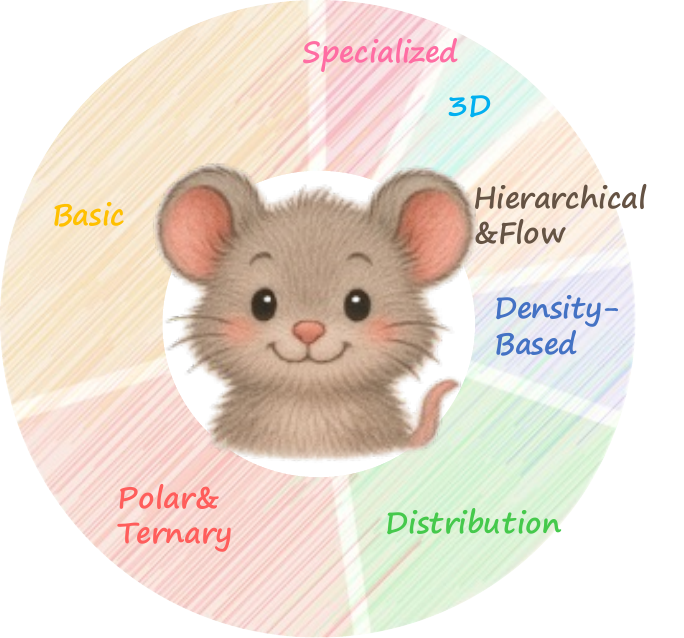}
\caption{Chart Type Distribution in \name.}
\label{fig:mouse}
\end{minipage}
\hfill
\begin{minipage}[t]{0.65\textwidth}
\centering
\includegraphics[width=\textwidth]{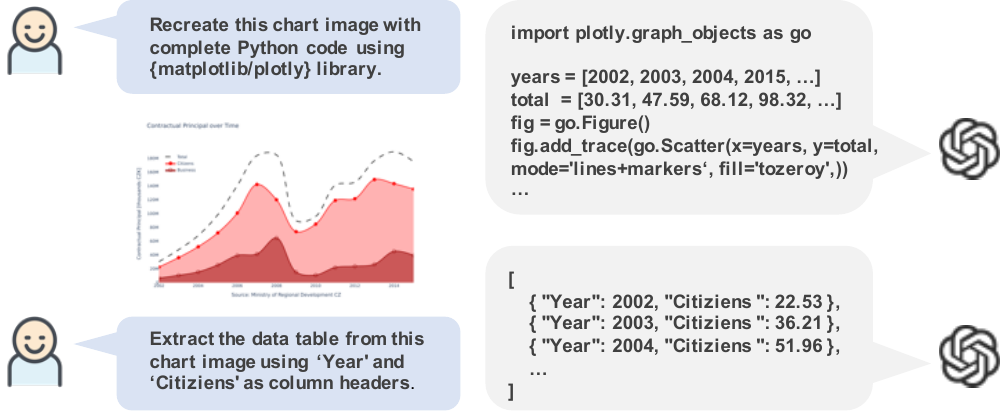}
\caption{Illustration of our chart-to-code generation task and controlled chart-to-table reconstruction task.}
\label{fig:task}
\end{minipage}
\vspace{-2mm}
\end{figure*}

\paragraph{MLLMs.}
Proprietary pioneers, notably Claude-Sonnet-4 \citep{Claude4}, GPT-5 \citep{gpt5}, and Gemini-2.5-Pro \citep{comanici2025gemini}, have unlocked potential for addressing complex realistic applications. Concurrently, the open-source community has released diverse high-quality models. This includes prominent series like QwenVL \citep{yang2025qwen3} and InternVL \citep{zhu2025internvl3}, alongside models such as MiMo-VL \citep{mimo}, GLM-4V \citep{GLM-4V}, and Deepseek-VL \citep{lu2024deepseek}. Together, they significantly advance the field of visual-textual understanding.

\textbf{Chart Benchmark.} 
The growing importance of chart comprehension in multimodal reasoning has driven the development of numerous benchmarks. Existing datasets target distinct capabilities, ranging from summarization \citep{kantharaj2022chart,rahman2023chartsumm} and VQA \citep{masry2022chartqa,methani2020plotqa,zhu2024multichartqa} to code generation \citep{yang2024chartmimic,wu2024plot2code,si2024design2code}. Although ChartBench \citep{xu2023chartbench} and ChartX \citep{xia2024chartx} move beyond isolated tasks by incorporating diverse reasoning challenges, they fail to unify visual understanding with structured semantic outputs. Moreover, existing datasets are limited by their diversity in chart types, fixed rendering libraries, and rigid metrics.

In contrast to existing benchmarks, we introduce ChartAnchor, a benchmark that formalizes chart grounding—a unified task mapping visual inputs to both executable code and tabular data. ChartAnchor bridges bridging chart-to-code generation and controlled table extraction within a single framework. It features 30 diverse chart types and over 8k+ instances rendered via multiple libraries, providing rich image–table–code tuples for robust evaluation.
\vspace{-5pt}

%% file: Section/MyBench.tex
\section{ The ChartAnchor Benchmark}

\begin{figure*}[t]
  \centering
  \begin{minipage}[b]{0.7\textwidth} 
    \centering
    \includegraphics[width=\linewidth]{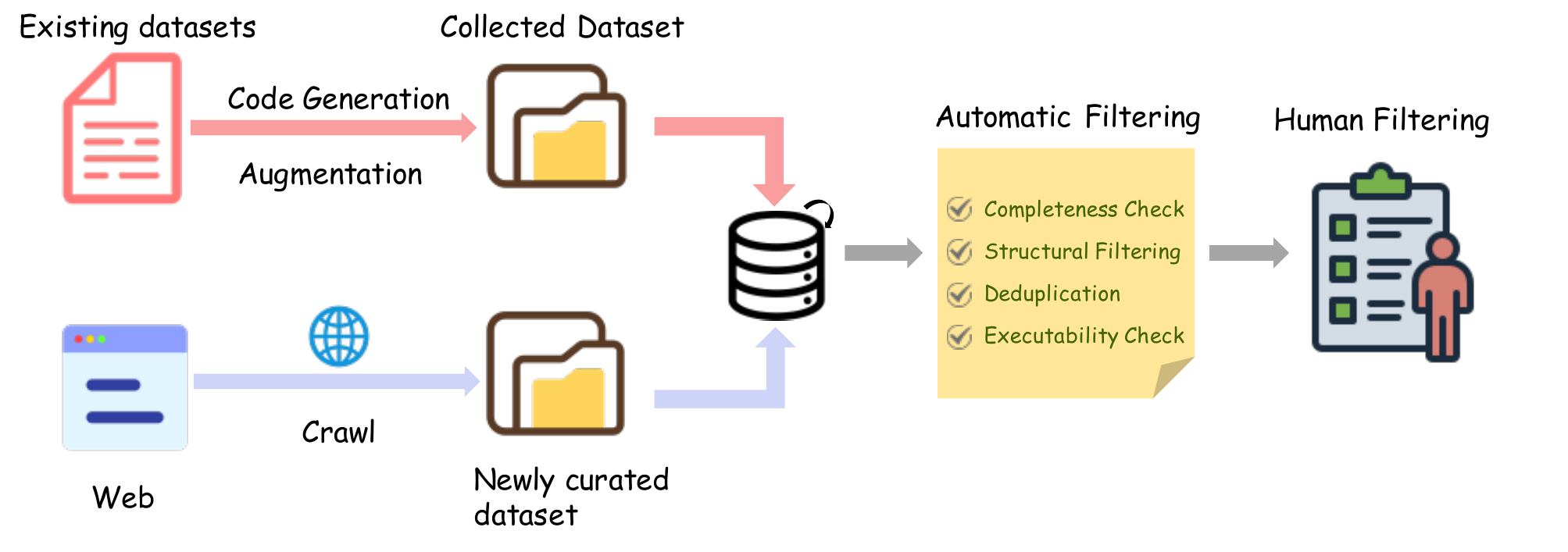}

    \captionof{figure}{Illustration of the Data Collection Pipeline.}
    \label{fig:collection}
  \end{minipage}
  \hspace{0.02\textwidth} 
  \begin{minipage}[b]{0.25\textwidth} 
    \centering
    \footnotesize
    \setlength{\tabcolsep}{4pt}
    \renewcommand{\arraystretch}{0.9}
    \begin{tabular}{l r}
      \toprule
      \textbf{Statistics} & \textbf{Value} \\
      \midrule
      \multicolumn{2}{l}{\textbf{Chart Images}} \\
      \quad Total Charts & 8,068 \\
      \quad Avg. Width & 3,346 \\
      \quad Avg. Height & 2,266 \\
      \quad Aspect Ratio & 1.43 \\
      \quad Brightness Std & 42.35 \\
      \quad Entropy & 1.41 \\
      \midrule
      \multicolumn{2}{l}{\textbf{Python Code}} \\
      \quad Mean Chars & 1,469 \\
      \quad Mean Tokens & 627.67 \\
      \midrule
      \multicolumn{2}{l}{\textbf{Table}} \\
      \quad Mean Rows & 20.35 \\
      \quad Mean Cols & 3.05 \\
      \bottomrule
    \end{tabular}

    \captionof{table}{Dataset statistics. }

    \label{tab:dataset_stats}
  \end{minipage}
\end{figure*}

\begin{table*}[t]
\centering
\renewcommand{\arraystretch}{0.6}
\resizebox{\textwidth}{!}{
\begin{tabular}{ccccccccccc}
\toprule
Type & Scatterpolar & Scatter3d & Line3d & Pie & Barpolar & Mesh3d & Violin & Line & Histogram2d & Bar \\
Num  & 653          & 223       & 62     & 700 & 350      & 103     & 214    & 617  & 174          & 659 \\
\midrule
Type & Box & Scatterternary & Waterfall & Heatmap & Scatter & Cone & Surface & Histogram & Carpet & Treemap \\
Num  & 423 & 142             & 111       & 372     & 675     & 44   & 84      & 482       & 23     & 121 \\
\midrule
Type & Parcoords & Funnelarea & Funnel & Sankey & Candlestick & Contour & Sunburst & Histogram2dcontour & Areachart & Ohlc \\
Num  & 71        & 97         & 175    & 231    & 116         & 246     & 283      & 142                  & 425       & 50 \\
\bottomrule
\end{tabular}
}
\caption{Chart counts across types.}
\label{tab:chart_type}
\vspace{-12pt}
\end{table*}

ChartAnchor evaluates chart grounding via chart-to-code generation and controlled table reconstruction. We construct a diverse corpus of chart–table–code triplets from existing and real-world sources (distribution in Fig.~\ref{fig:mouse}). Our pipeline integrates parametric generation with augmentation, followed by a rigorous hybrid filtering process to ensure high data quality and semantic fidelity.
\vspace{-8pt}

\subsection{Task Definitions}
As illustrated in Fig. \ref{fig:task}, we define two core tasks grounded in our chart–table–code triplets:

\textbf{Chart-to-Code Generation.} This task requires generating executable Python code from a chart image $I$. The model must produce a valid script that accurately reconstructs visual and structural elements (e.g., chart type, data, layout, and style). This assesses capabilities in visual abstraction and symbolic reasoning, demanding output that is both syntactically correct and semantically faithful to the input.

\textbf{Controlled Chart-to-Table Reconstruction.} This task recovers tabular data under constrained header supervision. Given an image $I$ and headers $\mathcal{H} = \{h_1, \dots, h_n\}$, the model generates a table $\mathcal{T} = [r_1, \dots, r_m]$ aligning numerical values to corresponding headers. The inclusion of headers removes label ambiguity, enabling a more targeted and reliable evaluation of data fidelity.

\subsection{Data Collection}

This section presents the data collection pipeline for \name, which comprises 8,068 samples: 1,535 (19\%) augmented from existing datasets and 6,533 (81\%) curated from real-world sources. Detailed procedures are provided in Appendix~\ref{app:coll}.

\subsubsection{\textbf{Data Source}}
As illustrated in Fig~\ref{fig:collection}, we collect a diverse set of chart–table–code triplets from two main sources: 
\begin{itemize}[leftmargin=*]
\vspace{-5pt}
\item \textbf{Existing datasets.} We leverage chart–table pairs from PlotQA \citep{methani2020plotqa}, DVQA \citep{kafle2018dvqa}, FigureQA \citep{kahou2017figureqa}, and Vistext \citep{tang2023vistext}. Given the absence of original code, we synthesize plotting scripts from metadata, applying systematic augmentations to visual and semantic attributes to ensure diversity and mitigate leakage.

\vspace{-5pt}
\item \textbf{Newly curated dataset.} We curate a large collection of chart\-table-code triples from publicly web sources, where charts are created and contributed by real users across a wide range of domains.
\vspace{-3pt}
 \end{itemize}

\subsubsection{\textbf{Code Generation and Augmentation for Existing Datasets.}}
To address the lack of source code in existing chart datasets, we propose a parametric pipeline that translates chart metadata into executable plotting scripts.  The pipeline consists of three main stages: semantic mapping of chart elements to primitives in visualization libraries, parameterization of visual attributes such as colors, fonts, and layout, and systematic augmentation through controlled style variations. These augmentations include changes to color schemes, marker types, axis scales, legend positions, font settings, and label orientations.

\subsubsection{\textbf{Filtering.}} Our filtering process consists the following two stages, more details are in Appendix~\ref{Filtering}.

\textbf{Automated Filtering.} To ensure data fidelity, we implement a four-stage pipeline: \textit{(a) Completeness}, pruning incomplete data triples; \textit{(b) Structural Filtering}, excluding unstructured formats such as rasterized tables or maps; \textit{(c) Deduplication}, removing redundancies via exact matching of compact statistical signatures; and \textit{(d) Executability}, discarding scripts that trigger runtime errors or rendering failures.

\textbf{Human Filtering.} We enforced rigorous quality control through a double-blind review process involving experts proficient in deep learning and data visualization. Samples were grouped by chart type and filtered within each group to ensure consistency across families. Independent reviewers evaluated the triples based on a standardized rubric for semantic accuracy, visual clarity, and stylistic diversity, with discrepancies resolved by a third adjudicator to guarantee high annotation reliability.

%% file: Section/dataset.tex
\begin{figure}[t]
  \centering

  \begin{subfigure}[t]{0.48\linewidth}
    \includegraphics[width=\linewidth,height=2cm]{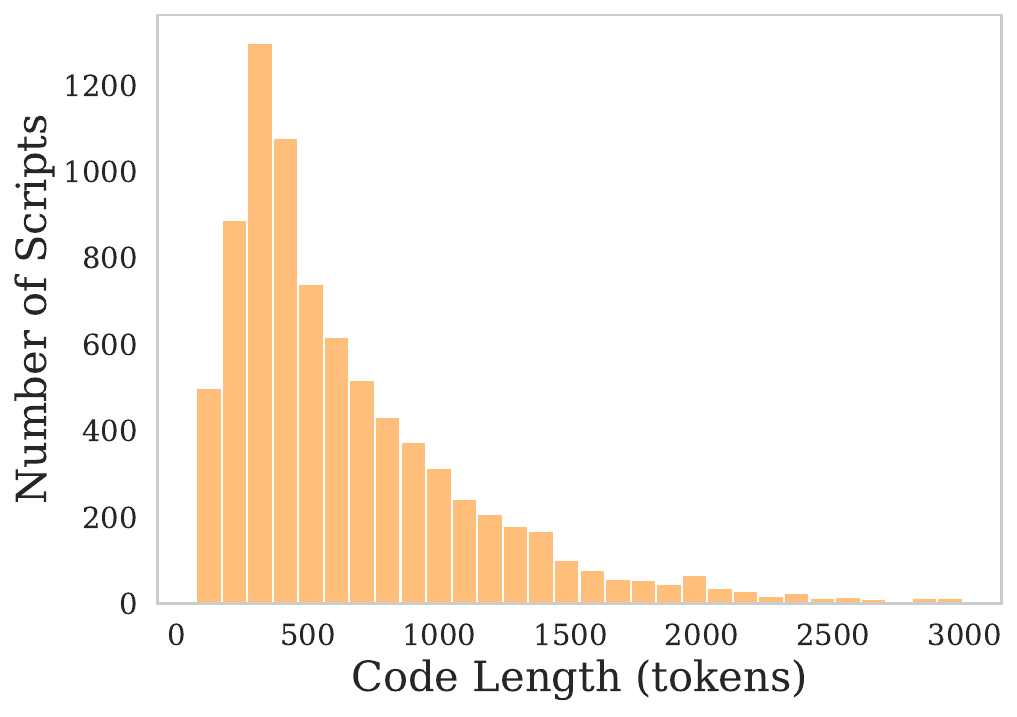}
    \caption{Code Length Distribution.}
  \end{subfigure}
  \hfill
  \begin{subfigure}[t]{0.48\linewidth}
    \includegraphics[width=\linewidth,height=2cm]{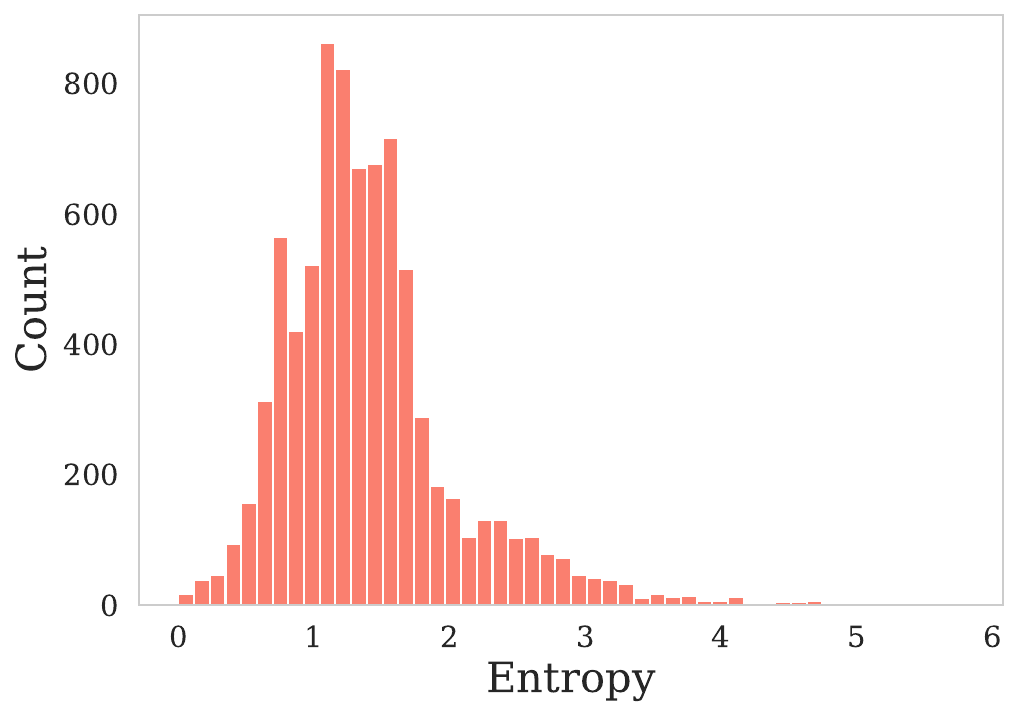}
    \caption{Image Entropy Distribution.}
  \end{subfigure}

  \vspace{0.5em}

  \begin{subfigure}[t]{0.48\linewidth}
    \includegraphics[width=\linewidth,height=2cm]{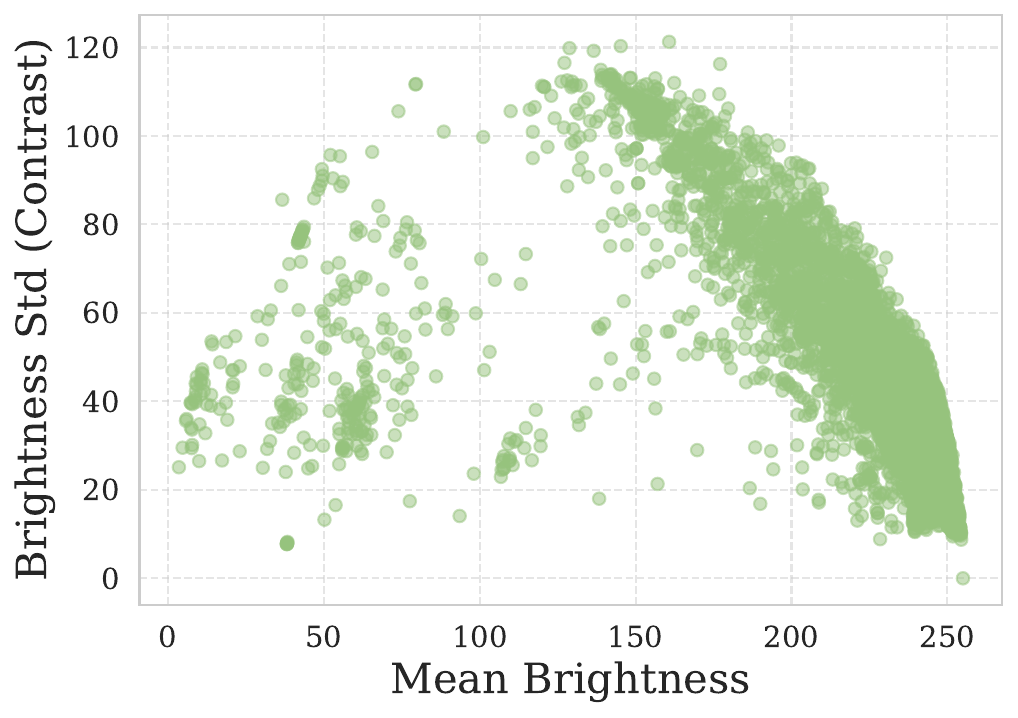}
    \caption{Brightness vs. Contrast.}
  \end{subfigure}
  \hfill
  \begin{subfigure}[t]{0.48\linewidth}
    \includegraphics[width=\linewidth,height=2cm]{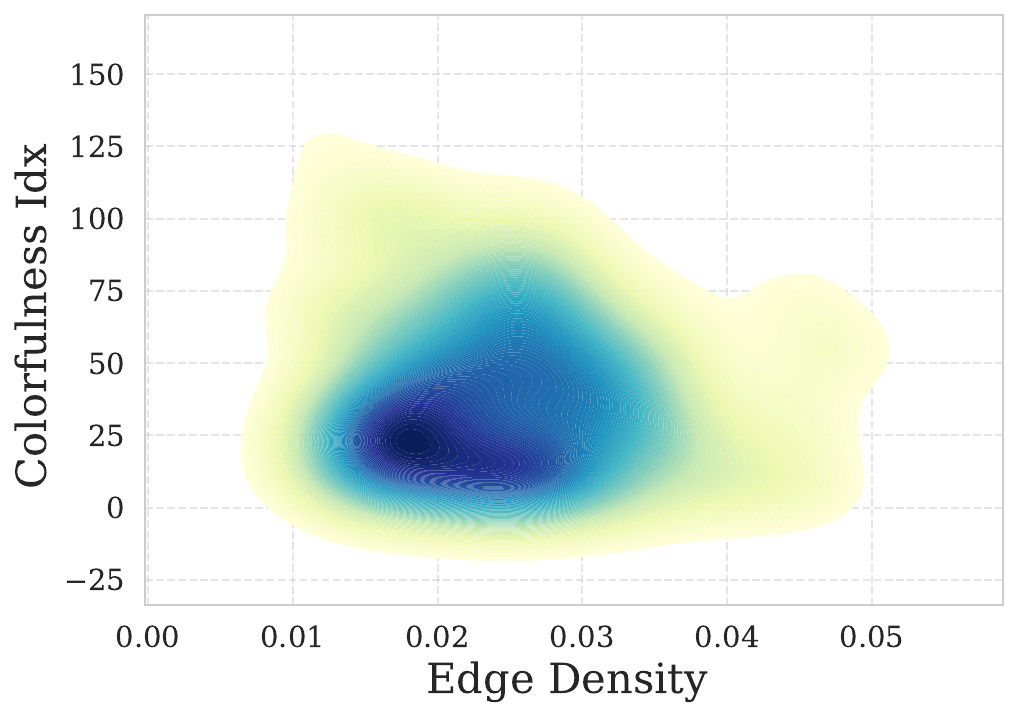}
    \caption{Visual / Color Complexity.}
  \end{subfigure}

  \caption{Chart-level visual statistics.}
  \vspace{-18pt}
  \label{fig:chart_sta}
\end{figure}

\vspace{-10pt}

\subsection{Dataset Analysis}

\input{table/compare_bench}

Tab.~\ref{tab:dataset_stats} and Fig.~\ref{fig:chart_sta} summarize the structural and visual diversity of \name: 8,068 charts with a range of resolutions, averaging at 3,346×2,266px, brightness variability ($\sigma$ = 42.35), and entropy ($\mu$ = 1.41). Code length spans from concise to complex scripts (Fig.~\ref{fig:chart_sta}a), while visual metrics show moderate entropy (Fig.~\ref{fig:chart_sta}b), diverse brightness/contrast (Fig.~\ref{fig:chart_sta}c), and broad color-structural distributions (Fig.~\ref{fig:chart_sta}d), reflecting its  real-world relevance by a balance between informativeness and realism. Token counts are computed using the GPT-4 tokenizer for tokenization.

As detailed in Tab.~\ref{tab:chart_type}, ChartAnchor encompasses \textbf{30} visualization families. Beyond canonical 2D plots, it covers diverse categories including advanced 3D charts, hierarchical diagrams, polar variants, and financial charts. The distribution mimics realistic usage patterns, balancing frequent types (e.g., \textit{bar}, \textit{line}) with long-tail complex charts. By integrating curated benchmarks with real-world code, ChartAnchor incorporates diverse plotting libraries (e.g., Matplotlib, Plotly) and styling idioms, establishing a robust benchmark for MLLM chart grounding. Detailed statistics and visual examples are provided in Appendices~\ref{app:dataset} and \ref{app:chart_examples}.

\subsection{Comparison with Other Benchmarks}
Table~\ref{tab:comdataset} presents a comprehensive comparison between our proposed benchmark and a wide range of existing datasets related to chart understanding and code generation.

Early benchmarks (e.g., ChartQA, PlotQA, Chart-to-Text) focus on NL tasks like QA and summarization, without structured protocols to evaluate bidirectional alignment between chart visuals and their underlying data or specifications. More recent multi-task datasets such as ChartArxiv and ChartBench broaden the scope, yet still omit code-based grounding, limiting their ability to assess executable understanding of visual semantics.

Code generation benchmarks like HumanEval and MBPP target text-to-code synthesis and lack visual input, making them unsuitable for evaluating multimodal alignment. Conversely, multimodal datasets like Plot2Code and ChartMimic introduce image-to-code tasks but fail to explicitly assess numerical fidelity or data reconstruction.

In contrast, \name is explicitly designed to evaluate comprehensive chart grounding. It uniquely integrates the following features:
(1) High diversity with 30 chart types and multiple plotting libraries to support varied symbolic mappings;
(2) Fidelity checks for both stylistic rendering and precise data recovery;
(3) Support for unannotated chart images to facilitate unsupervised visual reasoning; 
(4) Chart–table–code supervision for rigorous evaluation of symbolic, visual, and structural alignment.

%% file: table/compare_bench.tex
\begin{table*}[htbp]
\centering
\renewcommand{\arraystretch}{0.65}
\resizebox{0.8\textwidth}{!}{%
\begin{tabular}{l|cccccccccc}
\toprule
\multicolumn{1}{c|}{Dataset}  & Types & \makecell{Test\\ Set} & \makecell{Input \\Format} &\makecell{ Output\\ Format} & \makecell{Multi-task\\ Eval }& Visual & \makecell{Open\\
Source} & \makecell{Plot\\ Library} & \makecell{Data\\ Eval} & \makecell{Full\\ Ground} \\
\rowcolor[rgb]{ .851,  .882,  .949} \multicolumn{11}{c}{Chart Benchmarks} \\
ChartQA\citep{masry2022chartqa}  & 3     & 10k   & I+NL  & NL    & \xmark    & \xmark     & \cmark     & - & - & \xmark \\
PlotQA\citep{methani2020plotqa}   & 3     & 34k   & I+NL  & NL    & \xmark     & \xmark     & \cmark      & - & - & \xmark \\
Chart-to-text\citep{kantharaj2022chart}  & 6     & 44k   & I+NL  & NL    & \xmark      & \xmark      & \cmark      & - & - & \xmark\\
OpenCQA\citep{kantharaj2022opencqa}  & 5     & 1.2k  & I+NL  & NL    & \xmark     & \xmark     & \cmark      & - & - & \xmark \\
ChartSumm\citep{rahman2023chartsumm}  & 3     & 84k   & I+NL  & NL    & \xmark      & \xmark     & \cmark      &  -     & -  & \xmark\\
Charxiv\citep{wang2024charxiv}   & 18    & 1.32k   & I+NL  & NL    &    \cmark    & \cmark     &  \cmark     &  -     & -  & \xmark\\
MMC\citep{liu2023mmc}   & 6     & 2k    & I+NL  & NL    & \cmark      & \xmark      & \xmark     & - &  - & \xmark\\
ChartX\citep{xia2024chartx}  & 18    & 6k    & I+NL  & NL    & \cmark      & \xmark      & \xmark      & - & -  & \xmark\\
ChartBench\citep{xu2023chartbench}     & 9     & 2.1k  & I+NL  &   NL    & \cmark     & \cmark     & \cmark    &  -     &  -& \xmark\\
\rowcolor[rgb]{ .851,  .882,  .949} \multicolumn{11}{c}{Code Generation Benchmarks} \\
HumanEval\citep{chen2021evaluating}   & -     & 164   & Code  & Code  &   \cmark    & -    &    \cmark   &    -   & - & -\\
MBPP\citep{austin2021program}    & -     & 500   & NL+Code & Code  &  \cmark      & -     &   \cmark    &    -   & - & -\\
MMCode\citep{li2024mmcode}  & -     & 263   & I+NL  & Code  &    \cmark    & -     &    \cmark   &    -   & - & -\\
MatPLotBench\citep{yang2024matplotagent}    & 13    & 100   & NL    & Code  &   \cmark     & \cmark    &     \cmark  &  Mixture     & \xmark & \xmark\\
PLot2Code\citep{wu2024plot2code}  & 15    & 132   & I+NL  & Code  & \cmark     & \cmark     & \cmark    &    Singal   & \xmark & \xmark\\
ChartMimic\citep{yang2024chartmimic}   & 22    & 4.8k  & I+NL  & Code  & \cmark    & \cmark     & \cmark     &   Singal    & \xmark & \xmark\\
Design2Code\citep{si2024design2code}  & -  & 484   & I+NL  & Code  & \cmark     & \cmark     &  \cmark     &   Singal    & \xmark & \xmark\\
\midrule
\textbf{\name}  & 30    & 8.1k  & I+NL  & Code+NL & \cmark    & \cmark     & \cmark     & Mixture & \cmark & \cmark \\
\bottomrule
\end{tabular}
}
\caption{Comparison of benchmarks for chart understanding and code generation.  
“I” = image, “NL” = natural language. 
“Visual” assesses unannotated chart reasoning. 
“Plot Library” denotes the use of multiple plotting libraries in the chart-to-code task, while “Data Eval” indicates whether data fidelity is evaluated within the chart-to-code task.
}
\label{tab:comdataset}
\vspace{-10pt}
\end{table*}

%% file: Section/metrics.tex
\section{Evaluation Metrics}
\vspace{-5pt}

While existing chart-to-code benchmarks typically rely on basic metrics such as execution success or superficial visual similarity, they fail to capture semantic fidelity, particularly the correctness of underlying chart data. To address this, we introduce a multi-level evaluation framework that jointly examines the functional correctness, visual integrity, and data faithfulness of model outputs. This enables more precise assessment of chart grounding performance in real-world scenarios.

\textbf{\textit{Functional Validity.}} Pass Rate measures the proportion of model outputs that execute or parse without errors—i.e., valid chart-rendering code or well-formed tables—indicating baseline reliability.

\textbf{\textit{Visual Structure Consistency.}}
To go beyond functional success, we perform fine-grained evaluation of visual structure through four key aspects extracted directly from the rendered chart objects:

\textit{1) Textual Components Match.} We extract textual components (titles, axis labels, legends, annotations) and compare against reference charts to verify semantic and positional consistency.

\textit{2) Color Fidelity.} We quantify perceptual color differences using the CIEDE2000~\citep{sharma2005ciede2000} metric in the CIE Lab color space, which aligns with human vision sensitivity. This evaluation covers both static elements and dynamic color mappings. For multi-color comparisons, we apply the Hungarian algorithm to optimally pair generated and reference colors, minimizing the total perceptual deviation ($\Delta E_{00}$) across matched pairs. This ensures semantic alignment of color associations while preserving numerical fidelity.

\textit{3) Chart Type Identification.} Chart types indicate the structural intent of a visualization. We determine the type distribution for both generated and reference charts by identifying the type of each visual element, and measure accuracy through distributional comparison.

\textit{4) Layout Alignment.} We evaluate the presence, number, size, and arrangement of subplots to ensure structural correctness in multi-panel charts.

\paragraph{\textbf{\textit{Semantic Data Fidelity.}}}

\input{table/main_code}
To evaluate whether models can faithfully recover the underlying chart data, we introduce \textit{data-level fidelity metric}for both \textit{chart-to-code} and \textit{controlled chart-to-table} tasks. 

In the \textit{chart-to-code}, we first parse the generated figure objects and dispatch them to type-specific extractors \( \text{Extract}_T \), where each \( T \in \mathcal{T} \) corresponds to a supported chart type. Each extractor retrieves the relevant data fields—such as \( (x, y) \) for line or \( (x, \text{low}, \text{high}, \text{open}, \text{close}) \) for candlestick charts—and formats each data point as a normalized tuple \( \tau_i = (n, x_i, y_i, \ldots) \), where \( n \) is the trace name and the remaining elements are type-specific field values. The final structured form is a set \( \mathcal{L}_T = \{ \tau_1, \tau_2, \ldots, \tau_n \} \). In the \textit{controlled chart-to-table} task, each table row is similarly treated as a tuple \( \tau_i = (t_1, t_2, \ldots, t_m) \), resulting in a comparable structured set \( \mathcal{L}_T \), allowing both tasks to be evaluated under a unified tuple-based framework.

To assess prediction quality, we adopt a matching scheme inspired by the Structuring Chart-oriented Representation Metric (SCRM)~\citep{xia2023structchart}. For each predicted tuple \( p \) and ground-truth tuple \( q \), we compare corresponding fields \( p_i \) and \( q_i \), where \( i = 0, 1, \dots, n{-}1 \) and \( n \) is the tuple length. String fields are evaluated using edit distance \( J(p_i, q_i) \), and numerical fields using relative error \( e(p_i, q_i) \). A tuple is considered correct only if \textbf{all fields satisfy their respective tolerance}, with three tolerance levels: \textit{strict} (\( J \leq 0 \) or \( e \leq 0 \)), \textit{slight} (\( J \leq 3 \) or \( e \leq 0.05 \)), and \textit{high} (\( J \leq 5 \) or \( e \leq 0.10 \)).

We evaluate structural reconstruction using Precision ($P=n_m/n_p$), Recall ($R=n_m/n_{gt}$), F1 score ($2PR/(P+R)$), and IoU ($n_m/(n_p+n_{gt}-n_m)$). Here, $n_m$, $n_p$, and $n_{gt}$ denote the counts of matched, predicted, and ground-truth tuples, respectively. These metrics provide a principled assessment of semantic alignment and numerical fidelity.

\textbf{\textit{Perceptual Similarity.}}
To bridge the gap between syntax and human perception, we employ CLIPScore to measure semantic consistency via embedding alignment. This complements structured metrics by quantifying conceptual fidelity grounded in visual cognition.

%% file: table/main_code.tex
\renewcommand{\arraystretch}{1}
\begin{table*}[htbp]
  \centering
  \scalebox{0.55}{ 
  \begin{tabular}{l|c|ccccc|c|c|c|c|c|c}
    \toprule
    \multicolumn{1}{c|}{\multirow{3}[6]{*}{\textbf{Model}}} & \multirow{3}[6]{*}{\textbf{\makecell{Pass \\Rate}}} & \multicolumn{8}{c|}{\textbf{Visual Structure Consistency}} & \multirow{3}[6]{*}{\textbf{\makecell{Data \\Fidelity}}} & \multirow{3}[6]{*}{\textbf{\makecell{Clip\\ Score}}} & \multirow{3}[6]{*}{\textbf{\makecell{Over\\all}}} \\
    \cmidrule{3-10}
    & & \multicolumn{5}{c|}{\textbf{Text}} & \multicolumn{1}{c|}{\multirow{2}[4]{*}{\textbf{Color}}} & \multirow{2}[4]{*}{\textbf{Type}} & \multirow{2}[4]{*}{\textbf{Layout}} & & & \\
    \cmidrule{3-7}
    & & Legend & Title & \makecell{Axis\\ Label} & Annos & Avg. & \multicolumn{1}{c|}{} & & & & & \\
    \midrule
    \rowcolor[rgb]{ .851,  .882,  .949}\multicolumn{13}{c}{\textbf{Proprietary Multimodal Large Language Models}} \\
    GPT4o \citep{openai2024} & 91.88 & 63.53 & 72.83 & 67.20 & 76.36 & 69.98 & 34.64 & 70.06 & 80.73 & 35.85 &74.25 & 58.25 \\
    GPT-5 \citep{gpt5}  & 91.93 & 66.11 & 86.18 & 83.63 & 87.17 & 80.77  & 51.21 & 82.21 & 92.3  & 46.55 & 84.52 & 72.93  \\
    Claude-3-7-Sonnet \citep{claude-sonnet} & 78.60 & 54.97 & 69.11 & 67.19 & 70.57 & 65.46 & 40.79 & 59.89 & 76.53 & 30.30 & 65.82 & 54.59 \\
    Claude-4-5-Sonnet \citep{claude-sonnet} & 83.89 & 60.46 & 79.15 & 74.86 & 80.65 & 73.78  & 49.44 & 65.9  & 85.05 & 36.7  & 78.93 & 64.97  \\
    Gemini-3-Pro \citep{comanici2025gemini} & 83.86 & 60.31 & 79.84 & 76.97 & 83.25 & 75.09  & 48.32 & 71.98 & 87.38 & 41.58 & 75.33 & 66.61  \\
    \midrule
    \rowcolor[rgb]{ .851,  .882,  .949}\multicolumn{13}{c}{\textbf{Open-Source Multimodal Large Language Models}} \\
    InternVL3-2B \citep{zhu2025internvl3}& 28.97 & 18.75 & 17.41 & 17.63 & 20.88 & 18.67 & 8.98 & 16.01 & 21.55 & 13.56 & 17.93 & 15.75 \\
    Qwen2.5-VL-3B-Instruct \citep{yang2024qwen2}& 48.07 & 28.24 & 31.80 & 26.86 & 39.44 & 31.58 & 12.29 & 27.01 & 40.19 & 16.71 & 38.63 & 25.55 \\
    Gemma-3-4B-it \citep{team2025gemma} & 66.20 & 42.29 & 38.72 & 36.41 & 57.50 & 43.73 & 18.74 & 33.53 & 60.46 & 24.21 &47.48 & 36.13\\
    DeepSeek-VL2-27B-A4B \citep{wu2024deepseekvl2mixtureofexpertsvisionlanguagemodels} & 7.98  & 5.93  & 5.53  & 6.38  & 5.41  & 5.81  & 2.7   & 6.08  & 5.54  & 3.25  & 4.51  & 4.65  \\
    DeepSeek-VL-7B \citep{lu2024deepseek} & 32.05 & 17.59 & 20.41 & 17.62 & 23.19 & 19.70 & 9.56  & 17.37 & 24.83 &7.74& 20.35 & 15.84\\
    LLaVA-v1.6-Mistral-7B \citep{li2024llava} & 15.32 & 8.74 & 8.37 & 4.96 & 4.91 & 6.75 & 6.03 & 7.79 & 5.05 & 4.76 & 4.39 & 6.07 \\
    Qwen2.5-VL-7B-Instruct \citep{yang2024qwen2}& 67.59 & 41.75 & 41.46 & 43.60 & 61.90 & 47.18 & 18.45 & 41.83 & 64.30 & 27.85 &51.85 & 39.92\\
    MiMo-VL-7B-SFT \citep{coreteam2025mimovltechnicalreport} & 35.56 & 24.95 & 28.95 & 28.52 & 24.6  & 26.76  & 15.06 & 26.91 & 25.64 & 18.53 & 26.91 & 23.30  \\
    MiMo-VL-7B-RL \citep{coreteam2025mimovltechnicalreport}  &  41.51 & 30.19 & 34.79 & 34.01 & 30.26 & 32.31  & 18.76 & 32.77 & 31.96 & 22.92 & 29.9  & 28.10  \\
    MiniCPM-V-2.6-8B \citep{yao2024minicpm}& 26.87 & 16.76 & 16.88 & 14.21 & 16.18 & 16.01 & 7.91 & 13.77 & 16.92 & 12.95 & 14.35 & 13.51\\
    Qwen3-VL-8B-Instruct \citep{bai2025qwen3vltechnicalreport}  &62.16 & 43.11 & 48.8  & 48.82 & 52.92 & 48.41  & 30.97 & 48.45 & 55.56 & 27.27 & 54.96 & 44.27  \\
    InternVL3-9B \citep{zhu2025internvl3}& 69.19 & 44.75 & 47.48 & 45.06 & 59.57 & 49.20 & 22.06 & 40.72 & 66.20 & 29.20 & 55.14 & 41.47\\
    GLM-4V-9B \citep{glm2024chatglm}& 46.04 & 26.55 & 27.86 & 24.55 & 43.78 & 30.68 & 12.02 & 23.32 & 45.69 & 15.51 & 36.66 & 25.44\\
    CogVLM2-Llama3-Chat-19B \citep{hong2024cogvlm2}& 7.82 & 7.12 & 6.03 & 4.79 & 1.25 & 4.80 & 3.77 & 5.61 & 1.21 & 3.52 & 6.15 & 3.78\\
    InternVL3-14B \citep{zhu2025internvl3}& 80.60 & 52.80 & 55.69 & 54.90 & 75.12 & 59.63 & 26.13 & 53.08 & 80.63 & 30.59 &64.7 & 50.01\\
    Qwen2.5-VL-32B-Instruct \citep{yang2024qwen2}& 67.59 & 42.71 & 48.69 & 48.91 & 73.55 & 53.47 & 23.56 & 48.29 & 82.08 & 21.14 &66.58 & 45.70\\
     Qwen3-VL-32B-Instruct \citep{bai2025qwen3vltechnicalreport} & 73.56 & 51.15 & 46.67 & 50.64 & 66.39 & 53.71  & 38.51 & 58.46 & 70.14 & 33.86 & 63.87 & 53.09  \\
    GLM-4.5V-106B \citep{5team2025glm45agenticreasoningcoding} & 66.77 & 49.18 & 61.44 & 58.12 & 55.49 & 56.06  & 32.55 & 55.54 & 57.97 & 34.03 & 52.58 & 48.12  \\
     Qwen3-VL-235B-A22B-Instruct \citep{bai2025qwen3vltechnicalreport} & 79.62 & 57.23 & 64.54 & 62.65 & 73.67 & 64.52  & 37.2  & 62.88 & 76.81 & 38.07 & 68.62 & 58.02  \\

    \bottomrule
   
  \end{tabular}
  }
  \caption{Comprehensive performance comparison of proprietary and open-source MLLMs on the Chart-to-Code generation task. For data fidelity, we report IoU score under a slight tolerance stage. As our collected matplotlib images do not include annotations by design, they are excluded from the annotation metric evaluation. }
  \label{tab:code_task}
  \vspace{-15pt}
\end{table*}

%% file: Section/exp.tex
\section{Evaluation}

\vspace{-3pt}

This section benchmarks diverse proprietary and open-source MLLMs using \name, offering a holistic performance comparison.

\subsection{Evaluated Models and Implementation}

We benchmark 24 MLLMs across two categories. For the open-source models, total parameters ranging from 2B to 235B, including: Qwen2.5-VL series, Qwen3-VL series, InternVL3 series, DeepSeek-VL series, GLM series, and MiMo-VL-7B variants, alongside Gemma-3-4B-it, LLaVA-v1.6-Mistral-7B, MiniCPM-V-2.6-8B, and CogVLM2-Llama3-Chat-19B. For proprietary models, we include GPT-4o, GPT-5, Claude-3.7-Sonnet, Claude-4.5-Sonnet, and Gemini-3-Pro. For reproducibility, detailed experimental settings and the full list of prompts are relegated to Appendix~\ref{app:imple}.

\input{table/main_table}
\begin{figure*}
\vspace{-8pt}
\centering
\includegraphics[width=0.9\textwidth]{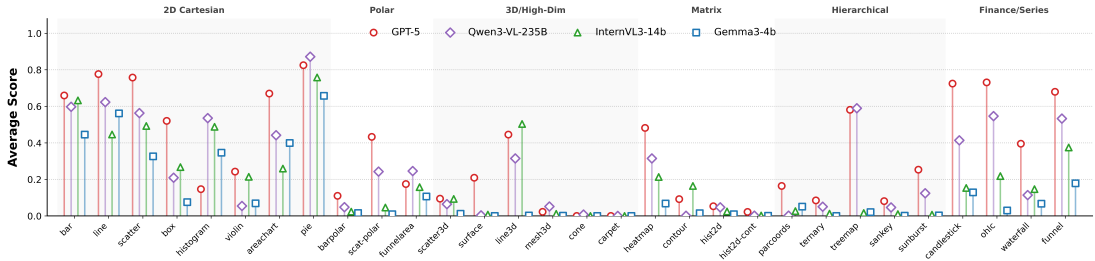}

\caption{Comparison of Data Attribute Prediction Across Chart Types for Different Models} 
\label{fig:data-attribute-accuracy}
\end{figure*}
\subsection{Key Insights from Results}
Tab. \ref{tab:code_task} and \ref{tab:tabletask} present the results for the Chart-to-Code and controlled Chart-to-Table reconstruction tasks, respectively. Key insights are discussed below, with further details in Appendix~\ref{app:imple}.

\textbf{\textit{1) GPT-5 establishes a new state-of-the-art across both tasks.}} GPT-5 achieves the highest overall score in chart-to-code generation (72.93) and strong F1 scores in chart-to-table reconstruction (F1-High: 55.36). Its superior visual reasoning capabilities are evidenced by exceptional pass rates and leading scores in both Visual Structure Consistency and Data Fidelity. While open-source models exhibit recent advancements, a distinct performance gap remains, particularly regarding generation accuracy and structural robustness compared to the proprietary baseline.

\textbf{\textit{2) The "Thinking" Paradox: Reasoning enhances structural alignment but risks amplifying perceptual errors.}} Models equipped with "thinking" modes (e.g., GPT-5, Gemini 3) excel in layout alignment and chart type recognition by logically deducing structural composition. However, this capabilities does not guarantee improvements in data or color fidelity; in fact, reasoning chains may inadvertently amplify hallucinations by logically rationalizing initial perception errors. Ultimately, "thinking" upgrades logical planning rather than low-level visual acuity.

\textbf{\textit{3) On-Policy RL catalyzes structural reasoning and code synthesis.}} Relative to MiMo-VL-7B-SFT, MiMo-VL-7B-RL achieves a substantial qualitative leap in Chart-to-Code generation, elevating the pass rate from 35.56 to 41.51 and overall score from 23.30 to 28.10. Driven by the Mixed On-policy RL (MORL) framework, the model transcends the surface-level mimicry characteristic of SFT to internalize visual logical hierarchies. This advancement is evidenced by superior Visual Structure Consistency—where RL effectively aligns COT reasoning with the rigorous global topological constraints of code syntax.

\textbf{\textit{4) Chart-to-code generation reveals consistent weaknesses in color replication and data fidelity.}}
Even top models (e.g., GPT-5) achieve lower scores in Color and Data Fidelity compared to structural elements like Title or Layout, indicating that fine-grained visual detail understanding remains a major bottleneck in generative chart reasoning.

\vspace{-6pt}

\subsection{Type Analysis}

\noindent

Figure~\ref{fig:data-attribute-accuracy} shows a comparative analysis of data attribute prediction accuracy across chart types in the Chart-to-Code task. Overall, GPT-5 establishes a \textit{performance ceiling}, particularly in 2D Cartesian and Finance/Series categories, while Qwen3-VL remains a competitive baseline, often matching GPT-5 in fundamental types like \texttt{pie} and \texttt{histogram}.

Crucially, we observe a substantial \textbf{Geometric Reasoning Gap}. Performance for all models degrades sharply in Polar and 3D categories. While models handle rectilinear layouts effectively, performance collapses on their radial or multi-dimensional counterparts. This indicates that current VLMs rely heavily on visual pattern recognition rather than intrinsic geometric reasoning. They struggle to map pixel-space information to non-orthogonal coordinate systems or to resolve spatial occlusions in 3D projections, highlighting a lack of spatial inductive bias for complex topologies.

Furthermore, all models, including GPT-5, exhibit a \textbf{Struggle with Statistical Abstraction}. They underperform on Matrix and abstract statistical charts (\textit{e.g.}, \texttt{box}, \texttt{violin}, \texttt{heatmap}). Unlike scatter plots that map raw data points directly, these charts require interpreting aggregated statistical distributions or dense grid encodings. This implies that chart-to-code capabilities are currently constrained by the models' limited ability to deconstruct high information density and interpret abstract visual summaries.

\subsection{Alignment with Human Evaluation}

\vspace{-8pt}
\begin{table}[h]
\centering
\small
\resizebox{\linewidth}{!}{
\begin{tabular}{lcc}
\toprule
\textbf{Metric} & \textbf{Acc (\%)} & \textbf{Kendall's $\tau$} \\
\midrule
Visual Structure Consistency & 88.0 & 0.76 \\
Perceptual Quality & 84.0 & 0.68 \\
Semantic Data Fidelity & 85.0 & 0.71 \\
\bottomrule
\end{tabular}
}
\caption{Human alignment analysis.}
\label{tab:human_alignment}
\vspace{-8pt}
\end{table}

To verify whether our evaluation metrics reflect human preferences, we conduct a pairwise comparison study. We sample 300 chart pairs generated by different models from the same ground-truth input and ask domain experts to indicate their preference along \textit{Visual Structure Consistency}, \textit{Semantic Data Fidelity}, and \textit{Perceptual Quality}. As shown in Tab.~\ref{tab:human_alignment}, both Consistency Accuracy and Kendall's $\tau$ demonstrate strong agreement between metric-induced rankings and human judgments. Experimental setup and annotation details are provided in Appendix~\ref{app:human_alignment}.

%% file: table/main_table.tex
\renewcommand{\arraystretch}{1}
\begin{table*}[htbp]
  \centering
  \scalebox{0.55}{
    \begin{tabular}{l|c|ccc|ccc|ccc}
    \toprule
    \multicolumn{1}{c|}{\textbf{Model}} & \textbf{Pass Rate} & \textbf{P} \textcolor{softred}{\rule{0.5em}{0.5em}} & \textbf{R} \textcolor{softred}{\rule{0.5em}{0.5em}} & \textbf{F1} \textcolor{softred}{\rule{0.5em}{0.5em}} & \textbf{P} \textcolor{softyellow}{\rule{0.5em}{0.5em}} & \textbf{R} \textcolor{softyellow}{\rule{0.5em}{0.5em}} & \textbf{F1} \textcolor{softyellow}{\rule{0.5em}{0.5em}} & \textbf{P} \textcolor{softgreen}{\rule{0.5em}{0.5em}} & \textbf{R} \textcolor{softgreen}{\rule{0.5em}{0.5em}} & \textbf{F1} \textcolor{softgreen}{\rule{0.5em}{0.5em}} \\
    \midrule
    \rowcolor[rgb]{.851, .882, .949} \multicolumn{11}{c}{\textbf{Proprietary Multimodal Large Language Models}} \\
    \midrule
    GPT-4o \citep{openai2024} & 97.12 & 10.74 & 8.44  & 8.65  & 37.71 & 23.30 & 24.69 & 60.42 & 35.44 & 37.72 \\
    GPT-5 \citep{gpt5} & 99.94 & 19.62 & 17.88 & 17.64 & 61.41 & 44.63 & 45.89 & 77.48 & 53.79 & 55.36 \\
    Claude-3-7-Sonnet \citep{claude-sonnet} & 94.85 & 12.75 & 11.02 & 10.86 & 44.64 & 29.43 & 30.52 & 58.12 & 37.11 & 38.62 \\
    Claude-4-5-Sonnet \citep{claude-sonnet} & 98.66 & 14.89 & 14.46 & 13.65 & 43.94 & 40.22 & 38.29 & 54.87 & 49.6  & 47.11 \\
    Gemini-3-Pro \citep{comanici2025gemini} & 95.19 & 19.96 & 18.03 & 18.15 & 57.78 & 45.18 & 46.48 & 70.61 & 52.93 & 54.72 \\

    \midrule
    \rowcolor[rgb]{.851, .882, .949} \multicolumn{11}{c}{\textbf{Open-Source Multimodal Large Language Models}} \\
    \midrule
    InternVL3-2B \citep{zhu2025internvl3} & 55.16 & 5.84 & 3.86 & 4.08 & 22.18 & 10.88 & 11.98 & 51.04 & 18.29 & 21.19 \\
    Qwen2.5-VL-3B-Instruct \citep{yang2024qwen2} & 92.18 & 9.11 & 6.13 & 6.51 & 36.70 & 17.28 & 19.12 & 61.01 & 27.43 & 30.73 \\
    Gemma-3-4B-it \citep{team2025gemma} & 96.23 & 7.35 & 5.38 & 5.52 & 32.05 & 18.33 & 19.72 & 48.89 & 26.49 & 28.60 \\
    DeepSeek-VL2-27B-A4B \citep{wu2024deepseekvl2mixtureofexpertsvisionlanguagemodels}
    & 75.18  & 3.81  & 3.03   & 3.08  & 17.33  & 12.15  & 12.76  & 24.83  & 15.65  & 16.76 \\
    DeepSeek-VL-7B \citep{lu2024deepseek}& 98.34 & 1.72 & 0.36 & 0.49 & 5.47 & 0.90 & 1.33 & 26.95 & 4.79 & 7.02 \\
    LLaVA-v1.6-Mistral-7B \citep{li2024llava}& 76.11 & 4.57 & 1.66 & 2.02 & 22.22 & 6.33 & 7.72 & 44.38 & 12.15 & 14.89 \\
    Qwen2.5-VL-7B-Instruct \citep{yang2024qwen2} & 95.05 & 12.17 & 8.18 & 8.67 & 48.45 & 23.75 & 26.35 & 69.58 & 30.96 & 34.90 \\
    
    MiMo-VL-7B-SFT \citep{coreteam2025mimovltechnicalreport} & 97.5  & 12.38 & 9.71  & 9.91  & 47.48 & 28.72 & 30.5  & 70.72 & 38.2  & 41.09 \\
    MiMo-VL-7B-RL \citep{coreteam2025mimovltechnicalreport} & 97.36 & 12.55 & 9.86  & 10.07 & 47.63 & 28.92 & 30.64 & 68.91 & 37.96 & 40.64 \\
     MiniCPM-V-2.6-8B \citep{yao2024minicpm} & 79.66 & 5.07 & 3.03 & 3.30 & 26.80 & 11.07 & 12.51 & 53.36 & 18.98 & 22.09 \\
     Qwen3-VL-8B-Instruct \citep{bai2025qwen3vltechnicalreport} & 92.76 & 13.18 & 10.02 & 10.31 & 51    & 29.44 & 31.63 & 69.2  & 36.7  & 39.97 \\
       InternVL3-9B \citep{zhu2025internvl3}& 88.43 & 11.42 & 8.79 & 8.95 & 44.54 & 25.36 & 26.98 & 63.49 & 34.47 & 36.96 \\
       GLM-4V-9B \citep{glm2024chatglm} & 91.93 & 7.14 & 4.86 & 5.12 & 27.42 & 13.17 & 14.50 & 59.50 & 24.64 & 27.70 \\
           InternVL3-14B \citep{zhu2025internvl3} & 66.18 & 8.19 & 7.47 & 7.28 & 29.27 & 22.30 & 22.81 & 40.90 & 29.67 & 30.49 \\
        CogVLM2-Llama3-Chat-19B \citep{hong2024cogvlm2} & 78.41 & 2.43 & 0.87 & 1.06 & 10.28 & 3.92 & 4.56 & 25.84 & 8.37 & 10.36 \\
     Qwen2.5-VL-32B-Instruct \citep{yang2024qwen2} & 97.83 & 13.05 & 9.87 & 10.10 & 48.65 & 28.94 & 30.99 & 68.14 & 37.97 & 41.04 \\
     Qwen3-VL-32B-Instruct \citep{bai2025qwen3vltechnicalreport} & 97.56 & 16.04 & 12.92 & 13.15 & 55.91 & 35.9  & 37.86 & 72.7  & 43.53 & 46.37 \\
      GLM-4.5V-106B \citep{5team2025glm45agenticreasoningcoding}& 99.8  & 14.13 & 11.71 & 11.83 & 52.7  & 34.48 & 36.12 & 71.98 & 43.49 & 45.87 \\
    Qwen3-VL-235B-A22B-Instruct \citep{bai2025qwen3vltechnicalreport}& 99.58 & 17.59 & 13.75 & 14.19 & 63.76 & 37.97 & 40.57 & 85.25 & 45.7  & 49.45 \\

    \bottomrule
    \end{tabular}
  }
 \caption{Comparison of model performance on the controlled chart-to-table task. Pass Rate indicates the proportion of examples for which the model produced a parsable table. P, R, and F1 denote precision, recall, and F1-score, respectively. Colored markers represent different tolerance levels: 
\textcolor{softred}{\rule{0.5em}{0.5em}} for strict, 
\textcolor{softyellow}{\rule{0.5em}{0.5em}} for slight, and 
\textcolor{softgreen}{\rule{0.5em}{0.5em}} for high.}
  \label{tab:tabletask}%
\end{table*}%

%% file: Section/conclusion.tex
\section{Conclusion}

In this work, we introduce ChartAnchor, a comprehensive benchmark for evaluating chart grounding. Unlike prior tasks, ChartAnchor unifies chart-to-code generation and controlled table reconstruction to rigorously assess visual, structural, and numerical fidelity. The dataset encompasses 8,068 chart–table–code triplets spanning 30 diverse types and multiple plotting libraries. We further proposed a multi-level evaluation protocol covering functional validity, visual consistency, data accuracy and perceptual similarity. Experiments across 24 MLLMs reveal that even the performance leader, GPT-5, continues to struggle with fine-grained data recovery and visual structure consistency. These findings underscore the need for integrating symbolic reasoning with visual precision, and we hope ChartAnchor catalyzes research on robust chart comprehension in critical domains.

\section{Limitations and Future Work}
While ChartAnchor currently targets static charts, real-world visualizations increasingly leverage interactive and dynamic components. Future work aims to extend the benchmark to encompass a broader spectrum of dynamics, such as drill-down plots, animated data transitions, and multi-view dashboards. This expansion enables the evaluation of model proficiency in handling dynamic semantics and multi-state rendering, significantly broadening the benchmark’s relevance to realistic analytical environments.

%% file: Section/appendix.tex
\newpage
\section*{Appendix}
\tableofcontents

\addcontentsline{toc}{section}{Appendices}  
\input{Section/appendix/data_coll}

\input{Section/appendix/dataset}

\input{Section/appendix/license}

\input{Section/appendix/impact}

\input{Section/appendix/implement}

\input{Section/appendix/case}
\input{Section/appendix/vis}

%% file: Section/appendix/data_coll.tex
\section{Detailed Data Collection and Processing Procedures}
\label{app:coll}
In this appendix, we provide a comprehensive description of our data collection, code synthesis, and quality filtering procedures for constructing the ChartAnchor dataset, a large-scale corpus of chart–table–code triples designed to support chart understanding and structured generation tasks. The goal of our pipeline is to ensure semantic fidelity, visual diversity, and execution reliability, making the dataset suitable for rigorous evaluation of large multimodal models (MLLMs).

\subsection{Data Sources and Corpus Composition}

We collect a total of \textbf{230,549 raw chart--table pairs}, which serve as the foundation for constructing the ChartAnchor corpus. These samples are derived from two sources:
\begin{itemize}
    \item \textbf{218,549 chart--table--code triplets} crawled from open-source visualization platforms, where each sample includes a rendered chart image, the structured data table, and the corresponding plotting script (mainly written in Plotly).
    \item \textbf{12,000 chart--table pairs} sampled from four existing chart-realted  datasets that provide image--table pairs but do not include source code.
\end{itemize}

In the remainder of this section, we describe the composition and characteristics of both sources in detail.

\textbf{\textit{1. Existing Datasets.}}
To bootstrap the construction of ChartAnchor, we leverage four well-known chart-centric datasets: \textit{PlotQA}, \textit{DVQA}, \textit{FigureQA}, and \textit{VisText}. These datasets collectively comprise over \textbf{750,000 chart samples}, each offering paired chart images and tabular data, but lacking executable plotting code. From each dataset, we uniformly sample \textbf{3,000 representative chart--table pairs}, yielding a total of \textbf{12,000 initial samples} for downstream code synthesis. The sampled charts span \textbf{five major types}: bar, line, scatter, area, and pie charts.

Below, we briefly describe the origin and structure of each dataset:

\begin{itemize}[leftmargin=*]
  \item \textbf{PlotQA} is a large-scale dataset constructed from real-world sources such as the World Bank, government portals, and open data platforms. It covers diverse domains including economics, health, education, and the environment. It contains \textbf{224,377 chart images}, primarily bar, line, and scatter plots, each paired with structured table data and metadata.

  \item \textbf{DVQA} is a synthetic dataset focused on bar chart understanding, comprising \textbf{300,000 high-resolution images} generated via \texttt{Matplotlib}. Each sample is associated with its underlying data table and metadata describing chart layout elements.

  \item \textbf{FigureQA} contains \textbf{over 100,000 synthetic chart images} generated using the \texttt{Bokeh} library, covering five figure types with structural annotations such as bounding boxes and labels.

  \item \textbf{VisText} comprises \textbf{12,441 chart--caption samples}, each including a chart image, data table, and scene graph. Charts are rendered using \texttt{Vega-Lite} with real-world data from Statista.
\end{itemize}

\begin{table*}[htbp]
  \centering

  \resizebox{\textwidth}{!}{
    \begin{tabular}{cccccccccc}
    \toprule
    Type  & Scatterpolar & Cone  & Line3d & Carpet & Barpolar & Mesh3d & Ohlc  & Line  & Histogram2d  \\
    Num   & 8193  & 384   & 10000 & 164   & 1599  & 10000 & 3949  & 10000 & 4873   \\
    \midrule
    Type  & Areachart & Box   & Scatterternary & Waterfall & Heatmap & Scatter & Scatter3d & Surface & Histogram  \\
    Num   & 5740  & 9530  & 3915  & 395   & 9980  & 10000 & 7300  & 10000 & 10000  \\
    \midrule
    Type  & Treemap & Violin & Parcoords & Funnelarea & Funnel & Sankey & Candlestick & Contour & Sunburst  \\
    Num   & 1122  & 7960  & 3567  & 219   & 716   & 8910  & 4750  & 10000 & 2962  
    \\
    \midrule
     Type  & bar   & pie   & histogram2dcontour & Density Tile Map & Tile Map & Atlas Map & Choropleth Atlas Map & Choropleth Tile Map & Image-based Table \\
    Num   & 8920  & 10000 & 5250  & 585   & 9382  & 9675  & 9260  & 513   & 8736 \\
    \bottomrule
    \end{tabular}
  }
  \caption{Number of samples per chart type in the newly curated dataset.}
   \label{tab:rawnum}
\end{table*}

\textbf{\textit{2. Newly Curated Dataset.}}
To enhance chart diversity and obtain source code supervision, we crawl \textbf{218,549 chart--table--code triplets} from open-source visualization communities. These samples are created by users across a wide range of domains and include full plotting scripts. Each sample contains a rasterized chart image, the underlying data table, and a Python script  that can regenerate the visualization.

This collection spans \textbf{over 36 chart types}, including both standard (e.g., bar, line, scatter) and specialized forms (e.g., candlestick, violin, sankey). Table~\ref{tab:rawnum} summarizes the sample distribution by chart type.

\subsection{Detailed Code Generation and Augmentation for Existing Datasets }
To address the lack of source code in existing chart datasets, we design a parameterized code generation pipeline that translates chart metadata into executable Python plotting scripts. This enables us to construct chart--table--code triplets from datasets that originally only contain chart images and tabular data. The pipeline consists of three sequential stages: (1) semantic mapping, (2) visual attribute parameterization, and (3) controlled data augmentation.

\paragraph{Semantic Mapping.}
We first parse the metadata associated with each chart (e.g., chart type, data structure, labels, axis information) and map it to corresponding primitives in popular visualization libraries. For example, bar charts are mapped to calls like \texttt{plt.bar()}, line charts to \texttt{plt.plot()}, and scatter plots to \texttt{plt.scatter()}. We also infer high-level layout logic such as multiple series plotting, stacked vs. grouped bar configurations.

\paragraph{Visual Attribute Parameterization.}
We define a structured set of visual attributes that govern the appearance of chart renderings. These attributes cover visual elements such as colors, strokes, fonts, axes, legends, and layout. Each attribute corresponds to a configurable parameter in the plotting code and forms the basis for subsequent augmentation.

\begin{itemize}[leftmargin=*]
  \item \textbf{Color schemes}: parameters defining the color of key visual elements including lines, bars, markers, and background.
  \item \textbf{Stroke and marker styles}: properties such as line width, dash pattern, and marker shape applicable to strokes or data points.
  \item \textbf{Font settings}: parameters specifying font size, family, and weight for chart titles, axis labels, and tick labels.
  \item \textbf{Axis configuration}: includes axis visibility, tick mark density and orientation, label formatting, and scaling behavior (e.g., linear vs. log).
  \item \textbf{Legend configuration}: layout options including visibility, location, frame style, and padding.
  \item \textbf{Canvas layout}: overall figure width, height, and aspect ratio, affecting spatial organization and density.
\end{itemize}

\paragraph{Systematic Augmentation.}
To further increase style diversity, we apply controlled random perturbations over the sampled parameters. All augmentations are range-constrained to preserve semantic structure while introducing sufficient variability. The applied strategies include:

\begin{itemize}[leftmargin=*]
 \item \textbf{Color perturbation}: visual element colors are augmented through multi-level sampling and transformation. Initial base colors are perturbed in HSV color space by applying random shifts to hue ($\pm$0.2), saturation ($\pm$0.25), and brightness ($\pm$0.25), resulting in perceptually similar yet distinct styles. For elements requiring visual separation (e.g., gridlines or multiple series), two contrasting strategies are adopted: (i) \emph{same-family} perturbation with minimal hue deviation to maintain stylistic consistency, and (ii) \emph{complementary sampling}, where colors are rotated approximately 180 degrees in hue and slightly jittered to avoid exact symmetry. All foreground colors—such as text, ticks, and gridlines—are dynamically adjusted based on the chart background to ensure a minimum contrast ratio of 3:1, as computed using relative luminance. If automatic contrast resolution fails, fallback binary colors (black or white) are applied to maintain legibility.

  \item \textbf{Line and stroke styling}: visual stroke properties—including line width, line pattern, and marker shape—are randomly selected from constrained sets. Line widths are sampled uniformly from a predefined range (e.g., 1.2 to 3.5 pt), and styles are chosen from standard patterns such as solid, dashed, dotted, and dash-dot. This variation simulates a broad range of visual densities and chart semantics, while ensuring clarity and readability.

\item \textbf{Grid and frame styling}: the presence of gridlines is toggled with a fixed probability (e.g., 70\%). If enabled, grid properties including color, alpha transparency (range: 0.2–0.6), and line style are randomly assigned. Grid color may be either a low-contrast perturbation of the primary visual element (same-family) or a dynamically selected complementary color that satisfies a minimum contrast ratio relative to the background. Chart frame (spine) color and width are also independently perturbed if specified, or otherwise randomly assigned.

\item \textbf{Tick and axis label formatting}: font size for tick labels is sampled from a narrow range (e.g., 8–12 pt), and axis label font size is sampled separately (e.g., 10–14 pt). Tick direction, length, and visibility are randomized across axes. Label rotation angles are applied conditionally when present, or left unset to trigger automatic formatting. Axis spine visibility and style may also vary with domain-specific settings.

\item \textbf{Text and title layout}: title font size is sampled within a broader range (e.g., 12–18 pt) to accommodate both compact and expanded figure layouts. To preserve clarity in narrow plots, chart titles are automatically wrapped at fixed character widths (e.g., 50 characters per line) to avoid horizontal overflow and truncation, ensuring consistent rendering across canvas aspect ratios.

\item \textbf{Legend configuration}: legend display is toggled probabilistically. When shown, layout options including position (e.g., top-right, bottom-center), spacing, and frame style are randomly selected. This simulates visual clutter or compression effects common in real-world plots with many categories.

\item \textbf{Canvas and layout variation}: figure width and height are sampled from uniform distributions (e.g., width = 6 ± 3 units, height = 5 ± 2 units), producing aspect ratios ranging from portrait to landscape. These adjustments impact element scaling, whitespace, and overall plot density, thereby exposing models to varying layout constraints and visual balance conditions.

\end{itemize}

These attribute-level augmentations collectively simulate a broad range of real-world chart styles. By systematically varying color schemes, marker designs, axis configurations, legend layouts, font properties, and layout structures, the resulting code-image pairs exhibit substantial stylistic diversity while preserving semantic fidelity. This enables more robust training and evaluation of models in chart understanding and generation tasks. All scripts are programmatically verified for syntax correctness and rendering completeness, ensuring reproducibility and consistency across the constructed dataset.
\subsection{Filtering Strategy and Statistics}
\label{Filtering}
\begin{table}[h]
\centering
\begin{tabular}{lrr}
\toprule
\textbf{Filtering Stage}         & \textbf{Removed} & \textbf{Retained} \\
\midrule
Completeness Check               & 51,582           & 178,967 \\
Structural Filtering             & 52,963            & 126,004 \\
Deduplication                    & 62,525           & 63,479 \\
Executability Check              & 15,011            & 48,468 \\
Manual Filtering   & 40,400            & 8,068 \\
\bottomrule
\end{tabular}
\caption{Sample counts at each filtering stage.}
\end{table}

\textbf{Automated filtering.}

– \textit{Completeness filtering}: We removed all samples missing any of the core components: structured table, chart image, or generation code. This includes examples with null or corrupt image files, empty tables, or scripts lacking plotting calls. A total of \textbf{51,582} examples were eliminated at this step.

– \textit{Structural filtering}: We excluded chart instances whose structure could not be reliably reconstructed into code without external visual assets. This includes geospatial plots (e.g., choropleth tile map), rasterized or image-based tables, and charts containing embedded logos and background images that are not specified in the underlying table. In addition, we filtered out samples whose generated code exceeded a predefined length threshold, indicating excessive verbosity, redundant operations, or inclusion of unrelated plotting logic. This step removed \textbf{52,963} samples in total.

– \textit{Deduplication}: For each data table, we extracted column-level features including column type (categorical or quantitative), column length, and a representative statistic determined by type: the most frequent value for categorical columns and the mean for quantitative columns. These features were computed for all columns and concatenated in column order to form a string-based table signature. Charts with identical signatures were treated as duplicates, and only one instance was retained. This filtering step removed \textbf{62,525} structurally duplicated samples.

\paragraph{Example.} The following table:

\begin{center}
\begin{tabular}{ccc}
\toprule
City & Category & Score \\
\midrule
Paris & A & 88.0 \\
Paris & B & 92.0 \\
London & A & 84.0 \\
London & B & 94.0 \\
\bottomrule
\end{tabular}
\end{center}

is processed column by column to compute the table signature:

\begin{itemize}[leftmargin=*]
\item \textbf{Column 1 (City)}:
  \begin{itemize}
    \item Type: Categorical
    \item Length: 4
    \item Representative statistic: Most frequent value = \texttt{Paris}
  \end{itemize}
\item \textbf{Column 2 (Category)}:
  \begin{itemize}
    \item Type: Categorical
    \item Length: 4
    \item Representative statistic: Most frequent value = \texttt{A}
  \end{itemize}
\item \textbf{Column 3 (Score)}:
  \begin{itemize}
    \item Type: Quantitative
    \item Length: 4
    \item Representative statistic: Mean = 89.5
  \end{itemize}
\end{itemize}

These features are concatenated in column order to produce the table signature:

\begin{center}
\texttt{categorical4Pariscategorical4A-} \\
\texttt{quantitative489.5}
\end{center}

If another chart shares the same signature, it is considered structurally equivalent and is filtered out.

– \textit{Executability check}: Each Python script was executed in an isolated environment. We discarded any sample whose script resulted in errors (e.g., due to missing fields or malformed syntax), produced no visual output, or generated a blank or invalid image file. This step filtered \textbf{15,011} additional cases.

After automated filtering, a total of \textbf{48,468} high-confidence samples were retained from the crawled corpus.

\begin{table*}[h]
\centering
\renewcommand{\arraystretch}{1.35}
\resizebox{\textwidth}{!}{
\begin{tabular}{p{3.5cm}|p{3.8cm}|p{3.8cm}|p{3.8cm}}
\toprule
\textbf{Dimension} & \textbf{Accept} & \textbf{Borderline} & \textbf{Reject} \\
\midrule

\textbf{Semantic Accuracy} &
All columns in the table are correctly encoded in the chart; axis titles match column names; all data series are included; data-to-visual mappings are accurate and complete. &
Most relevant columns are included; minor mismatches in field-to-axis mapping, partial omission of non-critical fields, or inaccurate axis labeling may exist. &
One or more key columns are missing or misused; axis assignments do not match table structure; values are incorrectly encoded or hardcoded; chart misrepresents the data. \\

\midrule

\textbf{Visual Clarity} &
All text is readable; tick marks, labels, and gridlines are well-aligned; spacing and font sizes are appropriate; the chart has no overlaps or clipping. &
The chart is generally legible but contains minor issues such as crowded labels, small text, or slight misalignment of elements. &
The chart contains severe layout problems, including overlapping text, unreadable labels, distorted scaling, or clipped elements that obstruct interpretation. \\

\midrule

\textbf{Stylistic Diversity} &
The chart uses varied formatting choices in color, font size, spacing, or layout; visual elements (e.g., legend placement, label orientation) differ from other charts of the same type. &
Some formatting differences are present, but the chart closely resembles many others in the same category; variation is minimal. &
Formatting is nearly identical to multiple other charts; visual parameters (e.g., spacing, label orientation, font, and color) are reused without change. \\

\bottomrule
\end{tabular}
}
\caption{Review rubric for chart--table--code triples.}
\label{tab:review_rubric_integrated}
\end{table*}

\textbf{Human Filtering Protocol.} 

Each chart--table--code triple was manually reviewed to ensure semantic correctness, visual interpretability, and style variation. Reviewers were graduate-level annotators with experience in Python scripting and data visualization. A total of six reviewers participated in the process, which was completed over the course of ten days.

The review was conducted in two passes: independent blind annotation followed by adjudication in case of disagreement. To ensure intra-group consistency, samples were first grouped by structural chart type (e.g., bar, line, scatter). Each triple was then evaluated along three axes—\textit{semantic accuracy}, \textit{visual clarity}, and \textit{stylistic diversity}—following the rubric in Table~\ref{tab:review_rubric_integrated}.

Each dimension was assigned one of three labels: (1)\texttt{Accept}, (2)\texttt{Borderline}, or (3)\texttt{Reject}. Samples receiving a double-\texttt{Reject} on any axis were removed. Disagreements or borderline cases were reviewed by a third annotator, who was allowed to execute or minimally adjust code to resolve ambiguity.

In total, \textbf{48,468} samples were flagged for human review, of which \textbf{40,400} were removed,  \textbf{8,068} accepted without change.

%% file: Section/appendix/dataset.tex
\section{More Analysis about Dataset}

\label{app:dataset}

We tokenize each code script in the benchmark and compute the token length for each example. The analysis shows an average length of approximately 628 tokens with a standard deviation of 466, and a minimum of 58 tokens. This indicates that the dataset includes both concise, logically clear scripts and longer, more complex ones, reflecting a broad range of task difficulty and coverage of real-world scenarios. Such diversity provides a solid foundation for evaluating the generalization capabilities of multimodal models across varying levels of complexity.

We categorize all table columns in the dataset into three types: string, numeric, and date. Numeric columns include both integers and floating-point numbers. Columns that do not meet either criterion are classified as strings by default.  Our analysis shows that 71\% of the columns are numeric, 23\% are string, and 6\% are date. This distribution indicates that numeric fields dominate the dataset, aligning with the inherently quantitative nature of most data visualizations. At the same time, the substantial presence of string and date fields highlights the dataset's semantic diversity, supporting categorical labeling and temporal trends. These findings demonstrate that ChartAnchor offers broad coverage of semantic structures commonly found in real-world data analysis tasks.

%% file: Section/appendix/license.tex
\section{Model License}
\label{app:license}
Table~\ref{tab:model_licenses} summarizes the licenses of all models evaluated in ChartAnchor, including both model weights and accompanying code repositories.

\renewcommand{\arraystretch}{1}
\begin{table*}[htbp]
\centering
\begin{tabular}{l l l}
\toprule
\textbf{Model} & \textbf{Model License} & \textbf{Code License} \\
\midrule
GPT-4o & Proprietary & Proprietary \\
GPT-5 & Proprietary & Proprietary \\
Claude-3-7-Sonnet & Proprietary & Proprietary \\
Claude-4-5-Sonnet & Proprietary & Proprietary \\
Gemini-3-Pro & Proprietary & Proprietary \\
InternVL3-2B & Apache 2.0 & MIT \\
Qwen2.5-VL-3B-Instruct & Apache 2.0 & Apache 2.0 \\
Gemma-3-4B-it & gemma & Not Applicable \\
DeepSeek-VL2-27B-A4B & deepseek & MIT \\
DeepSeek-VL-7B & deepseek & MIT \\
LLaVA-v1.6-Mistral-7B & Apache 2.0 & Apache 2.0 \\
Qwen2.5-VL-7B-Instruct & Apache 2.0 & Apache 2.0 \\
MiMo-VL-7B-SFT & MIT & MIT\\
MiMo-VL-7B-RL & MIT & MIT\\
MiniCPM-V-2.6-8B & minicpm & Apache 2.0 \\
Qwen3-VL-3B-Instruct & Apache 2.0 & Apache 2.0 \\
InternVL3-9B & Apache 2.0 & MIT \\
GLM-4V-9B & glm-4 & Not Applicable \\
CogVLM2-Llama3-Chat-19B &llama3 + cogvlm2 & Apache 2.0 \\
InternVL3-14B & Apache 2.0 & MIT \\
Qwen2.5-VL-32B-Instruct & Apache 2.0 & Apache 2.0 \\
Qwen3-VL-32B-Instruct & Apache 2.0 & Apache 2.0 \\
GLM-4.5V-106B  & MIT & Apache 2.0\\
Qwen3-VL-235B-A22B-Instruct & Apache 2.0 & Apache 2.0 \\
\bottomrule
\end{tabular}
\caption{ Summary of licenses in models that are evaluated in ChartAnchor. Entries marked with “Not
Applicable” indicate that authors do not have an explicit code license displayed within the codebase
or model checkpoint page.}
\label{tab:model_licenses}
\end{table*}

%% file: Section/appendix/impact.tex
\section{Broader Impacts}
\label{app:impact}

This work introduces a benchmark designed to evaluate the chart grounding ability of multimodal models through structured generation tasks—specifically, producing executable code and aligned tabular data from visual input. By formulating chart-centric understanding as a code- and table-grounded task, the benchmark enables more precise assessment of a model’s capacity to recover structured semantics from complex visualizations. The dataset is constructed from publicly available and license-compliant sources, with a focus on semantic traceability, syntactic validity, and reproducibility.

To mitigate potential risks, we adopt several safeguards: (1) the dataset emphasizes structured and verifiable content; (2) our filtering and annotation protocols (see Appendix~\ref{app:coll}) enforce consistency across modalities; (3) the evaluation suite includes tests for code execution, alignment fidelity, and structural coverage.

While the benchmark is designed to support research in grounded and interpretable generation, we acknowledge the possibility of unintended use. For instance, models trained or evaluated on ChartAnchor might be applied in automated settings without verification, potentially leading to misleading outputs. Although the benchmark itself does not directly enable such misuse, we recommend that future applications incorporate human oversight, validation mechanisms, and appropriate deployment constraints. Ensuring output traceability is particularly important when models are used in domains such as scientific computing, data journalism, and business reporting.

We hope that ChartAnchor serves as a resource for advancing multimodal systems that prioritize structured reasoning, factual alignment, and transparency.

%% file: Section/appendix/implement.tex
\section{Experiments}
\label{app:imple}
\begin{figure*}

\centering
\includegraphics[width=0.7\textwidth]{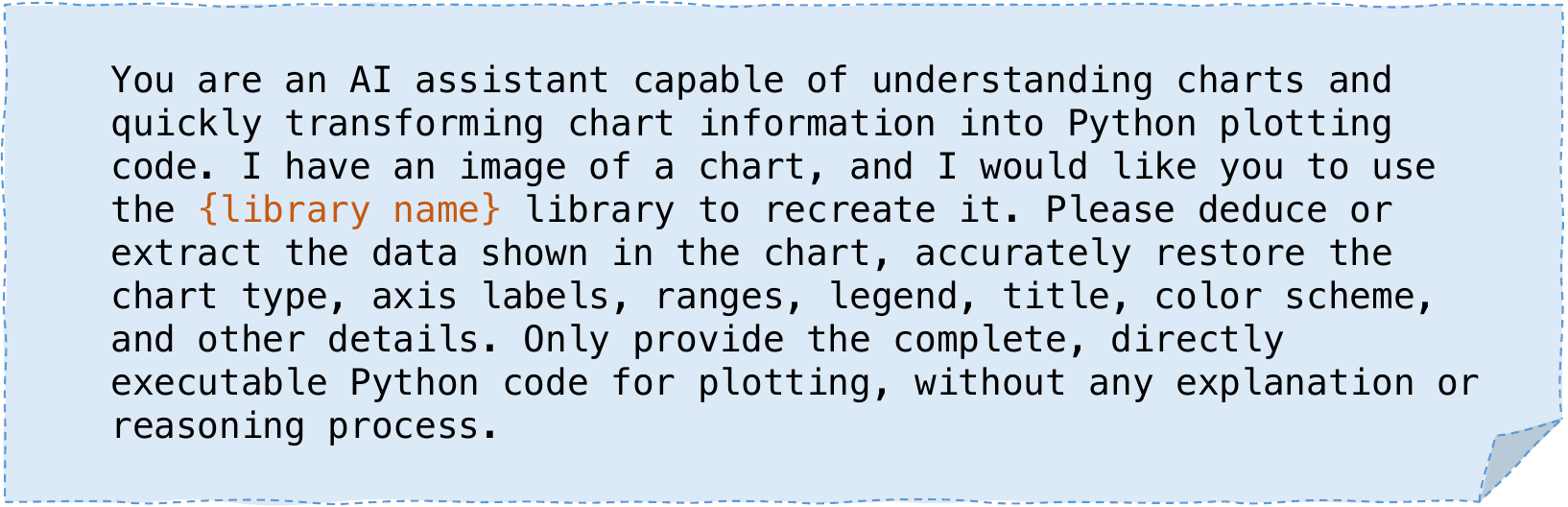}

\caption{Prompt for Chart-to-Code task. The orange text in curly braces denotes a selectable visualization library.} 
\label{fig:codeprompt}

\end{figure*}
\subsection{Implementation Details}
For all models, we set the temperature to $\tau = 0.1$ and use top-$p$ sampling with $p = 0.95$ for decoding. The maximum generation length is capped at 16384 tokens. For open-weight models, we adopt \texttt{bfloat16} precision during inference. All experiments are conducted on \texttt{H100 80G} GPUs. Results are reported from a single run per setting, with low-temperature decoding ensuring stability.

\subsection{Prompts}

\begin{figure*}

\centering
\includegraphics[width=0.7\textwidth]{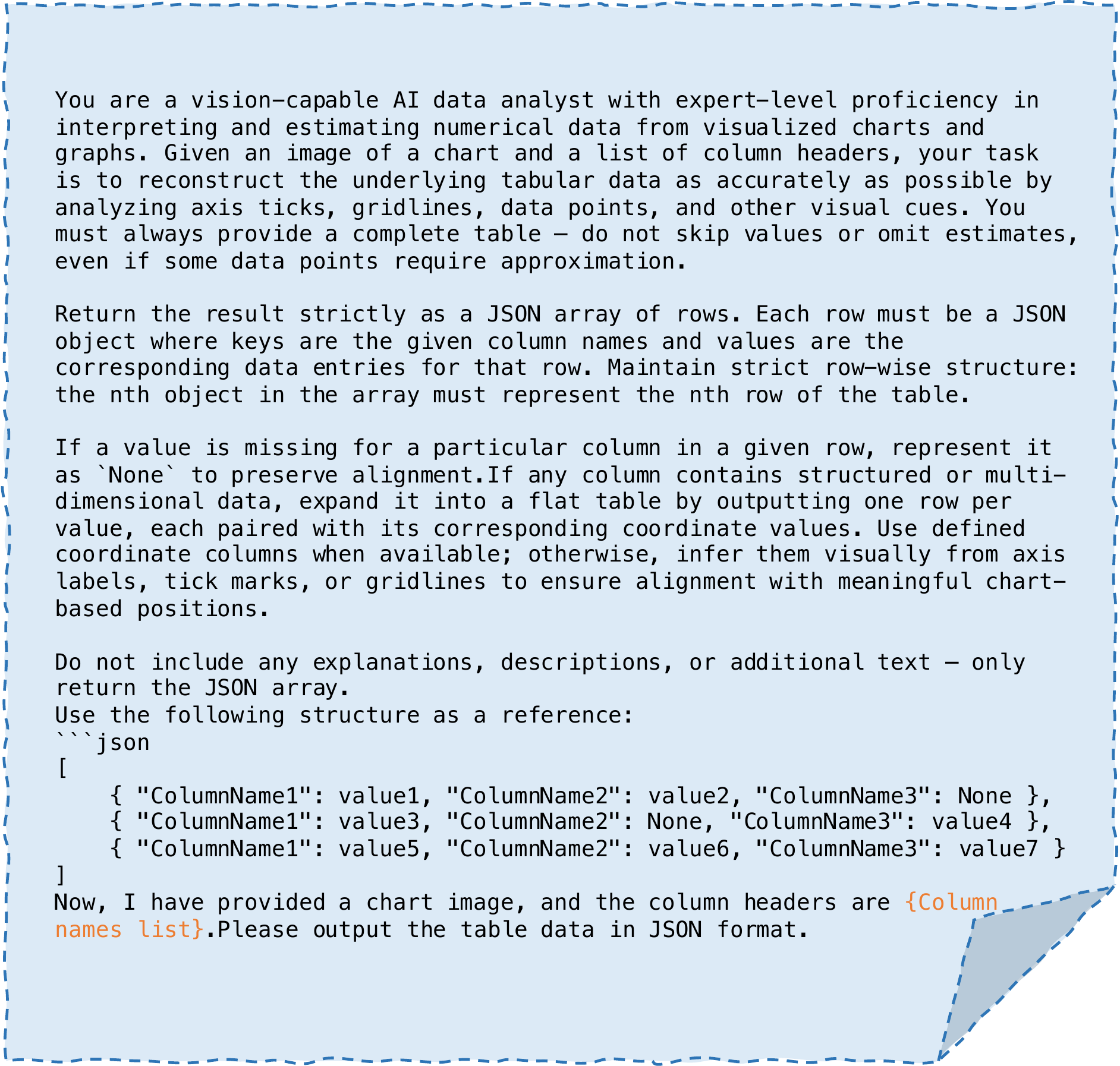}

\caption{Prompt for Controlled Chart-to-Table task. The orange text represents placeholders, which varies according to different table headers.} 
\label{fig:tableprompt}

\end{figure*}

We design tailored input prompts for each task to guide model behavior effectively. Figure~\ref{fig:codeprompt} illustrates the prompt used in the Chart-to-Code task, which specifies the desired output format and allows the model to choose between visualization libraries (e.g., Plotly or Matplotlib). Figure~\ref{fig:tableprompt} further presents the prompt format for the Controlled Chart-to-Table task, where key placeholders are dynamically adjusted based on the provided table headers.

\subsection{Additional Analysis}
As shown in Table ~\ref{tab:code_task} and Table ~\ref{tab:tabletask}, we can see: 

- \textit{Chart-to-Code Is Fundamentally Constrained by Long-Horizon Execution Requirements.}  
Both DeepSeek-VL and DeepSeek-VL2 exhibit consistently low pass rates on the Chart-to-Code task (32.05/7.98),  both operate under a maximum sequence length of 4096 tokens. Chart-to-Code requires long-horizon contextual coherence and strict execution fidelity, as successful generation depends on preserving object-level consistency across multiple code segments, including figure initialization, axis construction, and API-level correctness. Under limited context budgets, even minor inconsistencies in earlier object definitions can propagate into non-executable outputs. In contrast, tasks that do not enforce execution validity are far less sensitive to such long-context constraints. This indicates that Controlled Chart-to-Code inherently demands stronger long-context modeling and larger effective context windows, making it substantially more challenging than perception-oriented chart understanding tasks.

- \textit{Reasoning-Oriented Upgrades Yield Divergent Effects Across Chart Tasks.}  
While DeepSeek-VL2 shows clear improvements over VL on Controlled Chart-to-Table, it exhibits a performance drop on Chart-to-Code, revealing a model-level trade-off introduced by enhanced reasoning capabilities. Controlled Chart-to-Table benefits from stronger high-level abstraction and structural planning, as it primarily involves localized perception and short-context reasoning without strict execution constraints. In contrast, Chart-to-Code requires faithful continuation of previously established object states. Under limited effective context length, VL2’s stronger reasoning tendency may reconstruct code structures based on inferred intent rather than preserved execution history, leading to logically plausible but non-executable outputs. This effect may be further amplified by VL2’s mixture-of-experts architecture, where dynamic expert routing can improve specialization for localized reasoning but introduces additional challenges in maintaining stable, long-horizon state consistency across generation steps. As a result, reasoning-centric and MoE-based upgrades can jointly benefit short-context understanding tasks while exacerbating failure modes in long-context, execution-sensitive generation.

-  \textit{Color Feature Presents Unique Challenges in Visual Decoding.} Color accuracy remains the lowest-scoring aspect within the Visual Structure Consistency metrics, even for top-performing models such as GPT-5 (51.21), Claude-4-5-Sonnet(49.44) and Qwen3-VL-32B-Instruct(38.51). This suggests that fine-grained color differentiation poses unique challenges for current visual encoders, especially in complex or low-contrast chart regions. In many cases, the visual abstraction processes used by these models may reduce sensitivity to precise pixel-level color features. 

- \textit{High Pass Rate but Low Fidelity.Another failure mode is models generating syntactically valid outputs that “mask critical failures” in content.} For example, in controlled chart-to-table task, DeepSeek-VL-7B successfully produces parsable code nearly every time ( about 98\% pass rate) , yet its data extraction is almost nonexistent (strict F1 <1\%). It often writes basic chart code that runs but does not capture the actual data or visual details. This strategy yields a high functional score but extremely low data fidelity, indicating the model is defaulting to trivial or placeholder outputs to avoid errors rather than truly understanding the chart.

\textit{A High CLIP Score Does Not Guarantee Accurate Data Reconstruction.} A model may generate a chart that is visually convincing—possessing the correct chart type, structure, and color distribution—yet fails to preserve the underlying data. For instance, Claude-4.5-Sonnet achieves a high CLIP score (78.93), yet its data fidelity (36.70) falls slightly below that of Qwen3-VL-235B (38.07), despite the latter having a significantly lower CLIP score (68.62). This divergence illustrates that visual similarity and data accuracy are distinct dimensions. As noted in the introduction, a model may "reproduce a chart’s appearance while silently altering the data." Consequently, CLIP scores and Data F1 metrics must be viewed as complementary: the former detects visual discrepancies, while the latter reveals semantic errors. Both are indispensable for a comprehensive evaluation of chart fidelity.

\subsection{Analysis Across Chart Types}
\begin{figure*}

\centering
\includegraphics[width=1.0\textwidth]{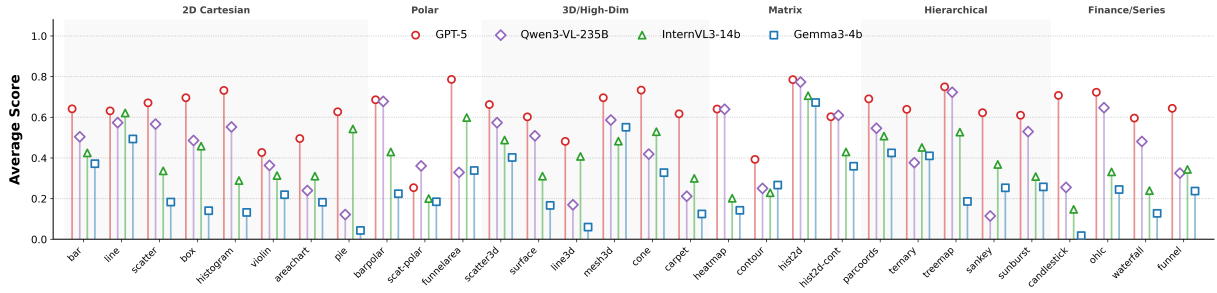}

\caption{Comparison of Overall Across Chart Types for Different Models on Chart-to-Code Task} 
\label{fig:codetypeoverall}
\end{figure*}

\begin{figure*}

\centering
\includegraphics[width=1.0\textwidth]{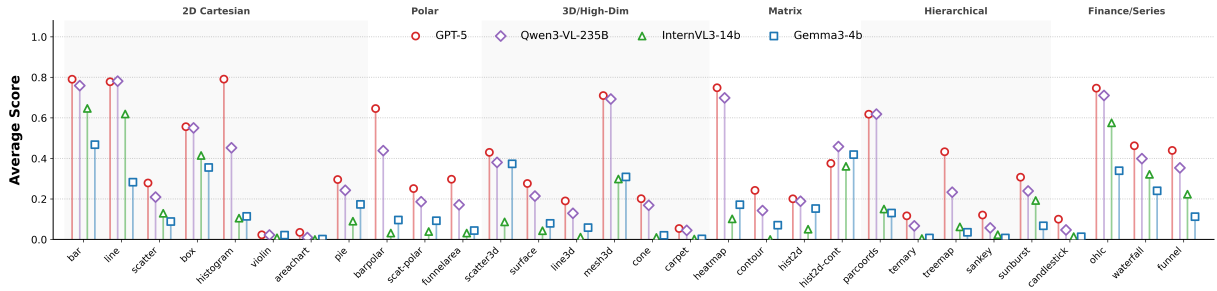}

\caption{Comparison of Slight F1-Score Across Chart Types for Different Models on Controlled Chart-to-Table Task} 
\label{fig:tabletypef1}
\end{figure*}

Figure~\ref{fig:codetypeoverall} shows that overall model performance varies significantly by chart type. Models perform best on simple 2D Cartesian charts like bar and line, which have consistent structures and are likely well-represented in training data. In contrast, performance drops sharply on 3D, matrix-style, and hierarchical charts, which involve complex layouts or dense data encoding. Financial charts show moderate stability, likely due to their standardized formats.

Compared to data  fidelity prediction (Figure~\ref{fig:data-attribute-accuracy}), some differences emerge. For instance, pie charts score high in data recovery but only moderately in overall performance, suggesting structural elements (e.g., legends) are hard to reproduce. Meanwhile, 3D and matrix charts perform worse in data accuracy than in overall score, highlighting the difficulty of recovering exact values from visually dense charts.

Figure~\ref{fig:tabletypef1} shows model performance on the chart-to-table task under slight tolerance. Basic Cartesian charts (e.g., bar, line) achieve the highest F1 scores due to their clear value mappings. In contrast, 3D, polar, and matrix-style charts perform poorly across models, reflecting challenges from visual distortion or dense layouts. Notably, hierarchical charts (e.g., sunburst, treemap) perform better here than in data attribute prediction (Figure~\ref{fig:data-attribute-accuracy}), suggesting that structured table formats help guide value extraction despite visual complexity.

\subsection{Correlation with Human Evaluation}
\label{app:human_alignment}
To assess the extent to which our multi-level evaluation framework aligns with human judgments, we conduct a human preference study based on pairwise comparisons in the chart-to-code generation task, which reduces annotator calibration bias and enables more reliable assessment of relative quality differences than absolute scoring.

\paragraph{Data Sampling.}
We randomly sample $N=300$ chart pairs from the ChartAnchor test set. For each pair $(C_A, C_B)$, two different models generate executable plotting code from the same ground-truth chart input, ensuring that any observed differences arise solely from model behavior rather than input variation.

\paragraph{Annotation Protocol.}
We recruit domain experts with prior experience in data visualization and chart analysis as annotators. For each chart pair, annotators are presented with the reference chart image together with the two rendered candidate charts. Following the core dimensions of our evaluation framework, annotators independently indicate their preference (\emph{``A is better''} or \emph{``B is better''}) along three axes:
\begin{itemize}
    \item \textbf{Visual Structure Consistency}, focusing on chart type, layout, textual elements, and color usage;
    \item \textbf{Semantic Data Fidelity}, emphasizing numerical accuracy and preservation of underlying data relationships;
    \item \textbf{Perceptual Quality}, capturing overall visual plausibility and semantic coherence.
\end{itemize}
No ties are allowed, enforcing a strict preference decision for each dimension.

\paragraph{Agreement Metrics.}
Rather than evaluating absolute quality, this study focuses on relative ranking alignment. For each chart pair and evaluation dimension, we compare the chart preferred by human annotators with the chart assigned a higher score by the corresponding automated metric. We report Consistency Accuracy (Acc) as the fraction of pairs for which the metric agrees with human preference. In addition, we compute Kendall’s rank correlation coefficient ($\tau$) to measure global agreement between metric-induced rankings and human preference orderings.

%% file: Section/appendix/case.tex
\section{Case Study of ChartAnchor}
\label{app:case}

\begin{figure*}[htbp]
  \centering 
  
  \begin{subfigure}[t]{0.35\linewidth} 
    \centering
    \includegraphics[width=\linewidth]{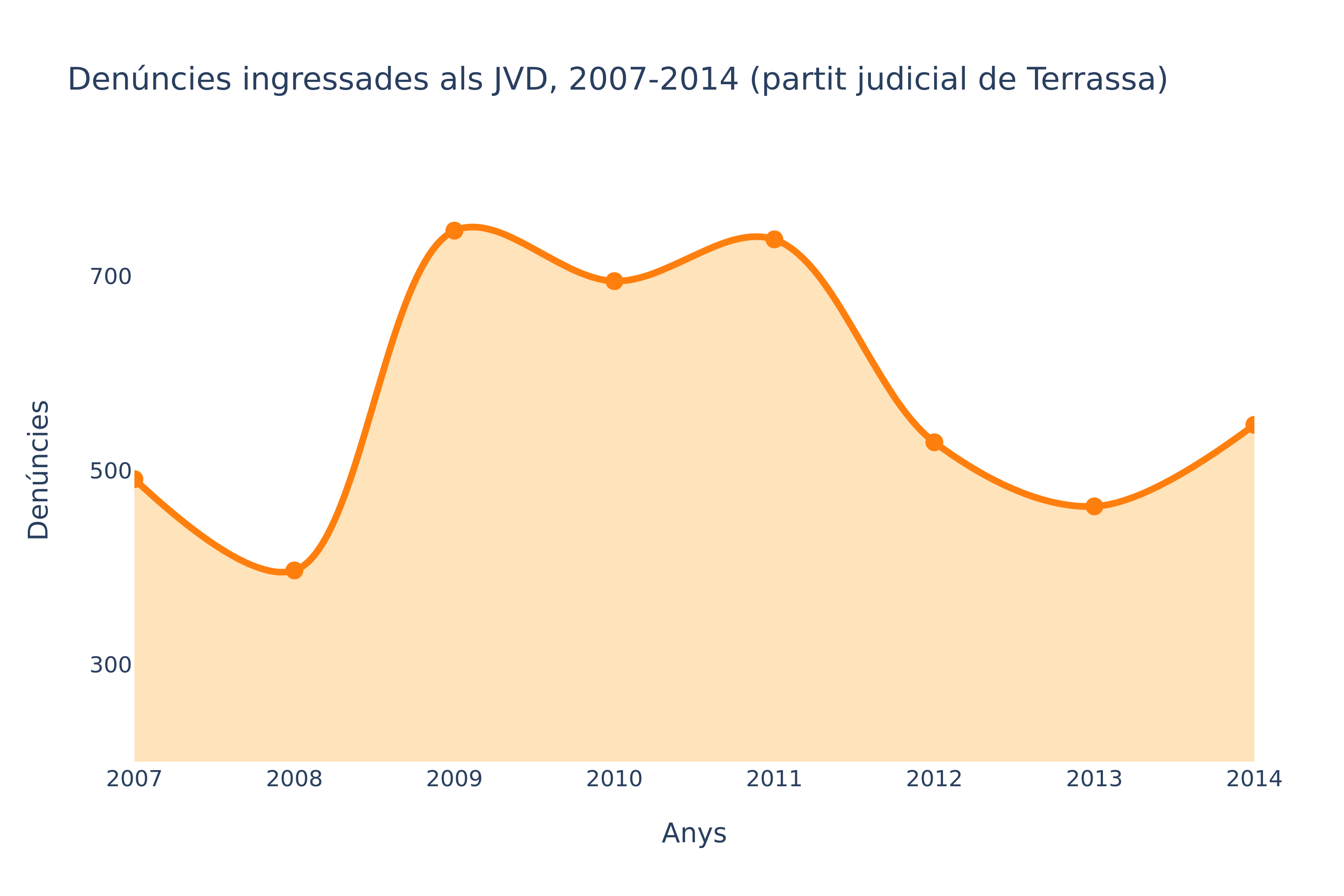}
    \caption{Gold Image}
  \end{subfigure}
  \hspace{1em} 
  \begin{subfigure}[t]{0.35\linewidth}
    \centering
    \includegraphics[width=\linewidth]{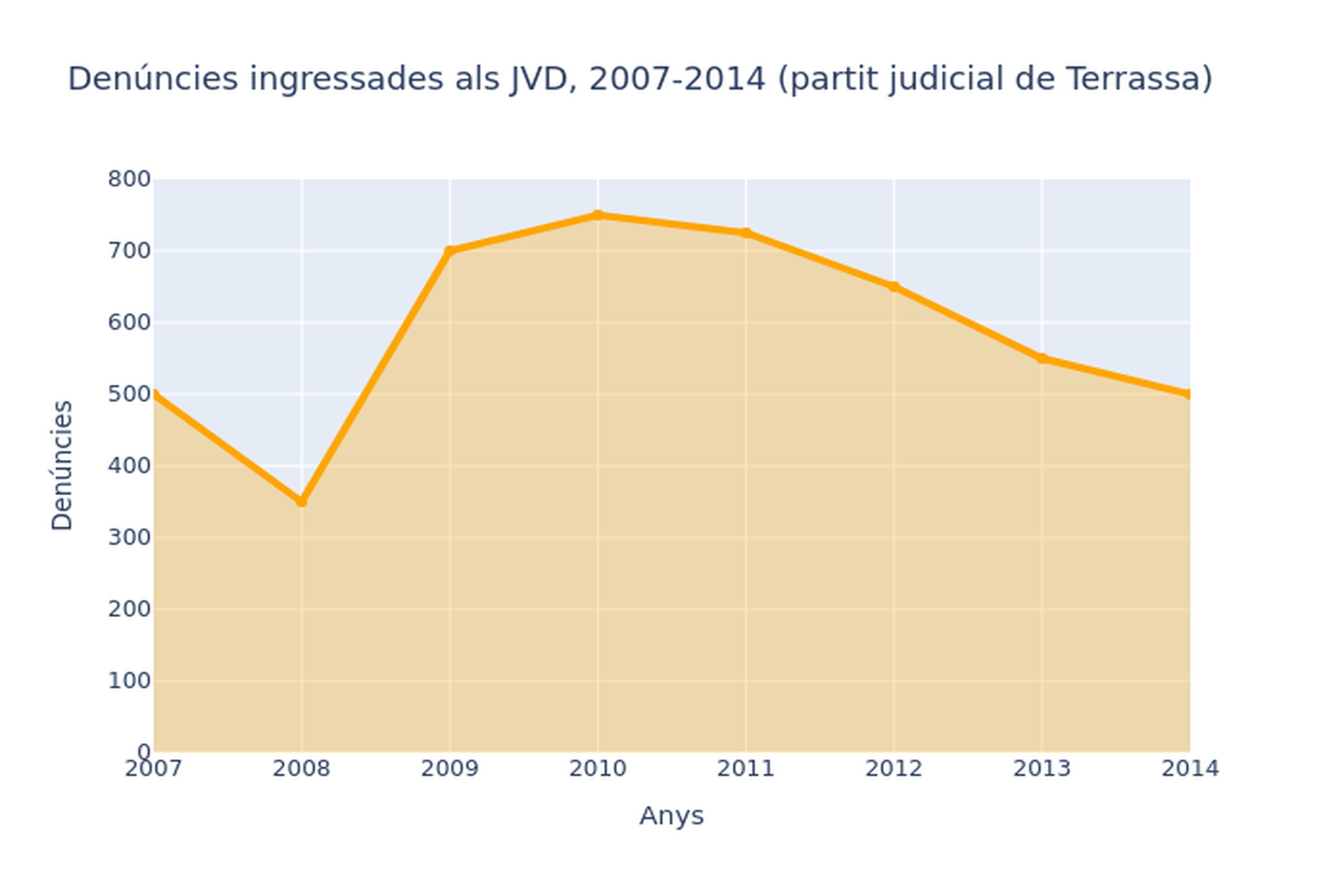}
    \caption{GPT4-o}
  \end{subfigure}
  
  \vspace{1em}

  \begin{subfigure}[t]{0.35\linewidth}
    \centering
    \includegraphics[width=\linewidth]{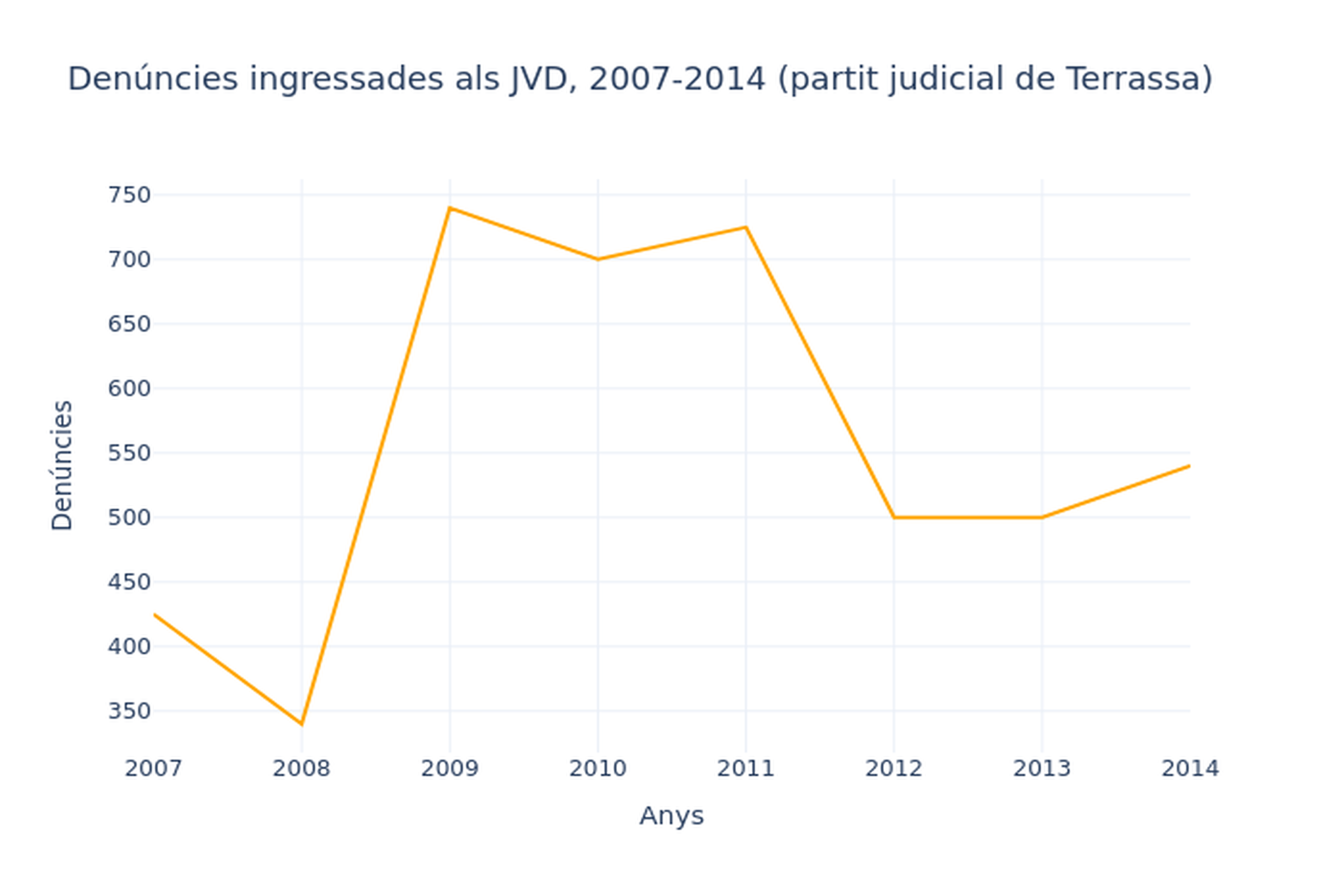}
    \caption{Gemma3-4b}
  \end{subfigure}
  \hspace{1em}
  \begin{subfigure}[t]{0.35\linewidth}
    \centering
    \includegraphics[width=\linewidth]{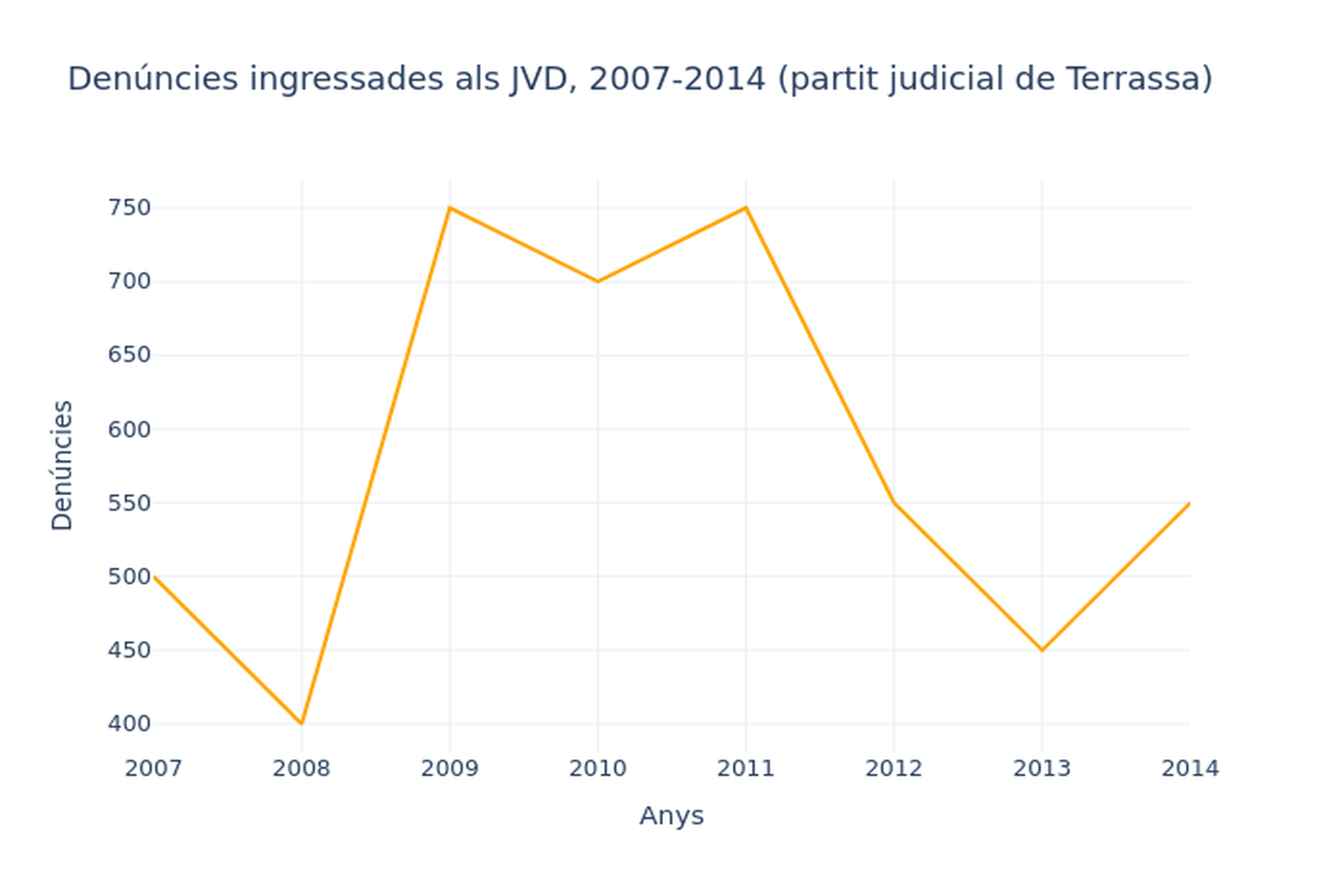}
    \caption{InternVL3-14b}
  \end{subfigure}
  
  \caption{Image Example1 for Chart-to-Code Task.}
  \label{fig:case1}
\end{figure*}

\textbf{Example 1}: Figure~\ref{fig:case1} illustrates the output of a Chart-to-Code task, comparing the gold reference image with the charts generated by three models: GPT-4o, Gemma3-4b, and InternVL3-14b. Among them, GPT-4o demonstrates the highest fidelity in visual styling, closely mimicking the original chart’s color fill, line smoothness, and overall aesthetic. In contrast, InternVL3-14b more accurately captures the data trends and year-to-year fluctuations, reflecting the original curve’s structure with greater numerical precision. 

\begin{figure*}[htbp]
  \centering
  \begin{subfigure}[t]{0.35\linewidth}
    \centering
    \includegraphics[width=\linewidth]{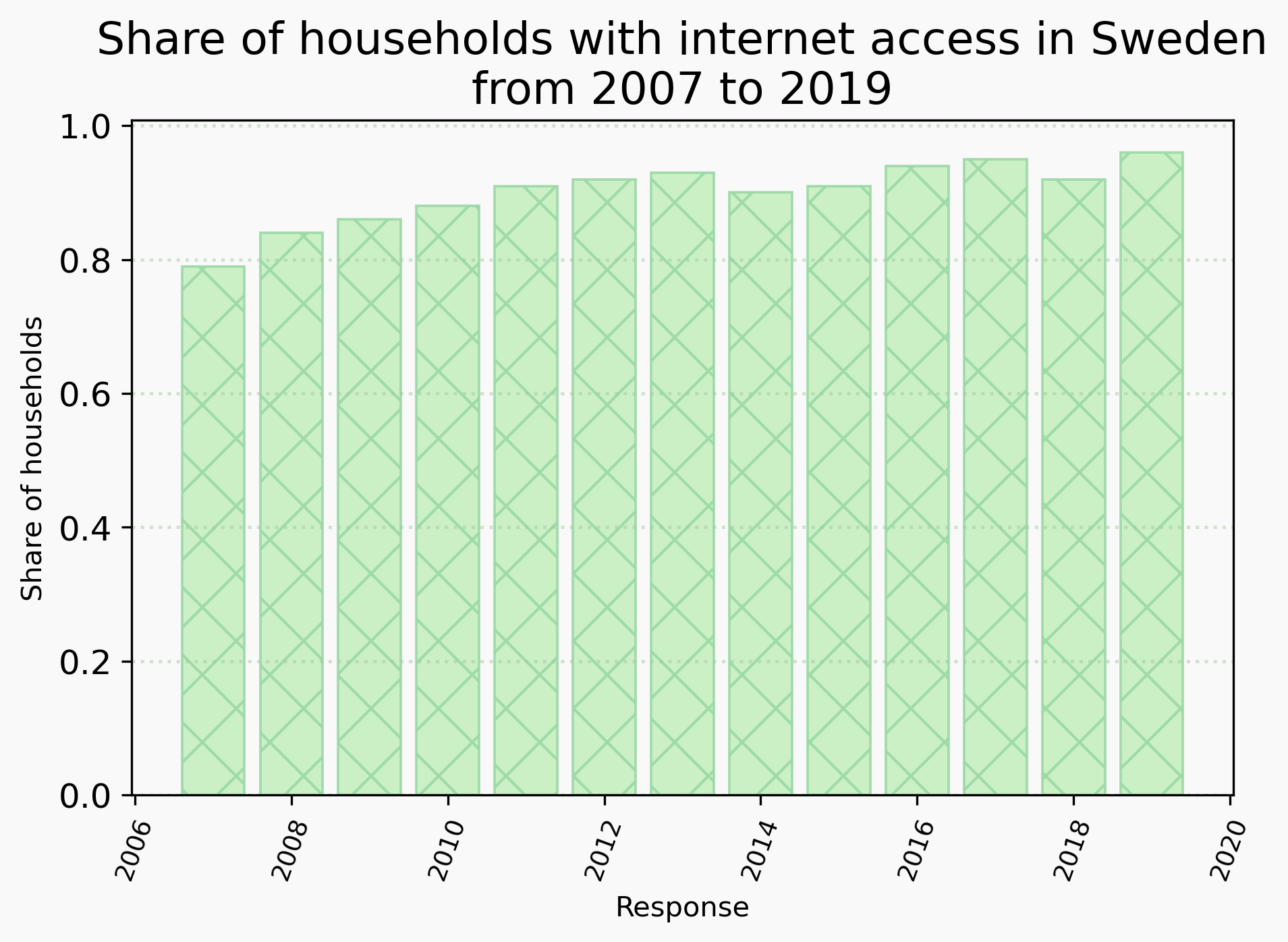}
    \caption{Gold Image}
  \end{subfigure}
\hspace{1em}
  \begin{subfigure}[t]{0.35\linewidth}
    \centering
    \includegraphics[width=\linewidth]{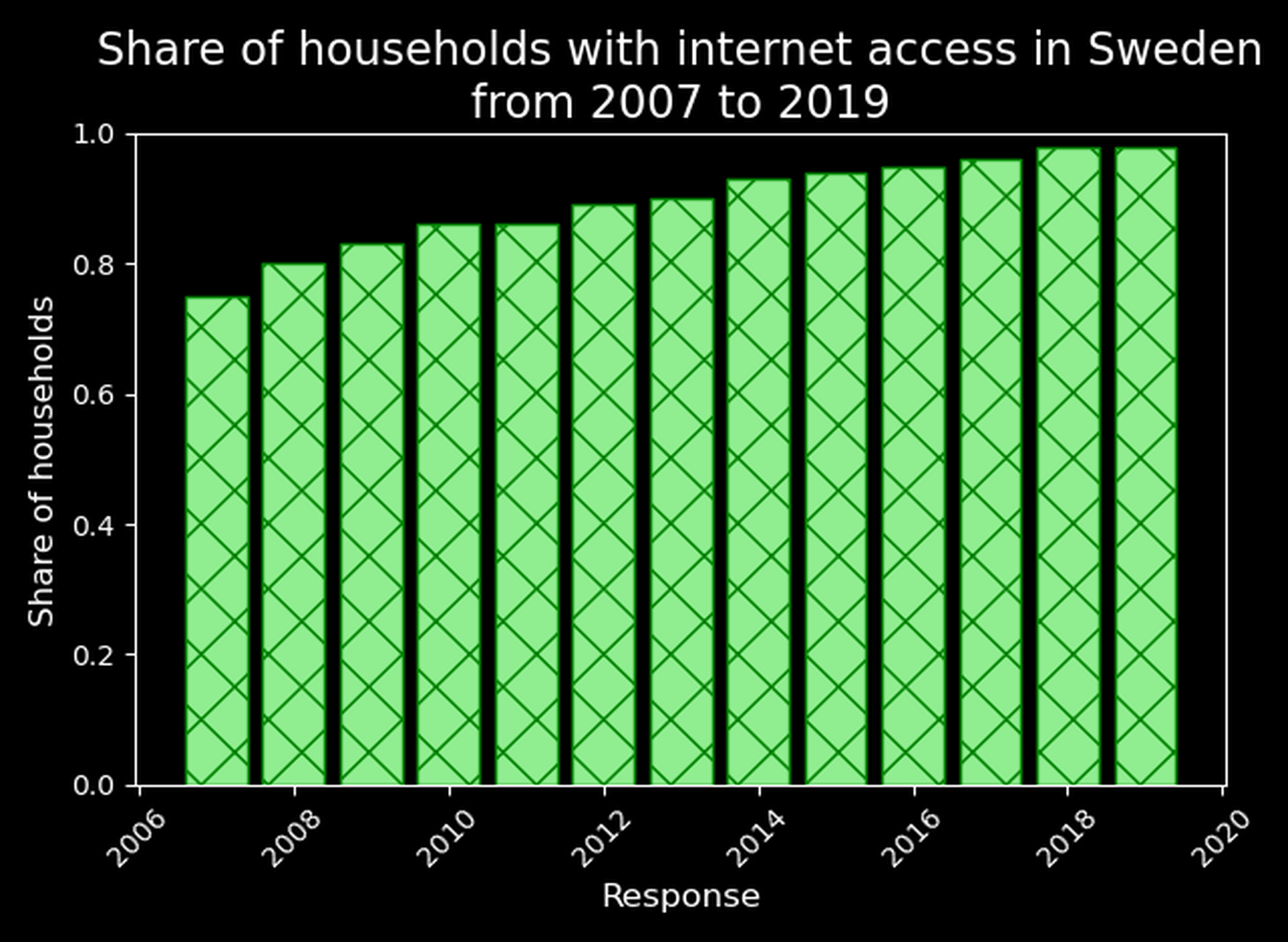}
    \caption{GPT4-o}
  \end{subfigure}
  \vspace{1em} 
  \\ 
  \begin{subfigure}[t]{0.35\linewidth}
    \centering
    \includegraphics[width=\linewidth]{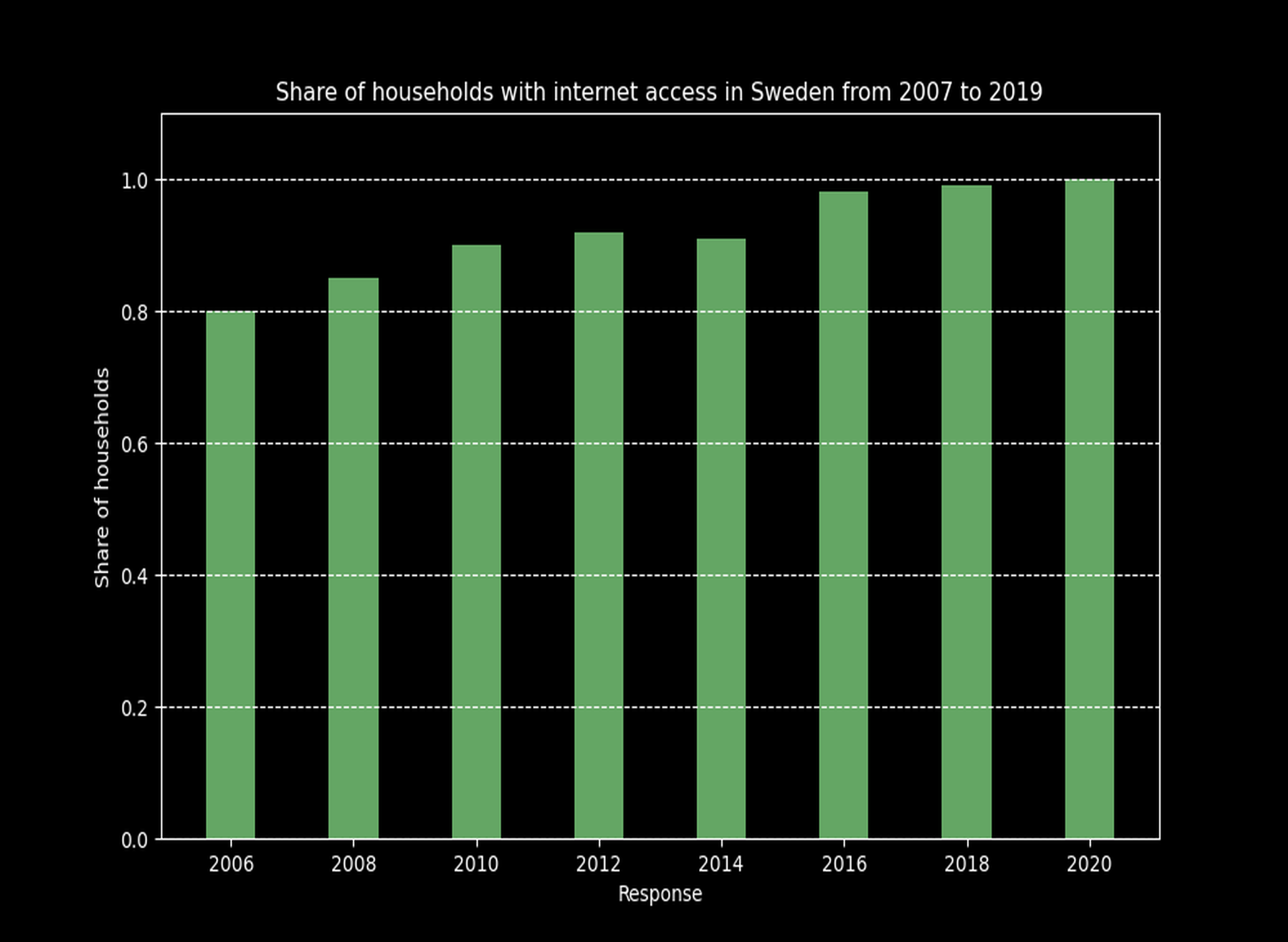}
    \caption{Gemma3-4b}
  \end{subfigure}
  \hspace{1em}
  \begin{subfigure}[t]{0.35\linewidth}
    \centering
    \includegraphics[width=\linewidth]{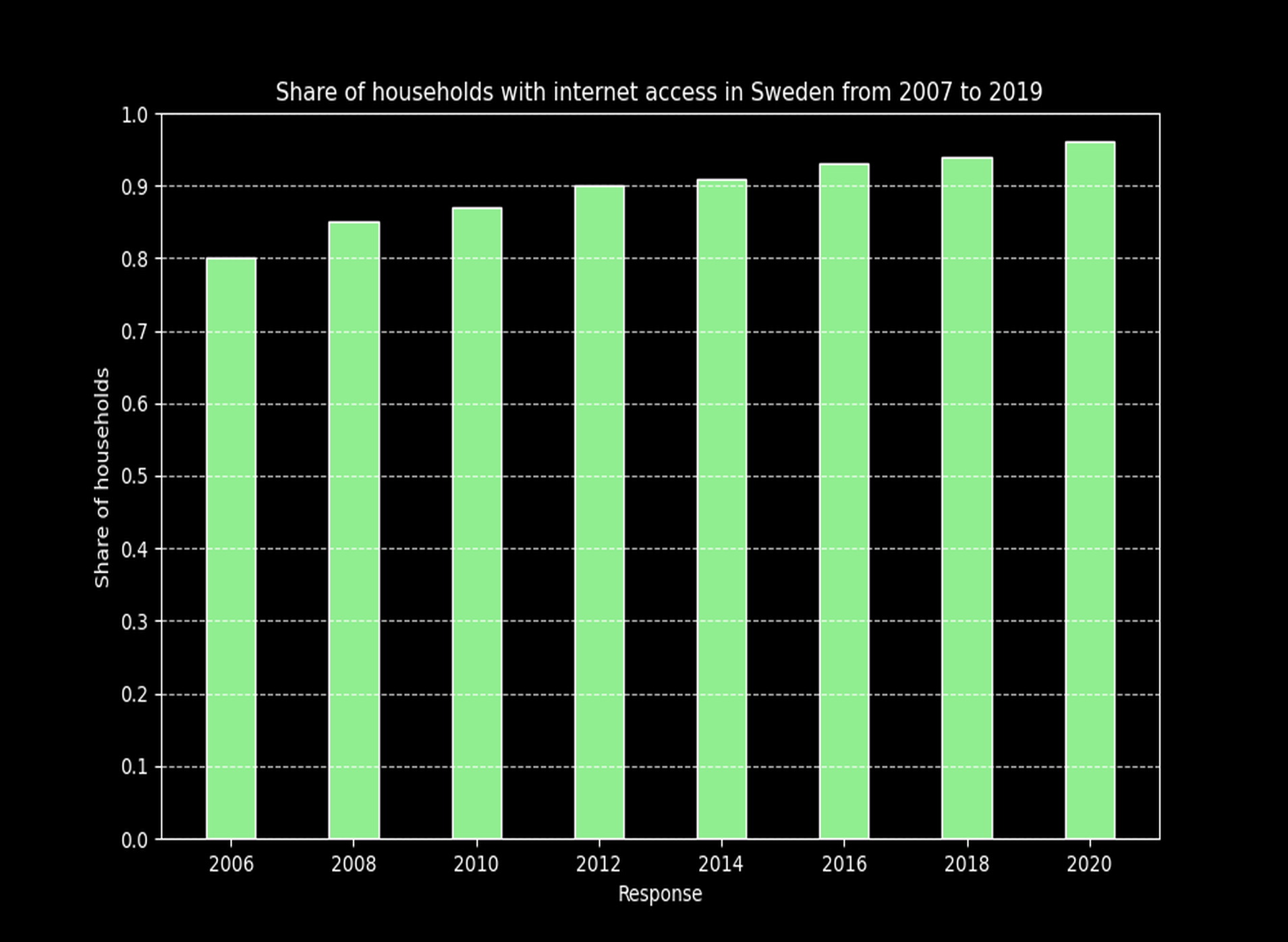}
    \caption{InternVL3-14b}
  \end{subfigure}
  \caption{Image Example2 for Chart-to-Code Task. }
  \label{fig:case2}
\end{figure*}

\textbf{Example 2}: As shown in figure~\ref{fig:case2}, GPT-4o offers the most faithful reproduction overall, closely matching the original chart’s data values and bar patterns, including the distinctive crosshatch fill—though it uses a dark background instead of light. In contrast, both Gemma3-4b and InternVL3-14b lack the patterned bars and display fewer, less precise data points, resulting in lower fidelity in both appearance and accuracy. While all models preserve the general upward trend, GPT-4o stands out for its high consistency in both visual styling and numerical detail.

\begin{figure*}[htbp]
\centering
\includegraphics[width=\textwidth]{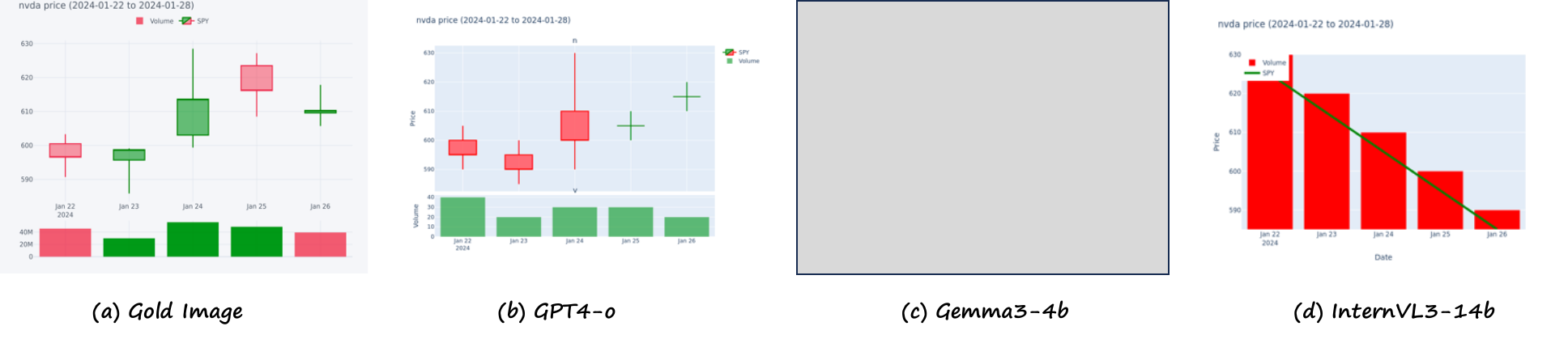}
\caption{Image Example3 for Chart-to-Code Task. The gray image indicates that the model did not generate a valid image or the code parsing failed} 
\label{fig:case3}
\end{figure*}

\textbf{Example 3}: As shown in figure~\ref{fig:case3}, in this example, GPT-4o is the only model that successfully replicates the candlestick chart structure from the gold image, preserving both the visual format and financial data elements. InternVL3-14b fails to produce a candlestick chart, instead outputting a misleading bar chart with incorrect trends. Gemma3-4b completely fails to render a valid image, as indicated by the gray placeholder.

\begin{figure*}[htbp]

\centering
\includegraphics[width=1.0\textwidth]{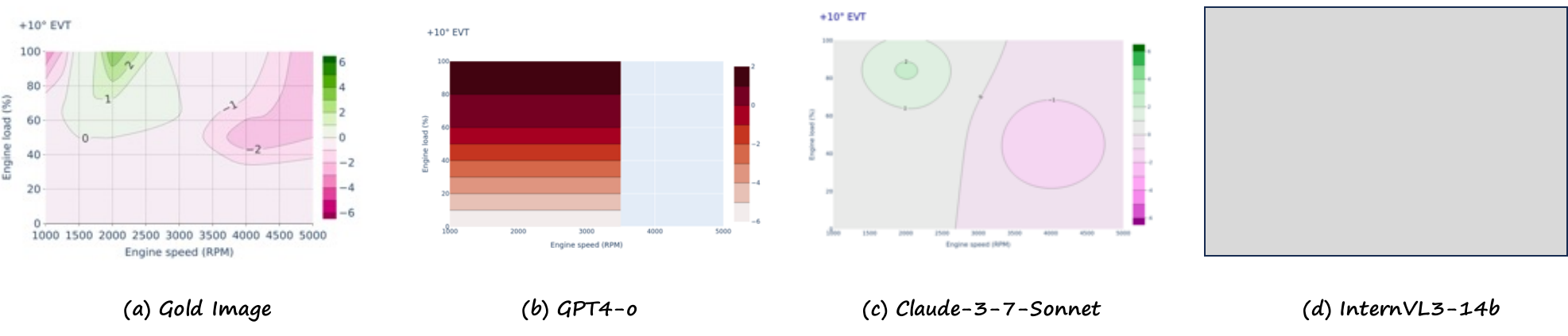}

\caption{Image Example4 for Chart-to-Code Task. The gray image indicates that the model did not generate a valid image or the code parsing failed} 
\label{fig:case4}
\end{figure*}

\textbf{Example 4}: As shown in figure~\ref{fig:case4}, in this example, Claude-3-7-Sonnet is the closest to replicating the contour style of the gold image, correctly preserving the shape and gradient regions, though with reduced detail. GPT-4o fails to generate contours, instead producing a simplified heatmap with horizontal bands that ignore spatial gradients. InternVL3-14b fails entirely, indicated by the gray placeholder.

\begin{figure*}[htbp]

\centering
\includegraphics[width=1.0\textwidth]{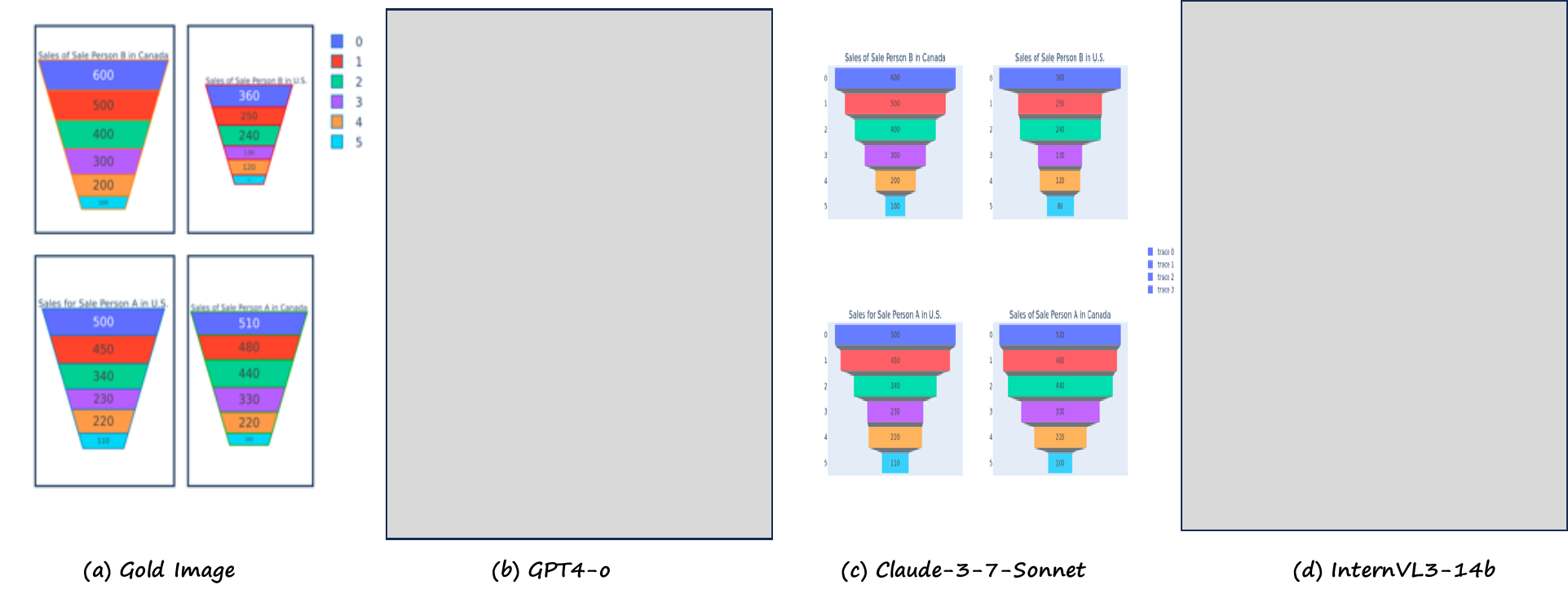}

\caption{Image Example5 for Chart-to-Code Task. The gray image indicates that the model did not generate a valid image or the code parsing failed} 
\label{fig:case5}
\end{figure*}

\textbf{Example 5}: As shown in figure~\ref{fig:case5}, only Claude-3-7-Sonnet successfully generates valid funnelarea charts, although the layout differs from the original: rearranges their positions. GPT-4o and InternVL3-14b both fail to render any output, shown as gray placeholders. While Claude’s version lacks stylistic fidelity, it maintains correct data values and category separation.

\begin{figure*}[htbp]
  \centering
  \begin{subfigure}[t]{0.45\linewidth}
    \centering
    \includegraphics[width=\linewidth]{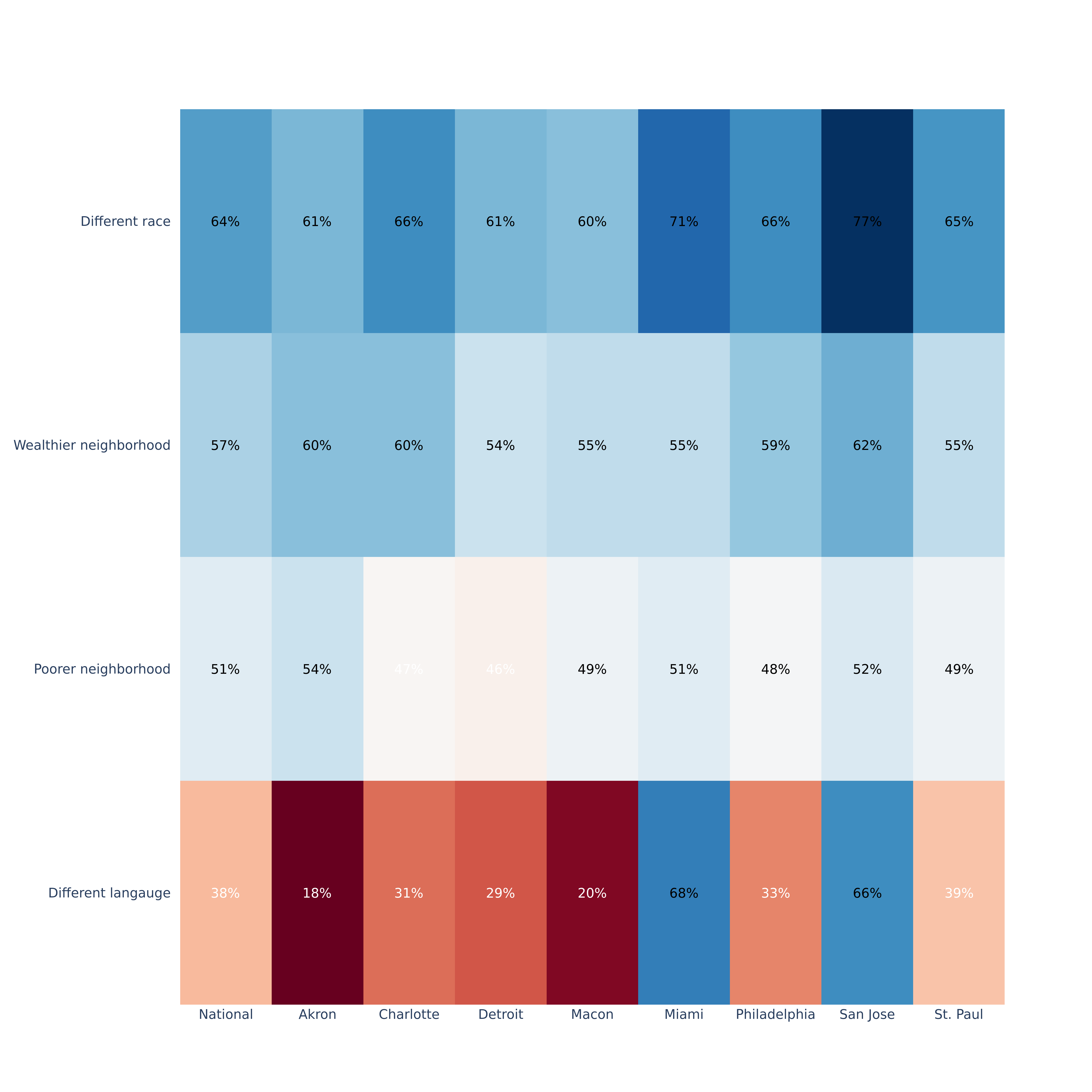}
    \caption{Gold Image}
  \end{subfigure}
  \hspace{1em}
  \begin{subfigure}[t]{0.45\linewidth}
    \centering
    \includegraphics[width=\linewidth]{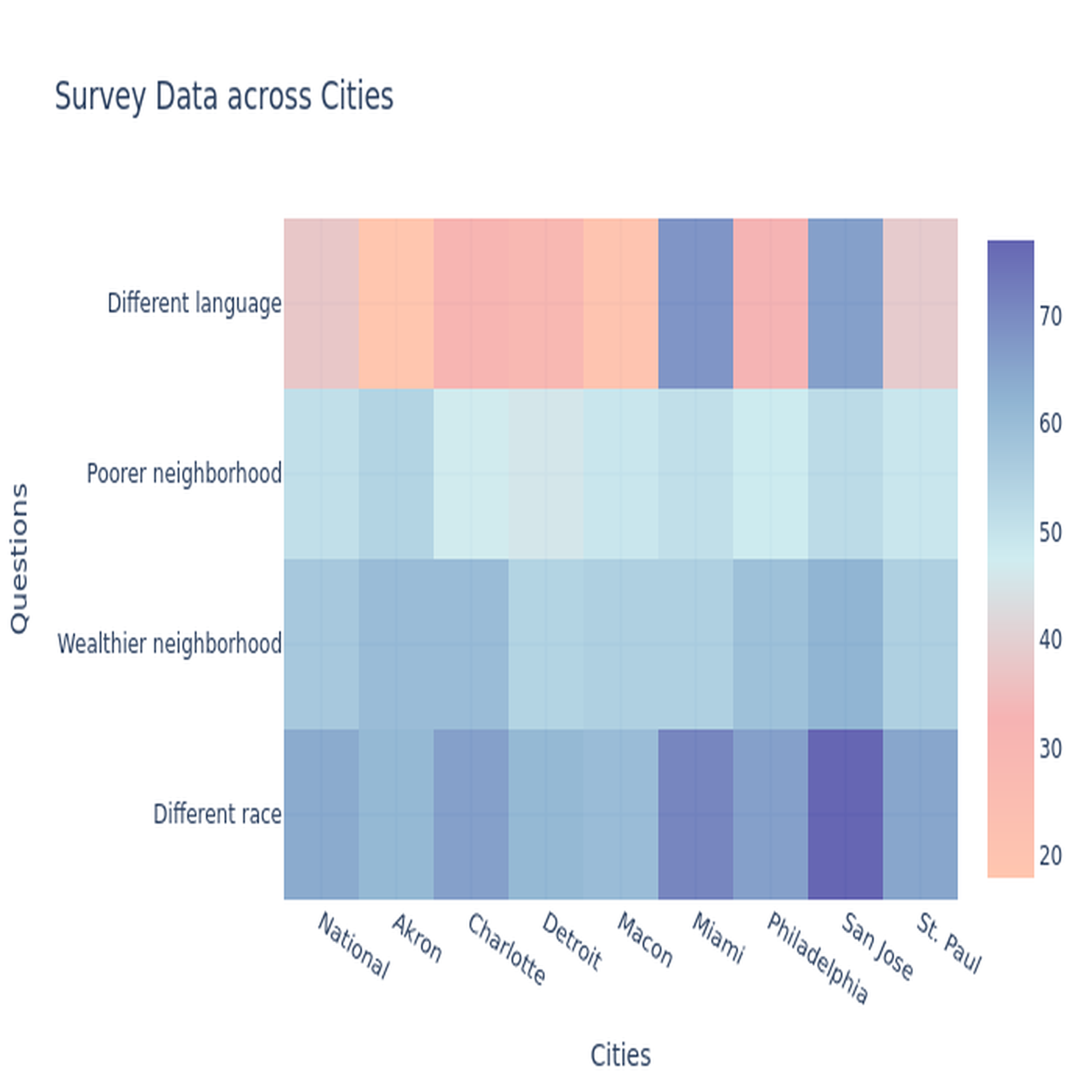}
    \caption{GPT4-o}
  \end{subfigure}
  \vspace{1em}
  \\
  \begin{subfigure}[t]{0.45\linewidth}
    \centering
    \includegraphics[width=\linewidth]{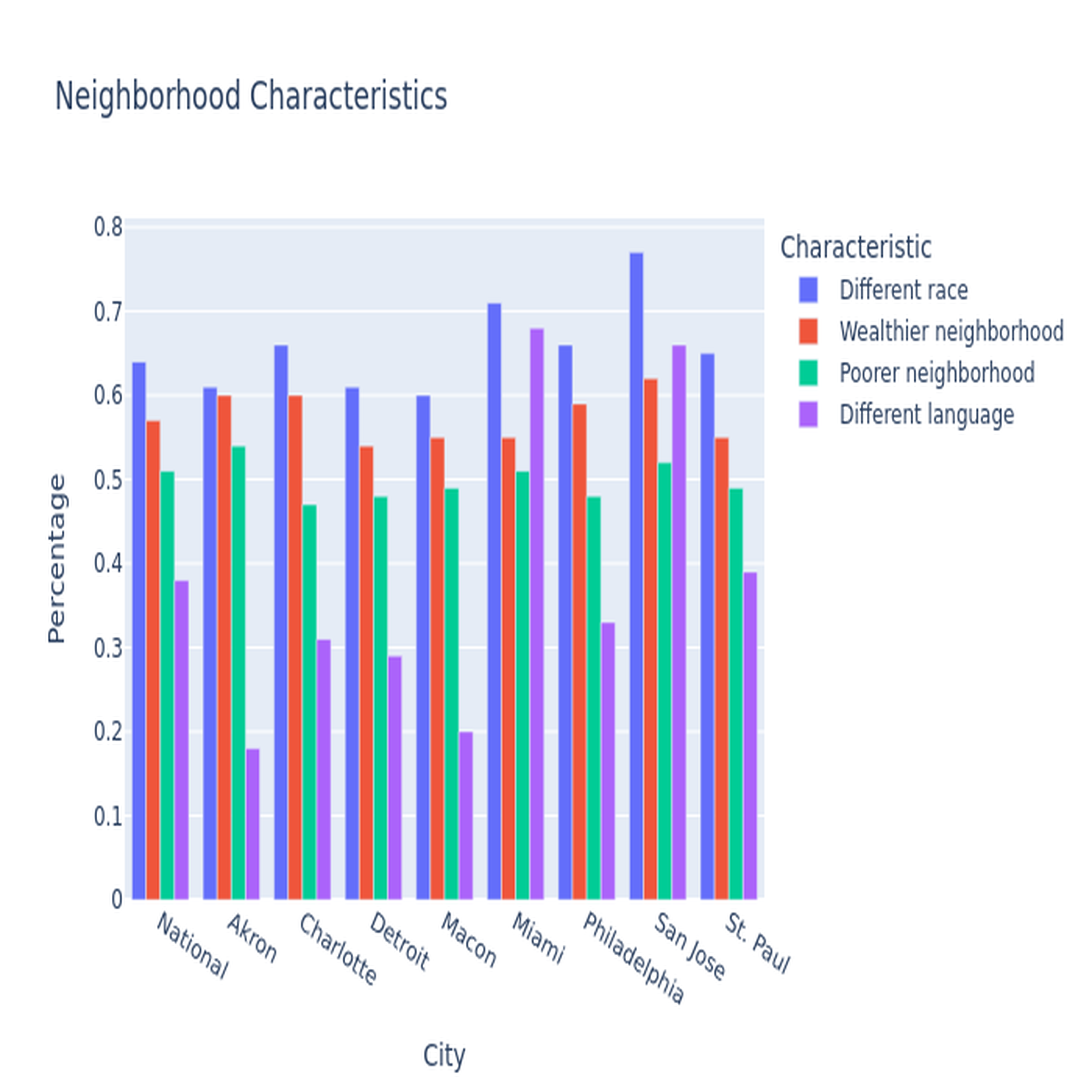}
    \caption{Gemma3-4b}
  \end{subfigure}
  \hspace{1em}
  \begin{subfigure}[t]{0.45\linewidth}
    \centering
    \includegraphics[width=\linewidth]{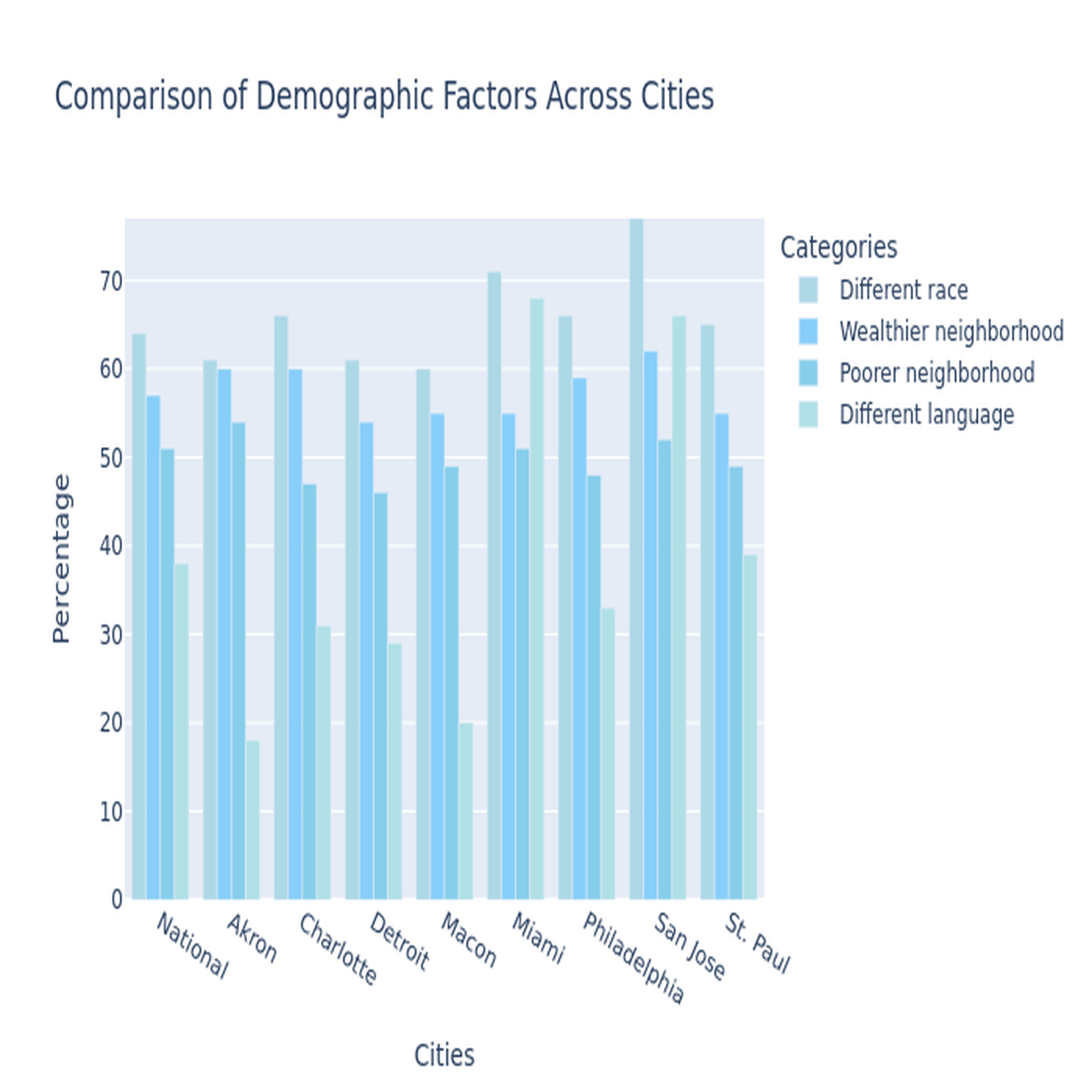}
    \caption{InternVL3-14b}
  \end{subfigure}
  \caption{Image Example6 for Chart-to-Code Task. }
  \label{fig:case6}
\end{figure*}

\textbf{Example 6}: As shown in figure~\ref{fig:case6}, GPT-4o is the only model that successfully reproduces the heatmap format of the gold image, including the layout of cities and categories. While the color mapping is slightly off, the structural format is well preserved. Gemma3-4b and InternVL3-14b both deviate significantly by converting the data into grouped bar charts. Although their bar heights roughly reflect the underlying values, they lose the compact matrix layout and visual impact of the original.

\begin{figure*}[htbp]
  \centering
  \begin{subfigure}[t]{0.23\linewidth}
    \centering
    \includegraphics[width=\linewidth]{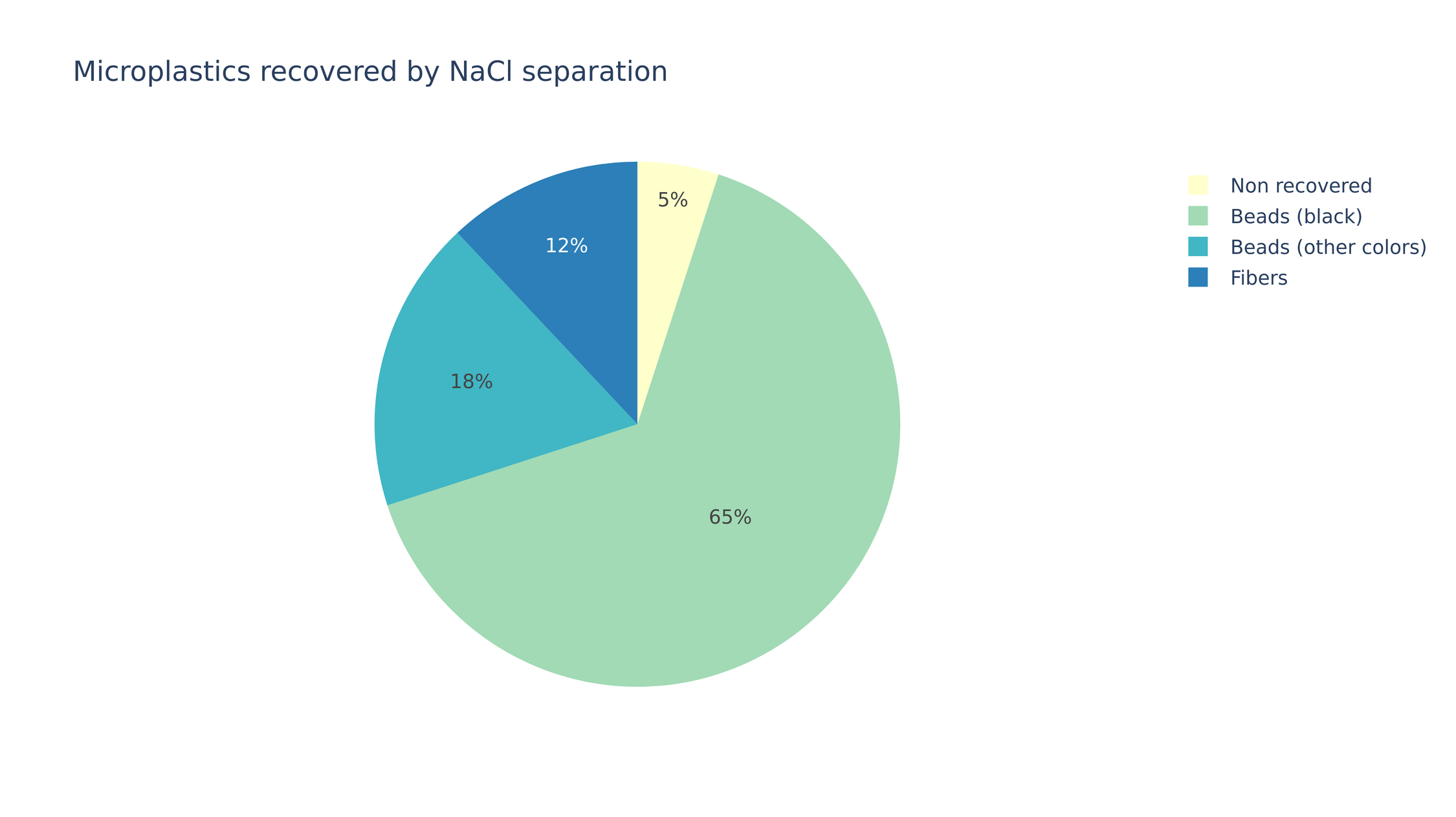}
    \caption{Gold Image}
  \end{subfigure}
  \hfill
  \begin{subfigure}[t]{0.23\linewidth}
    \centering
    \includegraphics[width=\linewidth]{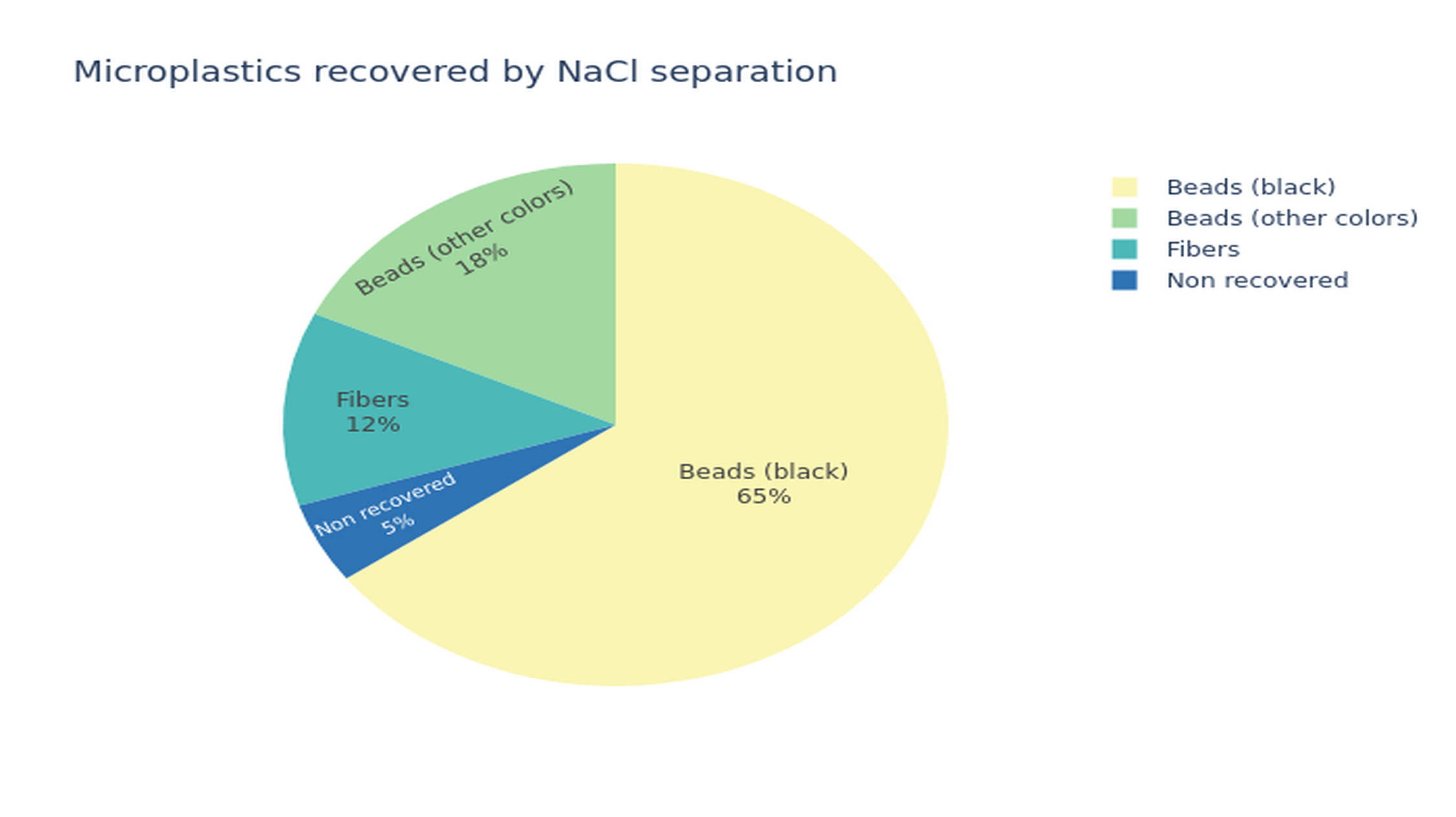}
    \caption{GPT-4o}
  \end{subfigure}
  \hfill
  \begin{subfigure}[t]{0.23\linewidth}
    \centering
    \includegraphics[width=\linewidth]{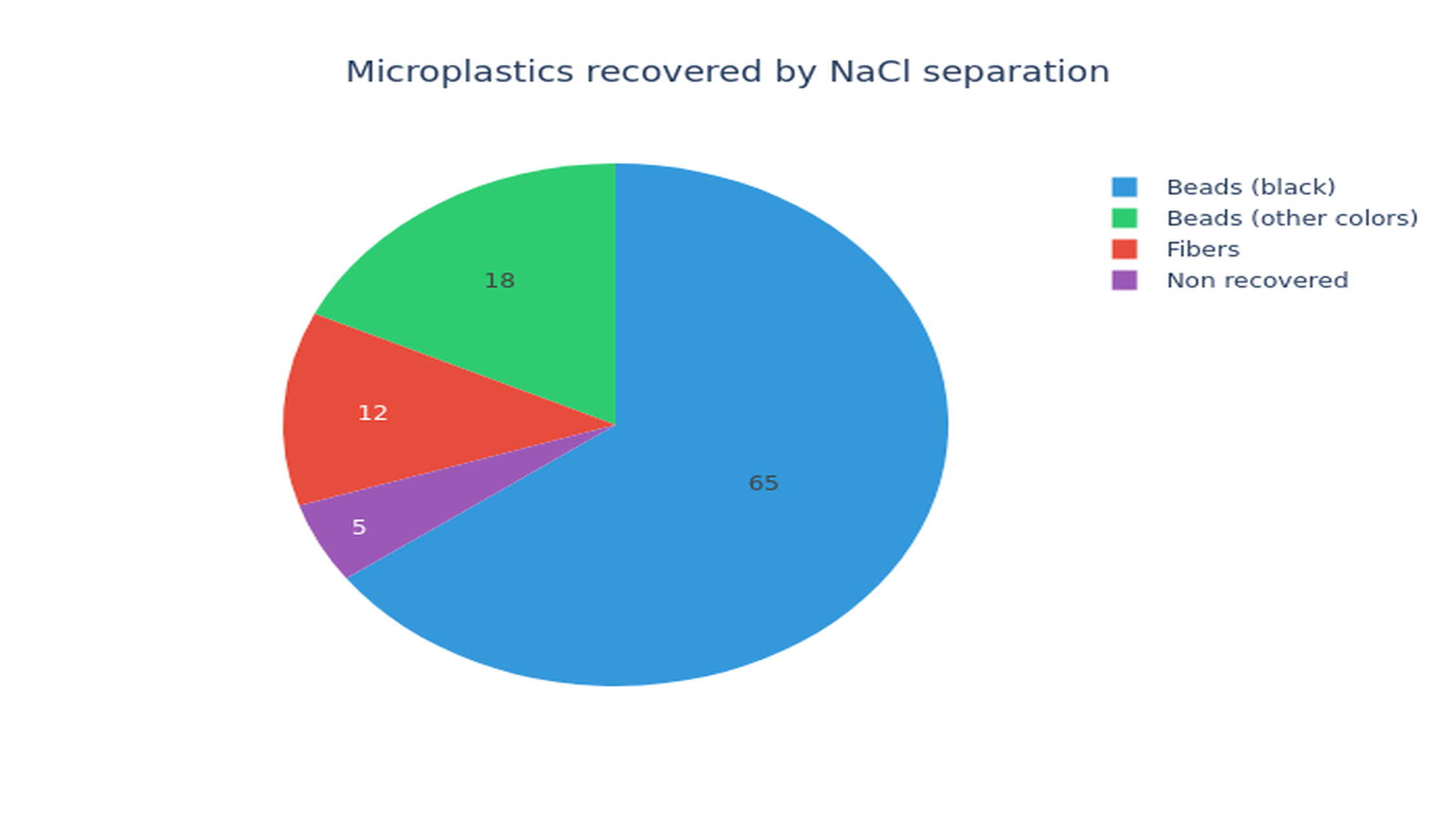}
    \caption{Gemma3-4b}
  \end{subfigure}
  \hfill
  \begin{subfigure}[t]{0.23\linewidth}
    \centering
    \includegraphics[width=\linewidth]{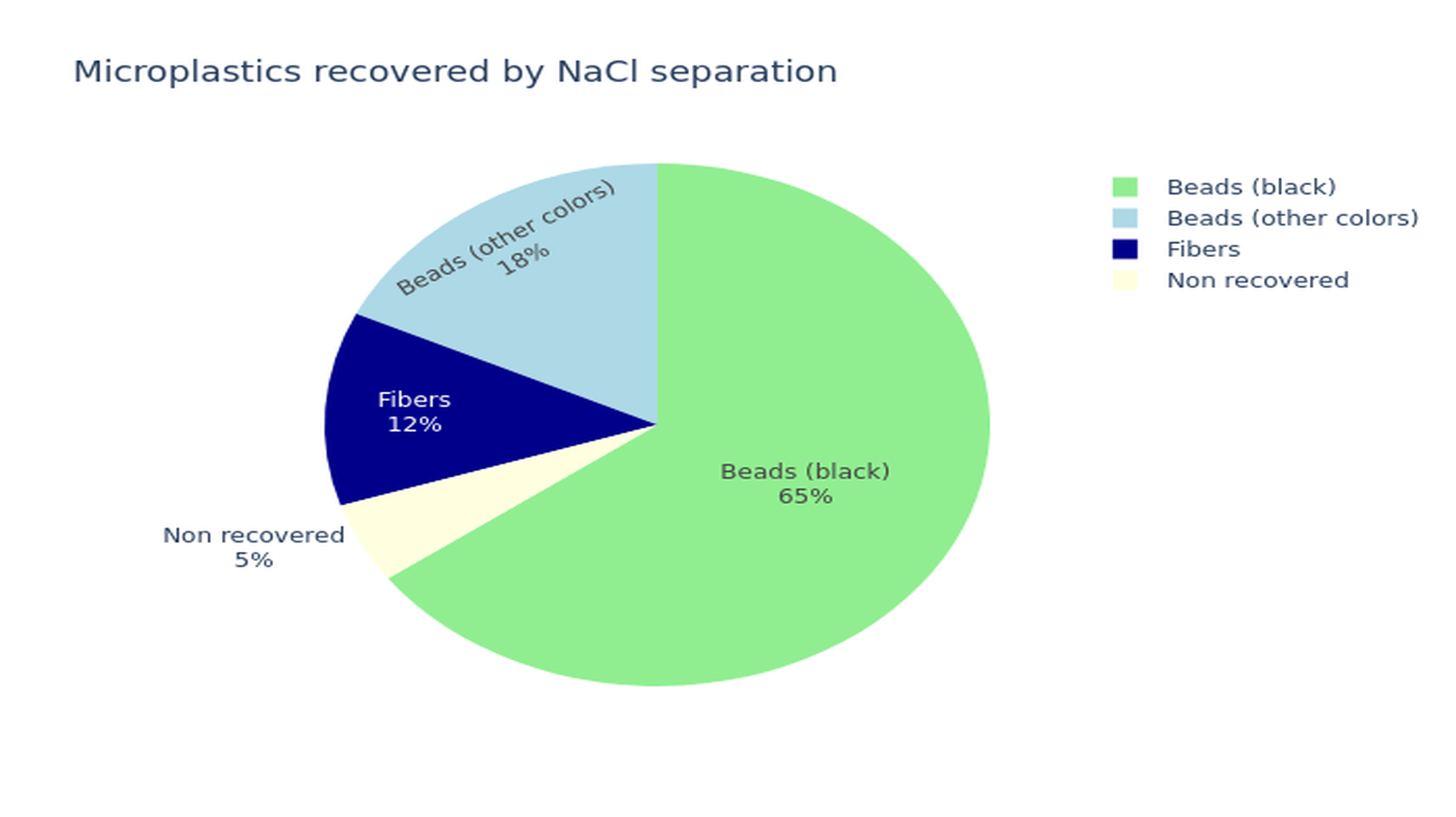}
    \caption{InternVL3-14b}
  \end{subfigure}
  \caption{Image Example7 for Chart-to-Code Task. Each subfigure shows the output from a different model.}
  \label{fig:case7}
\end{figure*}

\textbf{Example 7}: As shown in figure~\ref{fig:case7}, all models correctly reproduce a pie chart with the same numerical values as the gold image. GPT-4o and InternVL3-14b maintain both the correct labels and proportions, although GPT-4o’s layout is elliptical rather than circular. Gemma3-4b also retains the right proportions but uses a completely different color scheme and label placement. Overall, GPT-4o and InternVL3-14b achieve high fidelity, while Gemma3-4b is accurate in data but less faithful in appearance.

\begin{figure*}[htbp]
  \centering
  \begin{subfigure}[t]{0.48\linewidth}
    \centering
    \includegraphics[width=\linewidth]{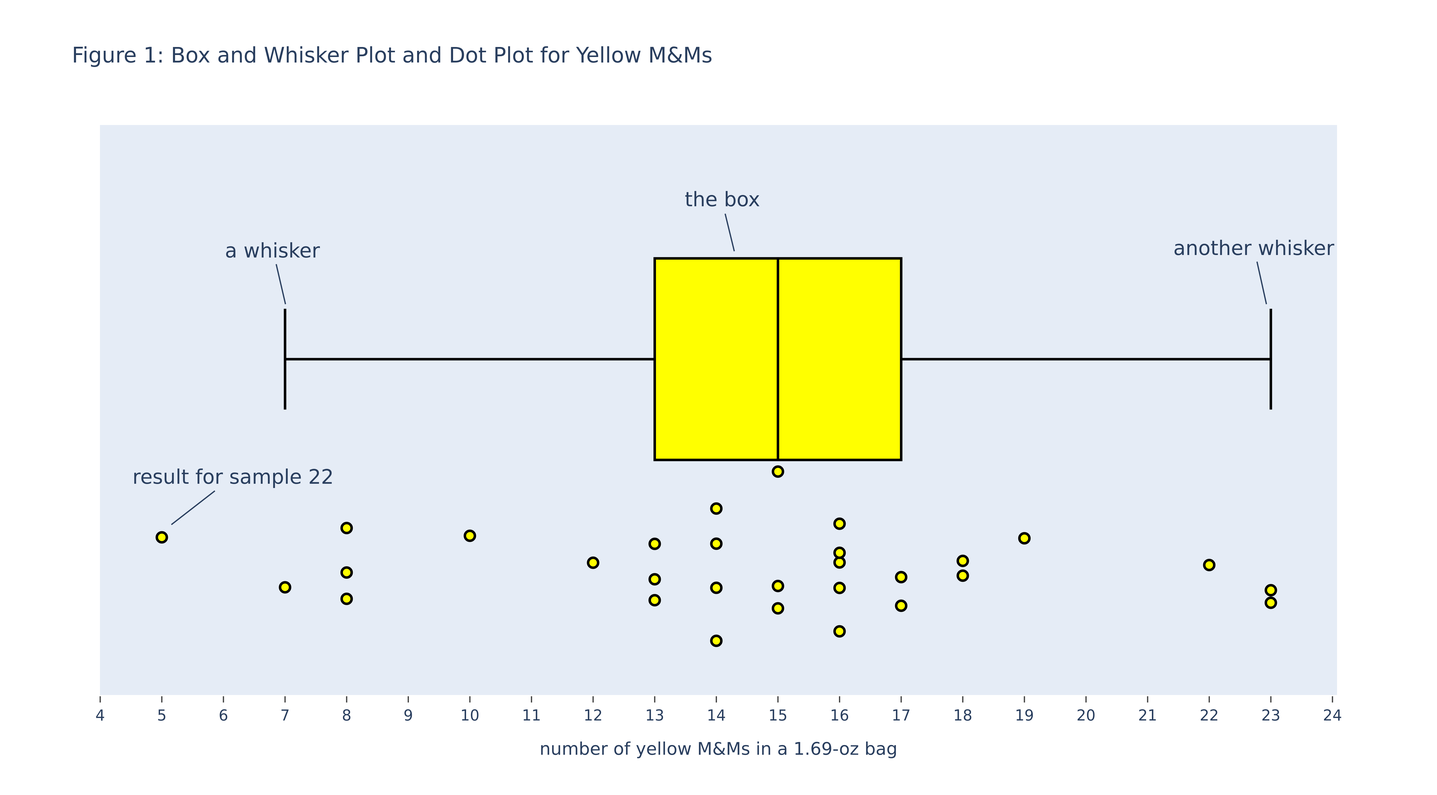}
    \caption{Gold Image}
  \end{subfigure}
  \hfill
  \begin{subfigure}[t]{0.48\linewidth}
    \centering
    \includegraphics[width=\linewidth]{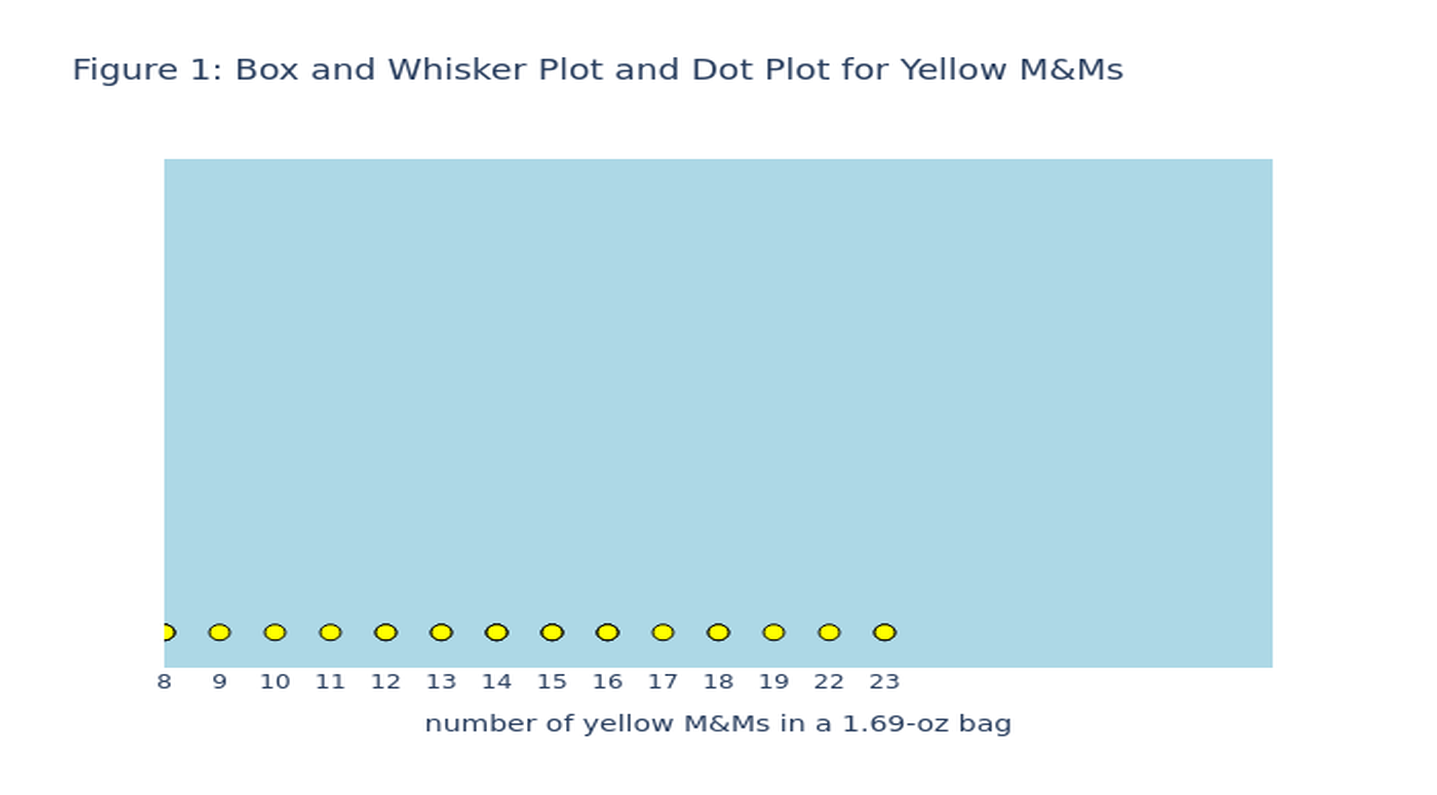}
    \caption{GPT4-o}
  \end{subfigure}
  \vskip\baselineskip
  \begin{subfigure}[t]{0.48\linewidth}
    \centering
    \includegraphics[width=\linewidth]{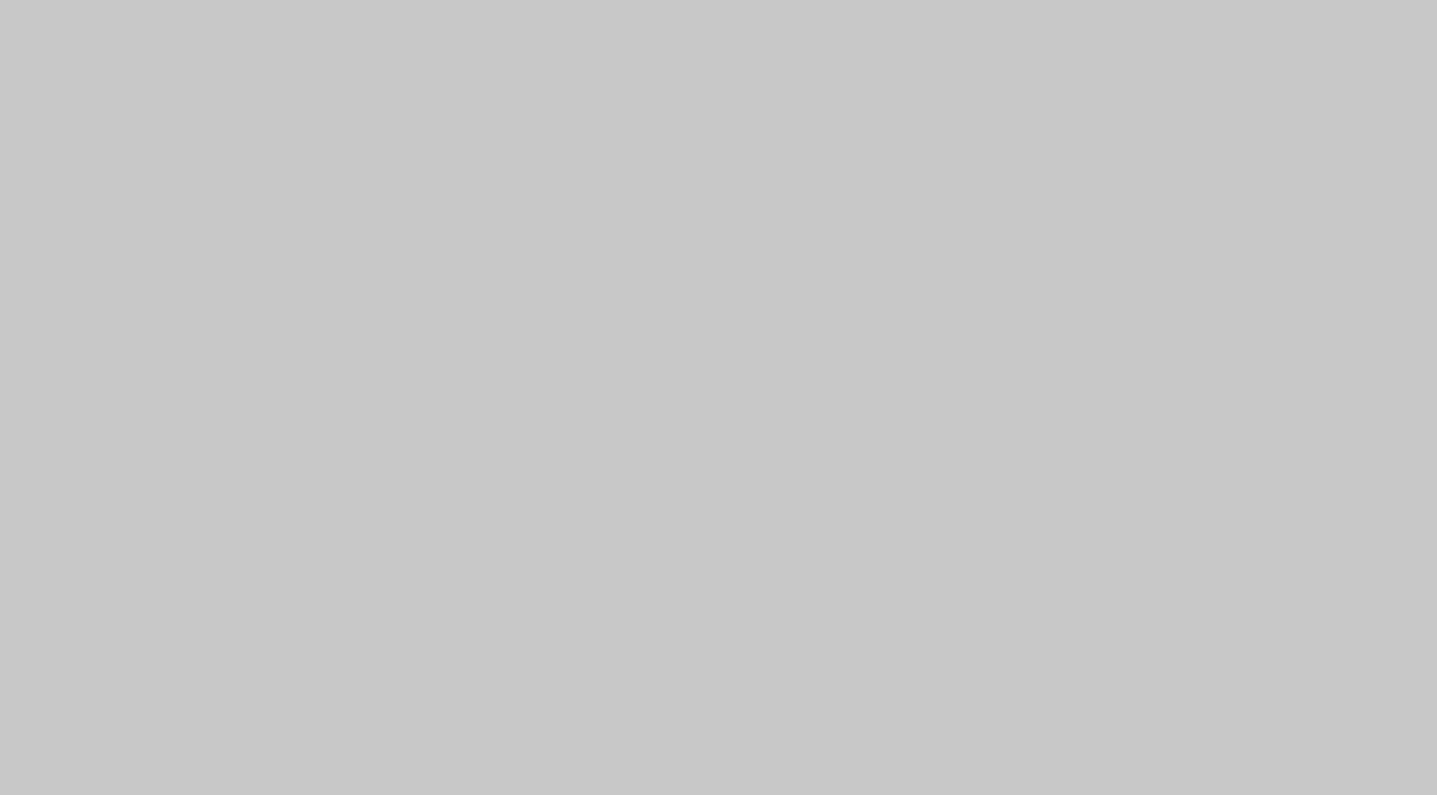}
    \caption{Gemma3-4b}
  \end{subfigure}
  \hfill
  \begin{subfigure}[t]{0.48\linewidth}
    \centering
    \includegraphics[width=\linewidth]{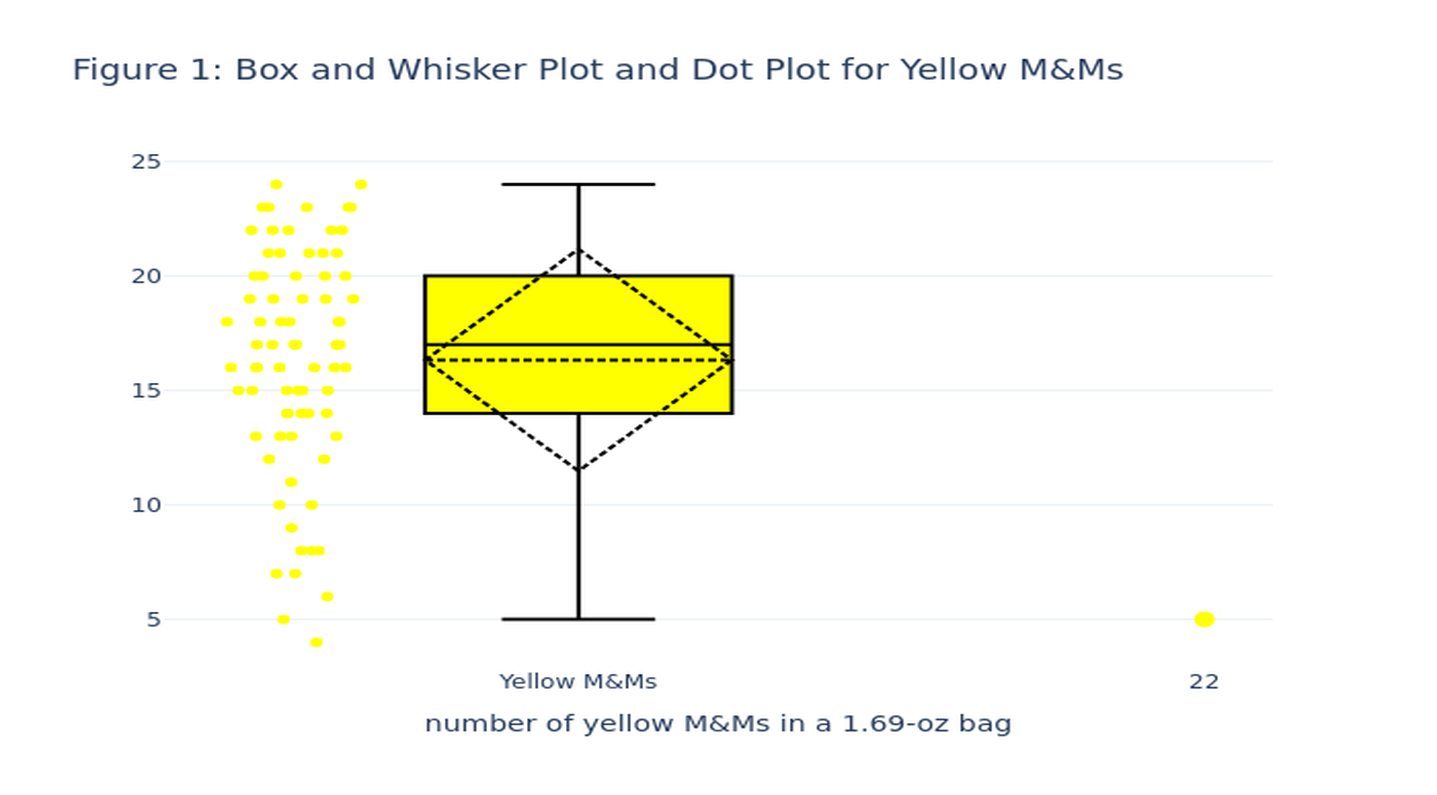}
    \caption{InternVL3-14b}
  \end{subfigure}
  \caption{Image Example8 for Chart-to-Code Task. The gray image indicates that the model did not
generate a valid image or the code parsing failed.}
  \label{fig:case8}
\end{figure*}

\textbf{Example 8}: From figure~\ref{fig:case8}, we can see, InternVL3-14b is the closest to the gold image, successfully generating both the box-and-whisker plot and the dot plot with a similar layout, color scheme, and annotated sample. GPT-4o captures only the dot plot and omits the box plot entirely, resulting in partial fidelity. Gemma3-4b fails to render any output, as shown by the gray placeholder. While InternVL3-14b slightly alters the orientation and spacing, it delivers the most accurate visual and structural reproduction overall.

\begin{figure*}[htbp]

\centering
\includegraphics[width=1.0\textwidth]{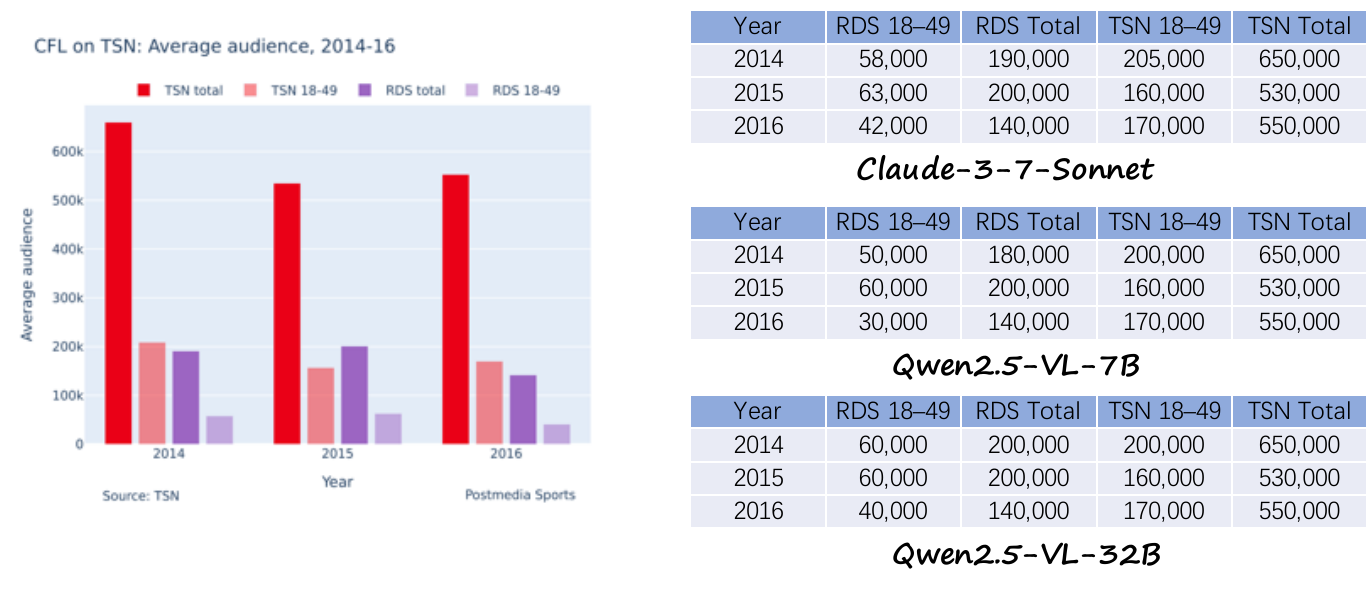}

\caption{Example9. Gold Image and Generated Tables for Controlled Chart-to-Table Task. } 
\label{fig:case9}
\end{figure*}

\textbf{Example 9}: From the gold table (Table~\ref{tab:gt1}) and the generated tables (Figure~\ref{fig:case9}), by comparing the row structures and numeric alignments, we observe that claude-3-7-Sonnet produces a table closely aligned with the gold table, with all values either correct or slightly off. Qwen2.5-VL-7B shows noticeable deviations in the "RDS 18–49" and "RDS Total" columns for 2014 and 2016. Qwen2.5-VL-32B gets closer but still introduces uniform values in 2015 that deviate from the gold data. Claude-3-7-Sonnet demonstrates the highest data fidelity and variation consistency across rows.

\begin{table}[htbp]
  \centering
  \caption{Example 9: Gold Table}
  \label{tab:gt1}
  \resizebox{\linewidth}{!}{%
    \begin{tabular}{ccccc}
    \toprule
    \textbf{Year} & \textbf{RDS 18\_49} & \textbf{RDS Total} & \textbf{TSN 18\_49} & \textbf{TSN Total} \\
    \midrule
    2014  & 58,000 & 191,000 & 209,000 & 660,000 \\
    2015  & 63,000 & 201,000 & 157,000 & 535,000 \\
    2016  & 41,000 & 142,000 & 170,000 & 553,000 \\
    \bottomrule
    \end{tabular}%
  }
\end{table}

\begin{table}[htbp]
\centering
\begin{tabular}{lcc}
\toprule
\textbf{theta} & \textbf{Today-r} & \textbf{Mean-r} \\
\midrule
AL29  & 66.87 & 44.09 \\
AL30  & 76.31 & 40.65 \\
AE38  & 26.49 & 39.08 \\
AL41  & 43.25 & 50.65 \\
AL35  & 40.96 & 45.12 \\
GD29  & 56.15 & 44.77 \\
GD30  & 51.42 & 43.54 \\
GD38  & 6.06  & 46.96 \\
GD41  & 25.54 & 51.53 \\
GD35  & 26.17 & 41.69 \\
\bottomrule
\end{tabular}
\caption{Example 10: Gold Table}
\label{tab:gtgt}
\end{table}

\begin{figure*}[htbp]

\centering
\includegraphics[width=1.0\textwidth]{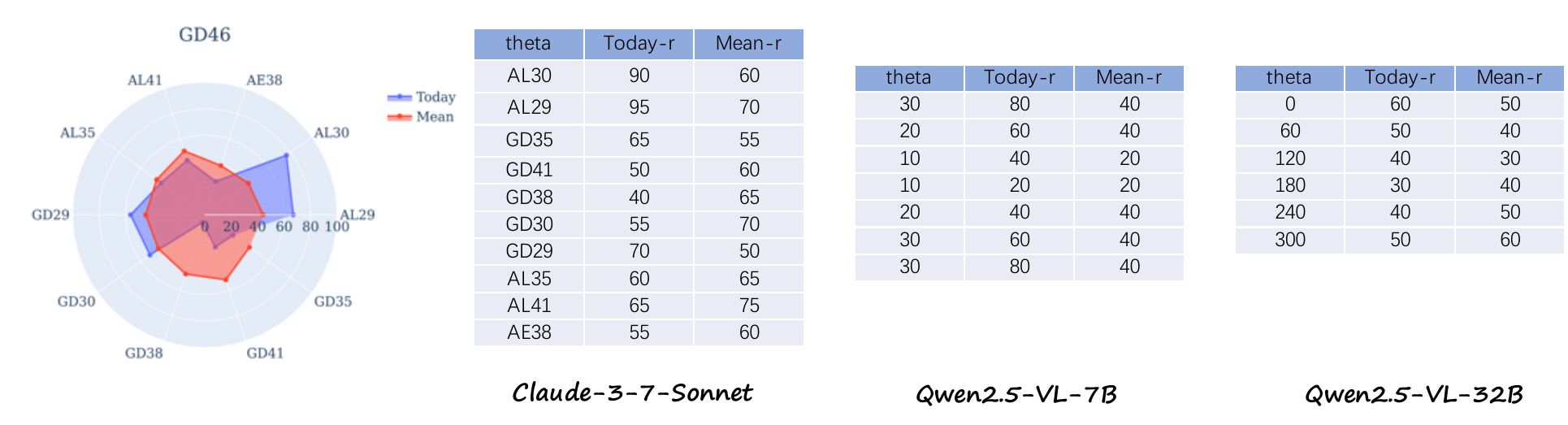}

\caption{Example10. Gold Image and Generated Tables for Controlled Chart-to-Table Task. } 
\label{fig:case10}
\end{figure*}

\textbf{Example 10}: From the gold table (Table~\ref{tab:gtgt}) and the generated tables (Figure~\ref{fig:case10}), we can see, in this example, Claude-3-7-Sonnet is the only model that correctly identifies the categorical theta labels (e.g., AL29, GD35) and extracts reasonable numerical values for both "Today-r" and "Mean-r," achieving the highest alignment with the gold table. In contrast, Qwen2.5-VL-7B and Qwen2.5-VL-32B fail to capture the label names and instead generate numerical theta angles, indicating a misunderstanding of the scatterpolar chart structure. Their values also significantly diverge from the reference data, limiting their usability for structured analysis.

\begin{figure*}[htbp]

\centering
\includegraphics[width=1.0\textwidth]{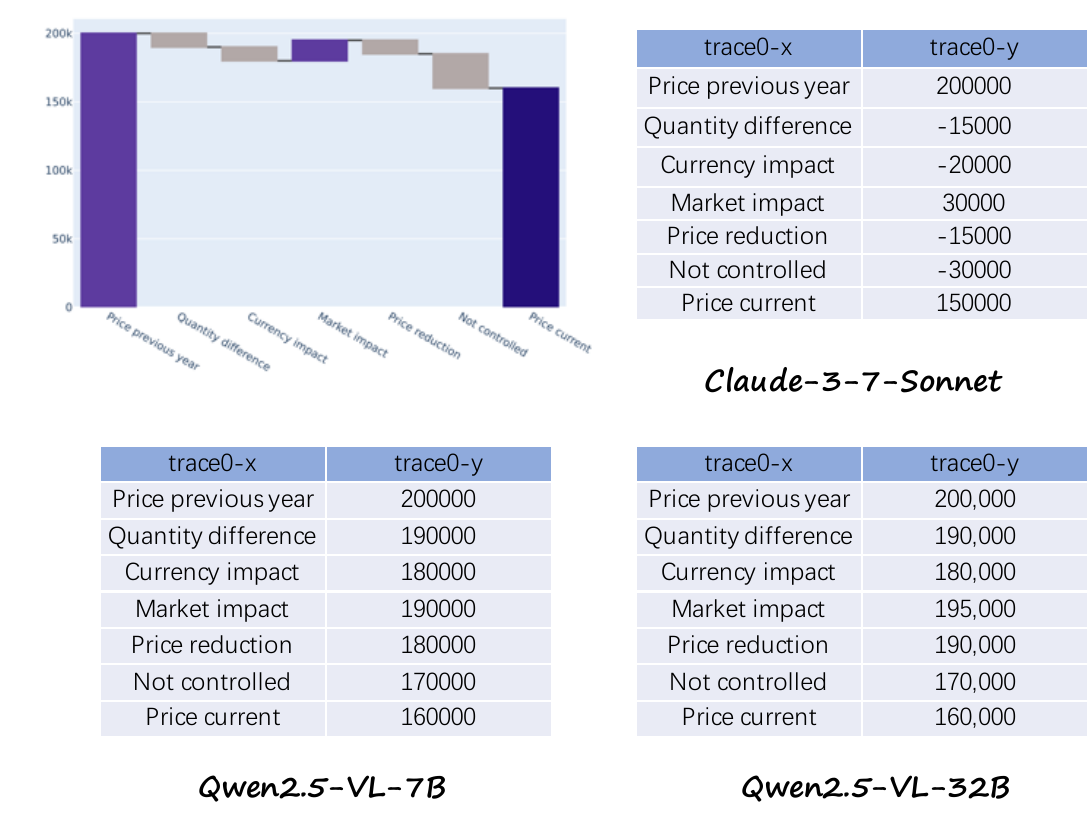}

\caption{Example11. Gold Image and Generated Tables for Controlled Chart-to-Table Task. } 
\label{fig:case11}
\end{figure*}

\textbf{Example 11}: From the gold table (Table~\ref{tab:gt3}) and the generated tables (Figure~\ref{fig:case11}), we can see, Claude-3-7-Sonnet is the only model that accurately captures both the structure and values of the gold table. It correctly identifies incremental changes (including positive and negative values) and matches the final "Price current" result. In contrast, Qwen2.5-VL-7B and Qwen2.5-VL-32B misinterpret the chart as cumulative values rather than stepwise deltas. As a result, their tables lose the core logic of a waterfall breakdown and deviate significantly from the reference.

\begin{table}[htbp]
\centering
\begin{tabular}{l r}
\toprule
\textbf{trace0-x} & \textbf{trace0-y} \\
\midrule
Price previous year & 200,000.0 \\
Quantity difference & -10,000.0 \\
Currency impact     & -10,000.0 \\
Market impact       & 15,000.0 \\
Price reduction     & -10,000.0 \\
Not controlled      & -25,000.0 \\
Price current       & 100,000.0 \\
\bottomrule
\end{tabular}
\caption{Example 11: Gold Table}
\label{tab:gt3}
\end{table}

%% file: Section/appendix/vis.tex
\section{Chart Examples}
\label{app:chart_examples}

\begin{figure*}[htbp]
\vspace{-8pt}
\centering
 
\includegraphics[width=1.0\textwidth]{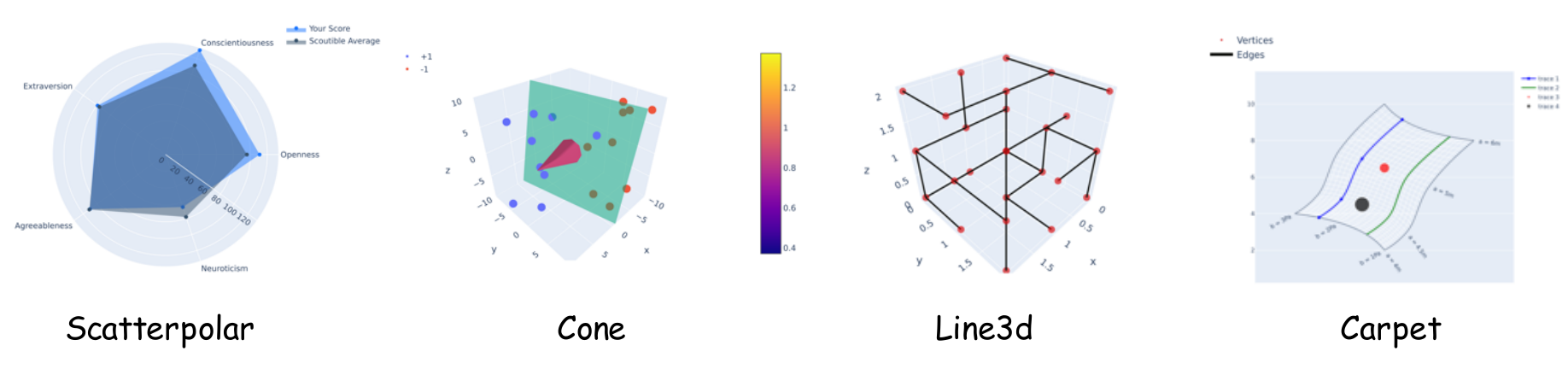}
\includegraphics[width=1.0\textwidth]{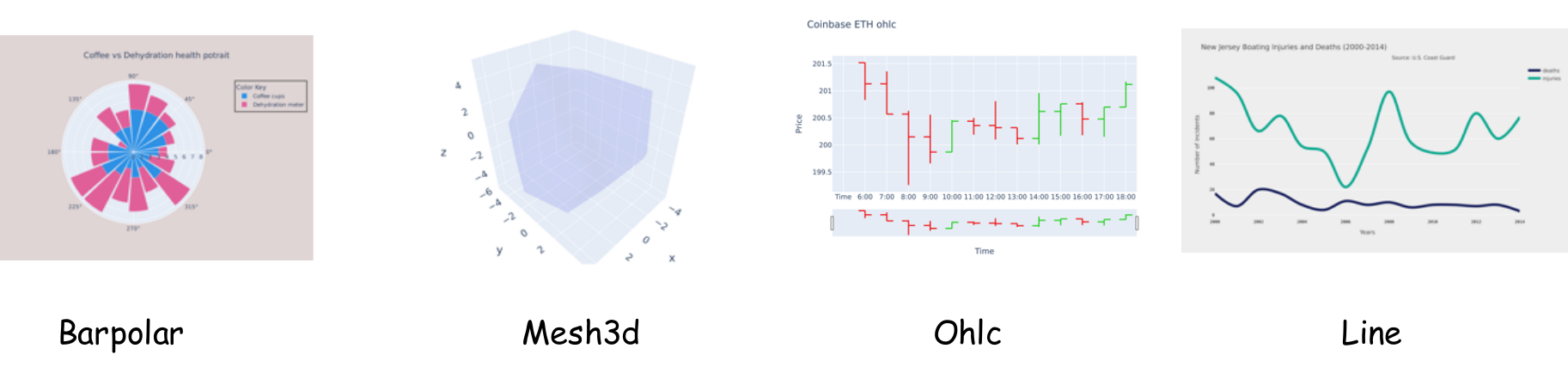}
\includegraphics[width=1.0\textwidth]{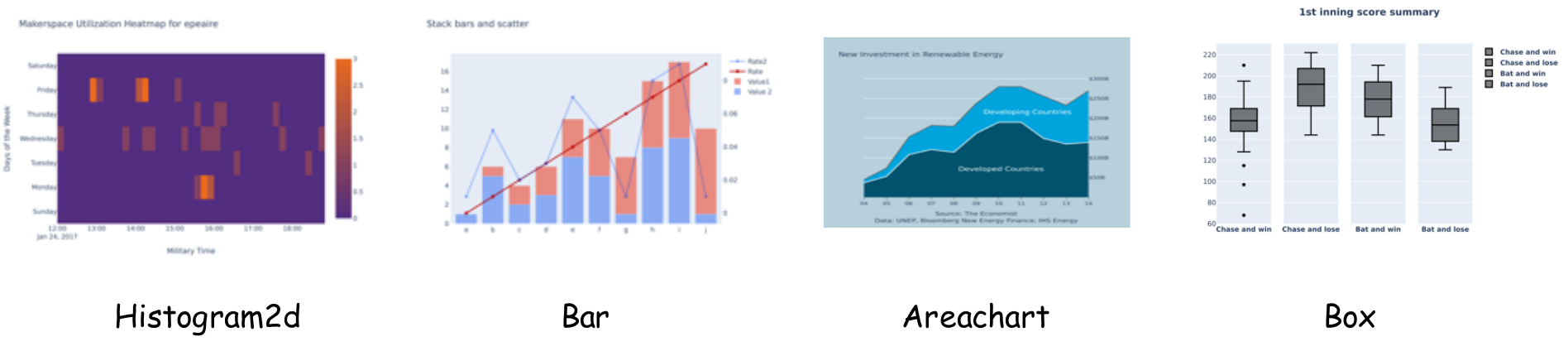}
\includegraphics[width=1.0\textwidth]{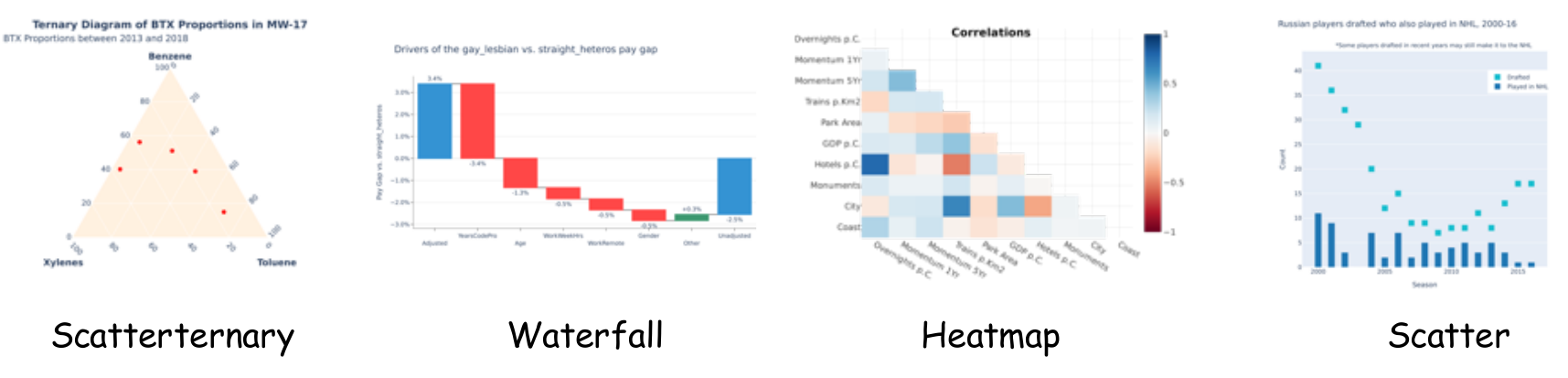}
\caption{Representative chart images in our benchmark.}

\label{fig:chart_examples_0}
\end{figure*}

\begin{figure*}[htbp]
\vspace{-8pt}
\centering

\includegraphics[width=1.0\textwidth]{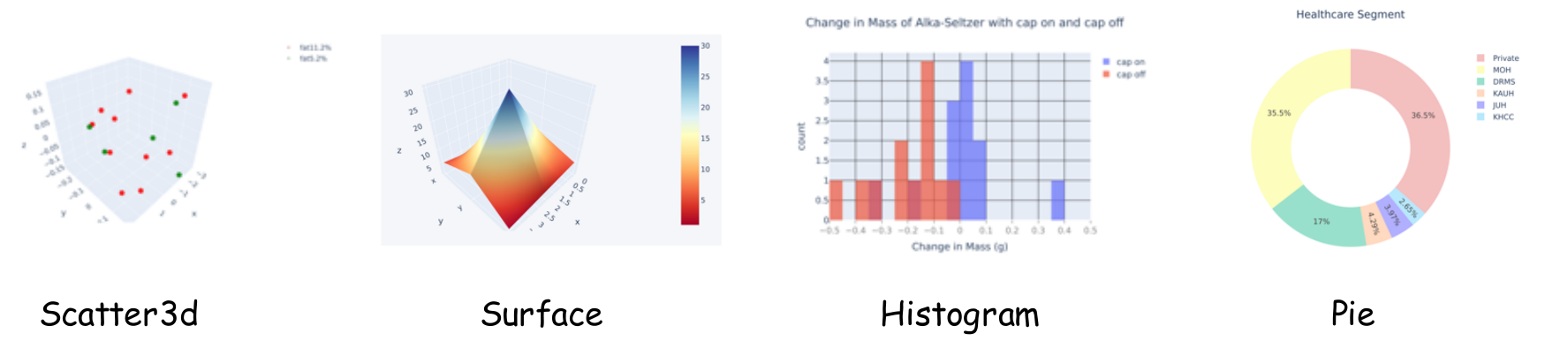}
\includegraphics[width=1.0\textwidth]{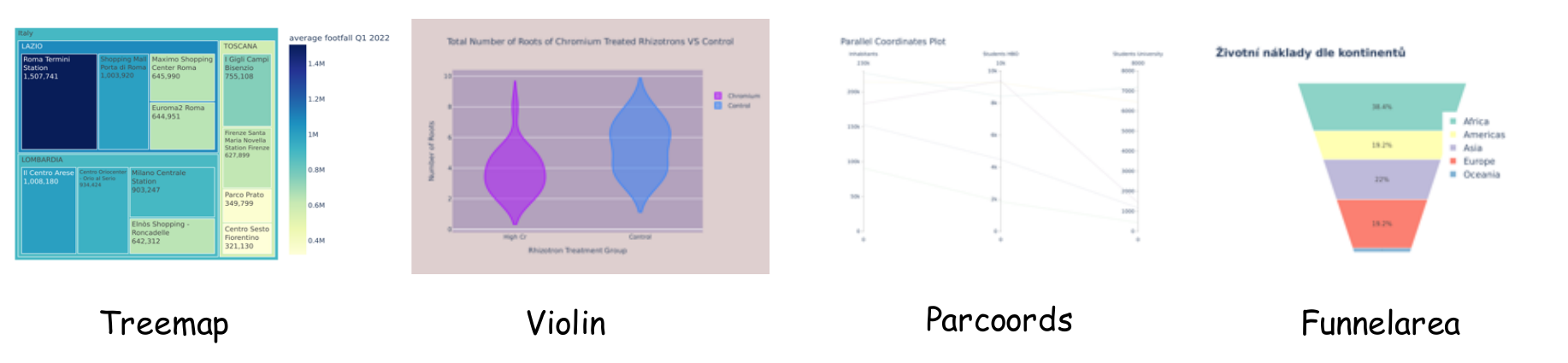}
 \includegraphics[width=1.0\textwidth]{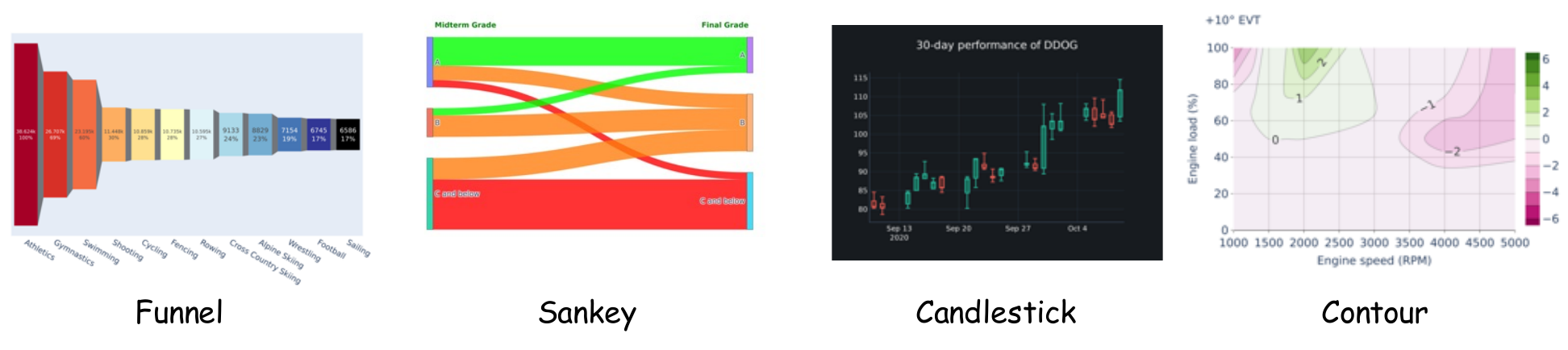}
    \includegraphics[width=1.0\textwidth]{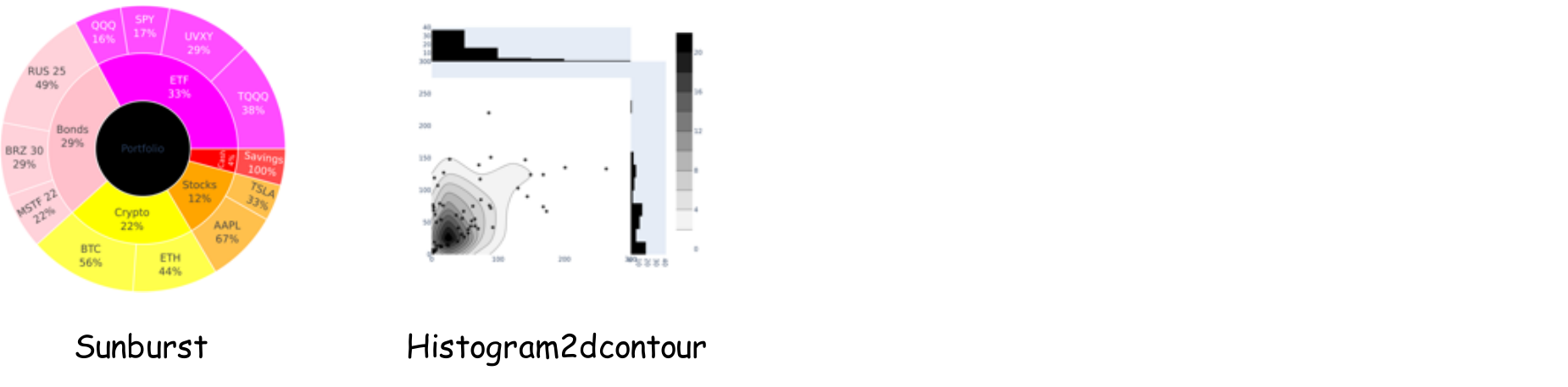}
    \caption{Representative chart images in our benchmark.}
\label{fig:chart_examples_1}
\end{figure*}

This section presents visual examples of the chart categories included in our benchmark. 
Each image illustrates a distinct chart type, showcasing the diversity in structure, data encoding, and visual design. 
Figures~\ref{fig:chart_examples_0} and Figures~\ref{fig:chart_examples_1} display representative samples from all 30 categories.

\section{Volunteer Recruitment and Payment}
All annotations were completed by five student volunteers from our institution. Given the light workload of the annotation task, participation was voluntary and uncompensated.

\section{Data Consent}
All data used in this work are publicly accessible and freely downloadable from their respective websites.

\section{AI Assistance Disclosure}
AI assistants were used solely for editorial assistance, including grammar and wording refinement, with all substantive content authored and reviewed by the authors.

\section{Data Screening and Privacy Protection}
We verified that all datasets used in this work are publicly available and do not contain personally identifiable information or offensive content. As the data are sourced from curated public repositories and websites, no additional anonymization or filtering was required.